\documentclass{article} 
\usepackage{iclr2026_conference,times}

\usepackage[utf8]{inputenc} 
\usepackage[T1]{fontenc}    
\usepackage{hyperref}       
\usepackage{url}            
\usepackage{booktabs}       
\usepackage{amsfonts}       
\usepackage{nicefrac}       
\usepackage{microtype}      
\usepackage{xcolor}         
\usepackage{amsmath, amsthm, bm}
\usepackage{comment}
\usepackage{enumitem}

\newtheorem{theorem}{Theorem}
\newtheorem{lemma}[theorem]{Lemma}

\usepackage{xfrac}

\newtheorem{open problem}[theorem]{Open problem}

\newtheorem{definition}{Definition}

\newcommand{\cB}{\mathcal{B}}
\newcommand{\cP}{\mathcal{P}}

\newcommand{\cW}{\mathcal{W}}
\newcommand{\cY}{\mathcal{Y}}

\newcommand{\e}{\mathbb{E}}
\newcommand{\p}{\mathbb{P}}
\newcommand{\R}{\mathbb{R}}
\newcommand{\vp}{\varphi}
\newcommand{\ve}{\varepsilon}

\newcommand{\hb}{\hat{\beta}}

\newcommand{\ld}{\dot{\ell}}

\newcommand{\rdd}{\ddot{r}}

\newcommand{\lddd}{\dddot{\ell}}

\newcommand{\diag}{\bm{{\rm diag}}}

\newcommand{\hbm}{\bm{ \hat \beta}_{\backslash \mathcal{M} }}

\newcommand{\bx}{\bm{x}}

\newcommand{\polylog}{{\rm polylog}}

\newcommand{\tbm}{\tilde{\bm \beta}_{ \backslash \mathcal{M}}}

\newcommand{\argmin}{\arg\min}
\newcommand{\RR}{\mathbb{R}}

\newcommand{\PP}{\mathbb{P}}
\newcommand{\EE}{\mathbb{E}}
\newcommand{\II}{\mathbb{I}}

\newcommand{\cD}{\mathcal{D}}
\newcommand{\cM}{\mathcal{M}}

\newcommand{\bb}{\bm{b}}

\newcommand{\bG}{\bm{G}}

\newcommand{\bSigma}{\bm{\Sigma}}
\newcommand{\bzero}{\bm{0}}

\newcommand{\bbeta}{\boldsymbol{\beta}}

\makeatletter
\def\namedlabel#1#2{\begingroup
    #2%
    \def\@currentlabel{#2}%
    \phantomsection\label{#1}\endgroup
}
\makeatother

\usepackage{hyperref}
\usepackage{url}
\usepackage{titlesec}
\usepackage{etoc}
\title{Gaussian certified unlearning in high dimensions: A Hypothesis Testing approach}

\author{Aaradhya Pandey\thanks{aaradhyapandey@princeton.edu corresponding author Department of Operations Research and Financial Engineering, Princeton University}, Arnab Auddy\thanks{auddy.1@osu.edu Department of Statistics, The Ohio State University}, Haolin Zou\thanks{hz2574@columbia.edu Department of Statistics,
Columbia University},
Arian Maleki \thanks{mm4338@columbia.edu Department of Statistics,
Columbia University}, Sanjeev Kulkarni\thanks{kulkarni@princeton.edu Department of Electrical and Computer Engineering and Department of Operations Research and Financial Engineering, Princeton University}
}

\iclrfinalcopy 
\begin{document}

\maketitle

\vspace{-.1 cm}
\begin{abstract}
Machine unlearning seeks to efficiently remove the influence of selected data while preserving generalization. Significant progress has been made in low dimensions $(p \ll n)$, but high dimensions pose serious theoretical challenges as standard optimization assumptions of $\Omega(1)$ strong convexity and $O(1)$ smoothness of the per-example loss $f$ rarely hold simultaneously in proportional regimes $(p\sim n)$.
In this work, we introduce $\varepsilon$-Gaussian certifiability, a canonical and robust notion well-suited to high-dimensional regimes, that optimally captures a broad class of noise adding mechanisms. Then we theoretically analyze the performance of a widely used unlearning algorithm based on one step of the Newton method in the high-dimensional setting described above. Our analysis shows that a single Newton step, followed by a well-calibrated Gaussian noise, is sufficient to achieve both privacy and accuracy in this setting. This result stands in sharp contrast to the only prior work that analyzes machine unlearning in high dimensions \citet{zou2025certified}, which relaxes some of the standard optimization assumptions for high-dimensional applicability, but operates under the notion of $\varepsilon$-certifiability. That work concludes 
that at least two steps are required to ensure both privacy and accuracy. Our result leads us to conclude that the discrepancy in the number of steps arises because of the sub optimality of the notion of $\varepsilon$-certifiability and its incompatibility with noise adding mechanisms, which $\varepsilon$-Gaussian certifiability is able to overcome optimally.
\end{abstract}
\vspace{-.1 cm}

\vspace{-.2 cm}
\section{Introduction and Theoretical contributions}
\vspace{-.2 cm}

\textbf{Overview and practice:} Modern ML models (from healthcare tools to systems like ChatGPT) are trained on data that often includes sensitive personal information, creating risks of memorization and leakage. Laws such as \citet{GDPR,CCPA,CPPA2022} enforce a ``right to be forgotten," requiring removal of both the data and its statistical influence on trained models. Because full retraining is costly, machine unlearning develops efficient methods to remove data traces while preserving model generalization. Recent years have seen rapid progress, fueled by substantial empirical results and rigorous theory \citet{cao_yang_2015}, \citet{bourtoule2021machine}, \citet{chen2021when}, \citet{gupta2021adaptive}, \citet{pmlr-v134-ullah21a}, \citet{nguyen2022survey}, \citet{chundawat2023zero}, \citet{warnecke2023machine}, \citet{tarun2024fast}, \citet{wang2024machine}, \citet{liu2025rethinking}.
 \par
 \textbf{Algorithms in theory:} In addition to substantial progress in the practical development of machine unlearning algorithms, their theoretical understanding has also advanced significantly in recent years. For example, \citet{pmlr-v119-guo20c} and \citet{sekhari2021remember} studied \textit{Newton-based} unlearning algorithms and concluded that a single Newton update, followed by appropriately calibrated noise, is sufficient to ensure privacy for users requesting data removal while preserving the accuracy of the resulting estimates. In addition, \citet{neel2021descent}, \citet{allouah2025the} examined the performance of approximate unlearning algorithms based on \textit{gradient descent} and \textit{stochastic gradient descent}.
 \par
\textbf{Low dimensionality:} An important caveat in these theoretical papers is they \textit{implicitly assume} the number of model parameters $p$ is significantly smaller than the number of observations $n$, by putting \textit{standard optimization assumptions on the loss, which fails to hold in high-dimensional proportional regimes}. \textit{We prove this point later in this section}. So, these results are not useful for many AI applications, wherein the number of model parameters $p$ is very large and comparable to the sample size$ n$. We refer to this as the high-dimensional setting, in contrast to the low-dimensional setting considered in \citet{pmlr-v119-guo20c}, \citet{sekhari2021remember},  \citet{neel2021descent}, and \citet{allouah2025the}.
\par
\textbf{Theoretical contributions of this paper:} We \textit{introduce} the canonical notion of $(\phi,\varepsilon)$ -Gaussian certifiability \eqref{sec:Gauss_cert} that is particularly well suited for high dimensions, and \textit{analyze} the performance of a widely used machine unlearning algorithm in the underexplored proportional regime $p\sim n$ . This requires us to relax the stringent standard optimization assumptions on the per-example loss $f$ in all previous works, including  \citet{pmlr-v119-guo20c}, \citet{sekhari2021remember}, \citet{neel2021descent}, \citet{allouah2025the}. In this setting, our work makes the following \textit{novel theoretical contributions}:
 \par
\textbf{Certifiability:} Inspired by the hypothesis testing interpretation of differential privacy 
\citet{wasserman2010statistical},
\citet{pmlr-v37-kairouz15}, \citet{liu2019investigating}, \citet{balle2020hypothesis},
\citet{bu2020deep},
\citet{dong2021central}, \citet{awan2022logconcave}, \citet{awan2023canonical}, \citet{dong2022gaussian}, we introduce a \textit{practical}, \textit{optimally achievable}, \textit{statistically interpretable}, and \textit{robust} notion of certifiability,  called $\ve$-Gaussian certifiability. This notion is particularly \textit{well-suited for developing theory} in high-dimensional situations. In fact, as proved in \citet{dong2021central}, a broad class of noise-adding mechanisms converge in behavior to $\varepsilon$-Gaussian certifiability as the dimensionality increases. This makes  $\varepsilon$-Gaussian certifiability the \textit{canonical} framework of certifiability in high dimensions. 
\par
\textbf{Performance: } Similar to \citet{pmlr-v119-guo20c, sekhari2021remember}, and \citet{zou2025certified}, we then theoretically investigate a class of unlearning algorithms that perform a few steps of Newton's method, rather than fully re-optimizing the empirical risk. In contrast to \citet{pmlr-v119-guo20c}, \citet{sekhari2021remember}, \citet{allouah2025the}, our studies explore the theoretical behavior of these methods in \emph{high‐dimensional} proportional setting of  $p \sim n$, by \textit{relaxing the standard optimization assumptions of $\Omega(1)$ strong convexity and $O(1)$ smoothness of the chosen per-example loss}. We prove that a \textit{single} Newton update with carefully calibrated independent Gaussian noise, simultaneously attains $\varepsilon$-Gaussian certifiability and incurs only a vanishing degradation in generalization error. Note that while  \citet{sekhari2021remember, pmlr-v119-guo20c} also concluded that a \textit{single Newton step is sufficient} for approximate removal of a single datapoint, but due to \textit{their restrictive assumptions on the loss}, their results fail completely in high-dimensions, even in the simplest setting of ridge regularized least squares. We explain this failure in detail next and also in Section ~\ref{sec:comparison} of the appendix. 
\par
\textbf{Removing multiple data points:} Our theory encompasses the simultaneous removal of data from $m$ users and show that, even when the number of removal requests $m$ grows with the sample size, accurate unlearning is still possible in high-dimensional settings provided that $m = o\!\bigl(n^{1/4}/\mathrm{polylog}(n)\bigr)$.
\par
\textbf{Comparison of conclusions and failure of assumptions in previous works:} \label{ssec:theory} In the remaining part, first, we show why the standard optimization assumptions of the per-example loss $f$ in  \citet{allouah2025the,sekhari2021remember} break down in high dimension. Then we compare our theoretical conclusions with \citet{zou2025certified}, as it is most closely aligned with ours in high dimensions.
\par
\textbf{Failure of the assumptions in \citet{allouah2025the}, \citet{sekhari2021remember}:}  Let $\mathbf{z}_i = (\bx_i,y_i)\in \mathbb{R}^{p+1}$ denote the \textit{feature-response} pair $(\bx_i$, $y_i)$ of datapoint $i\in [n]$. Both \citet{allouah2025the} and \citet{sekhari2021remember} consider estimating the parameters of the model, $\bbeta\in \mathbb{R}^p$ through the following optimization problem:
$\hat{\bbeta}= \arg\min_{\bbeta}  \frac{1}{n} \sum_{i=1}^n f(\bbeta, \mathbf{z}_i)= \arg\min_{\bbeta} \sum_{i=1}^n f(\bbeta, \mathbf{z}_i), \text{ where}$
  $f(\bbeta, z_i)$ is the per-example cost with  $n$ datapoints. In their theory \citet{allouah2025the} 
 assumes\footnote{\citet{sekhari2021remember} assumes a variation of standard optimization assumptions on per-example loss $f$ whose conclusions blow up in high dimensions as well. This important case is treated in Section ~\ref{sec:comparison} of the appendix.} that  $f$ satisfies the standard optimization assumptions. More precisely, they assume  $\bbeta \to f(\bbeta, \mathbf{z}_i)$  is simultaneously (1) worst case $\mu=\Omega(1)$ strongly convex $\iff \nabla^2 f(\bbeta, \mathbf{z}_i)\succeq \mu I_p $ for all $\mathbf{z}_i$, and (2) worse case $L=O(1)$ smooth $ \iff \nabla^2 f(\bbeta, \mathbf{z}_i) \preceq LI_p $ for all $\mathbf{z}_i$. These two  cannot hold simultaneously in high-dimensional proportional regime $p\sim n$. In fact these are even violated in simple examples such as linear regression. To prove this, we consider one of the simplest estimators based on ridge regularized least square loss: $\sum_{i=1}^n f(\bbeta, \mathbf{z}_i)= {\lambda}\|\bbeta\|^2 + \sum_{i=1}^n (y_i-\bx_i^T\bbeta)^2$. A computation of the Hessian gives $\nabla^2 \left( \sum_{i=1}^n f(\bbeta, \mathbf{z}_i)\right)= 2X^TX + 2\lambda I_p $.  In high dimensional proportional regime $p \sim n$ that is studied in this paper, the condition  $L=O(1)$ smoothness constant of the per-example loss $f$ inevitably requires  the  scaling  $\|\bx_i\|_2 \sim 1$ (with high probability) for all $i \in [n]$. But, this completely destroys the $\mu=\Omega(1)$ strong convexity of the per-example loss $f$, as $\lambda_{min} \left(\nabla^2 f(\bbeta, \mathbf{z}_i)\right) = 2\frac{\lambda}{n}$. A detailed explanation of this is provided in Section ~\ref{sec:comparison} of the appendix.
\par
\textbf{Comparing with conclusions of \citet{zou2025certified}:} In high dimensions, the privacy-accuracy interplay is subtler than in low dimensions.  \citet{zou2025certified} is the only theoretical work we know of that has studied  machine unlearning when $p\sim n$ by  \textit{relaxing some of the standard optimization assumptions on the loss}, relative to previous works with stringent assumptions \citet{allouah2025the},\citet{sekhari2021remember}. Our rigorous result that \textbf{one noisy Gaussian-Newton step suffices} \eqref{eq:randomization_step} to unlearn an RERM \eqref{eq:betahat} in high dimensions under $(\phi_n, \varepsilon)$-GPAR \ref{sec:Gauss_cert} stands in contrast with \cite{zou2025certified}. Their theoretical analysis relies on the $(\phi, \varepsilon)$-PAR  certifiability (section \ref{ssec:hypothesistesting}) and shows that even for removing a single data point, \textbf{at least two Newton steps are required} to ensure the accuracy. The sharp contrast between the analytical results highlights the tightness of our GPAR framework \ref{sec:Gauss_cert} in high dimensions.  The \textit{suboptimality of $\varepsilon$-certifiability stems from its incompatibility with natural noise addition strategies}, thereby necessitating the injection of disproportionately large noise to meet the certifiability criteria, an effect that is particularly pronounced in high-dimensional regimes.

\vspace{-.2cm}
\section{The machine unlearning problem: privacy and accuracy} \label{sec:machunle}
\vspace{-.2cm}

\textbf{The problem:} Let $\cD_n$ denote the dataset, $A(\cD_n)$ denote the machine learning model trained on $\cD_n$, and $\cM\subset[n]$ the indices of points to be removed, with corresponding data $\cD_{\cM}$. The \textit{problem of machine  unlearning} requires a procedure $\bar{A}$ to produce an update $\bar{A}_{\setminus \cM}:=\bar{A}(A(\cD_n),\cD_{\cM},T(\cD_n))$ that closely approximates the \textit{ideal}  model $A(\cD_n \setminus \cD_{\cM})$ retrained from scratch. We assume that $\bar{A}$ has access to the model output $A(\cD_n)$, the dataset of removal requests $ \cD_{\cM}$ and some auxiliary gradient or Hessian information $T(\cD_n)$ as in \cite{sekhari2021remember} and \cite{zou2025certified}.
\par
\textbf{Two approaches (approximate versus exact unlearning) and our conceptual focus:} We focus on  \emph{approximate unlearning} (e.g. \cite{pmlr-v119-guo20c}, \cite{sekhari2021remember}, \cite{neel2021descent}, \cite{allouah2025the}) where, $\bar{A}$ achieves \textit{statistical proximity} to the retrained model as opposed to \emph{exact unlearning} \cite{cao_yang_2015}, \cite{bourtoule2021machine},\cite{chowdhury2025s3t}, \cite{kuo2025sift}, which requires literal equality $\bar{A}_{\setminus \cM}=A(\cD_n \setminus \cD_{\cM})$ (almost surely or in distribution). Exact recomputation is often computationally infeasible, and since our \textit{focus is on extending the theory of the proposed unlearning framework to a large class of loss functions in high-dimensional problems of interest}, approximate removal methods offer a practical and scalable alternative. Note that the machine unlearning mechanism $\bar{A}_{\setminus \mathcal{M}}$ is expected to satisfy the following two properties: \par
\vspace{-.1 cm}
 \textbf{Protect user privacy:} The output of the unlearning procedure \(\bar{ A}_{\setminus \cM}\) must leak \textit{no information} about the removed subset \(\cD_{\cM}\), even under arbitrary post‐processing of the unlearned output. \par
\vspace{-.1 cm}
 \textbf{Preserve model accuracy:} The unlearned model \(\bar{A}_{\setminus \cM}\) must incur only a \textit{vanishing change} in generalization capabilities compared to the ideal retrained from scratch model \(A(\cD_n \setminus \cD_{\cM})\).

 \vspace{-.2cm}
\subsection{Generalized linear model (of data) and empirical risk minimization }\label{sec:erm-intro}
\vspace{-.2cm}

 We now describe our high-dimensional setup as a step  towards formalizing privacy and accuracy.
 \par
\textbf{GLM Data:} We assume our data  $ \cD_n := \{(\bx_i,y_i)\}_{i=1}^n$ with $\bx_i \in \R^p$ denoting the \textit{features} and $y_i\in\R$ denoting the \textit{response}. 
Further, $(\bx_i,y_i)$ are  
i.i.d. samples from $ dP_{\bbeta^*}(\bx,y)= dp(\bx) dq(y|\bx^{\top}\bbeta^*)$ with \textit{density} $p: \RR^p \to \R_{\geq 0}$ and \textit{kernel} $q(\cdot|\bx^{\top}\bbeta^*) :\cY \to \R_{\geq 0}$ with \textit{unknown} $\bbeta^* \in \cB_p:=\R^p$.
\par
\textbf{High-dimensional focus:} In many AI applications, the number of model parameters, $p$, is often comparable to the sample size, $n$. So, we aim to study models under the \textit{proportional high dimensional setting} in which both $n,p$ are large while $n/p = \gamma$. This sets the interpretation of all asymptotic expressions as the directional limit $n\to\infty, p=n/\gamma$. For example, a sequence of random variables $W_n = o_P(1)$ means $W_n$ converges to zero in probability under the asymptotic setting above. 
 \par
\textbf{RERM training:} The goal of a learning procedure $A$ is to estimate the parameter $\bbeta^*$ from training data. A widely used approach is \textit{regularized empirical risk minimization} (R-ERM), defined as:
\begin{equation} \label{eq:betahat}
\hat{{\bbeta}} =
A(\mathcal{D}_n) :=  \underset{{{\bbeta}} \in \mathbb{R}^p}{\arg\min} \quad  L({\bbeta}) := \underset{{{\bbeta}} \in \mathbb{R}^p}{\arg\min }\quad
\lambda r({\bbeta}) +\sum_{i=1}^n  \ell ( y_i \mid \bx_i^\top {\bbeta} )  ,
\end{equation}
where $\ell$ is a (individual) loss function and $r$ is a regularization term with factor $\lambda$, which we assume is $O(1)$ throughout.  A statistically interpretable choice of the individual loss $\ell(y| z)$ is the negative log likelihood $ -\log q(y|z)$, and the choice of the regularizer $r$ is often based on some prior information on $\bbeta^*$. Popular choices include ridge ($r=\|\cdot\|_2^2$), LASSO ($r=\|\cdot\|_1$), and their variants.
\par
\textbf{Retraining from scratch:} The objective of unlearning is to remove the influence of a subset $\mathcal{M}$ of observations, of size $|\mathcal{M}| = m$, that have requested removal from the training data.
The \textit{ideal unlearned estimator} would be the one retrained from scratch on the remaining data $\cD_{\setminus \cM}:= \cD_n \setminus \cD_{\cM}$:
\begin{equation} \label{eq:betahat-relearn}
      \hbm:=
         A(\cD_{\setminus \cM}) :=  \underset{{\bbeta} \in \RR^p}{\argmin }   \quad 
        L_{\setminus \cM}(\bbeta):= \underset{{\boldsymbol{\beta}} \in \mathbb{R}^p}{\arg\min}  \quad  \lambda r({\bbeta}) + \sum_{i\notin \cM}  \ell ( y_i|\bx_i^\top {\bbeta} ). 
    \end{equation}
\textbf{The approach of machine unlearning:} To process deletions efficiently, without full retraining, and obscure residual information, we update $\hat{\bbeta}$ via the randomized procedure below. \footnote{For more details on the claim that randomization is \textit{necessary} to hide residual information and preserve privacy of individuals, please refer to the literature of differential privacy \cite{dwork2006differential}.} 
\begin{equation}\label{eq:betatilde-unlearn}
    \tilde{\bbeta}_{\setminus\cM}
    :=\bar{A}(\hat{\bbeta},\cD_{\cM},T(\cD_n),\bb),  \text{ where } \bb \text{ is a random noise.}
\end{equation}
\textbf{Assumptions on $\bar{A}$ and motivations behind:} Given a trained model, $\hat{\bbeta}$, note that $\bb$ is the only source of randomization during unlearning. In addition to the perturbation $\bb$, the unlearning procedure $\bar{A}$ has access to the set $\cD_{\cM}$ corresponding to the data to be removed, and limited access to the outputs of the learning algorithm $A$ on $\cD_n$.  For example, (i) the original estimator $\hat{\bbeta}$, and (ii) some auxiliary information $T(\cD_n)$ such as \textit{gradient} or \textit{Hessian} of the objective function on $\cD_n$, at $\hat{\bbeta}$.

\vspace{-.1 cm}
\section{Formal definitions of certifiability and accuracy}\label{sec:formal:def}
\vspace{-.1 cm}

As discussed in Section \ref{sec:machunle}, machine unlearning algorithms are expected to satisfy two key criteria: (1) \textit{protect user privacy} and (2) \textit{preserve model accuracy}. The purpose of this section is to formalize the notions corresponding to these objectives by introducing:
(1) $(\phi, \varepsilon)$-Gaussian certifiability as a canonical measure of privacy, and
(2) Generalized error divergence as a measure of accuracy.
We will formally establish later that both notions are particularly well suited to high-dimensions.

\vspace{-.1 cm}
\subsection{General certifiability through the lens of hypothesis testing}
\vspace{-.1 cm}

 \textbf{A Hypothesis testing motivation:}\label{ssec:hypothesistesting} To motivate this, we take the perspective of an adversary who, given an output $\check{\bbeta}$ from an unlearning algorithm $\bar{A}$, performs the following hypothesis testing problem:
\begin{equation}\label{eq:unlearn-testing}
 H_0: \check{\bbeta}\sim \cP_{\rm re}
 \quad
 \text{vs.}
 \quad
 H_1: \check{\bbeta}\sim \cP_{\rm un}
\end{equation}
where $\cP_{\rm re}$ and $\cP_{\rm un}$ are the conditional distributions of $\bar{A}(\hbm,\emptyset,T(\cD_{\setminus\cM}),\bb)$ given the data $\cD$ and $\bar{A}(\hat{\bbeta},\cD_{\cM},T(\cD),\bb)$ given  the data $\cD$ respectively. Indeed, if the adversary has sufficient evidence to reject $H_0$, it would mean that $\check{\bbeta}$ has retained some information about $\cD_{\cM}$, defeating the purpose of machine unlearning. Note that in all settings, the adversary may make mistakes in accepting or rejecting the null. Hence, to consistently evaluate the quality of the adversary's decision, we have to evaluate the probability of Type I error (i.e. the error of rejecting $H_0$ when the truth is $H_0$), and Type II error (i.e. the error of not rejecting $H_0$ when the truth is $H_1$).
\par
\textbf{Trade-off functions (the key object):} The above  leads us to the notion of trade-off curve defined below and studied extensively in the Statistics and information theory  \cite{lehmann1986testing}, \cite{poor_1994_signal}. To introduce this notion, let us simplify the notations of the above problem to the following. Suppose that a random vector $W$ is given, and we would like to perform the following test:
\[
H_0: W \sim P \quad vs. \quad H_1:W \sim Q. 
\]

\begin{definition}   \label{def:TOF} (Trade-off function) 
Given two probability distributions $P, Q$ on a measurable space $(\cW, \mathcal{F}_{\cW})$, we define the  \emph{trade-off function} as the map $T(P,Q):[0,1]\to[0,1]$   as
\begin{equation}\label{eq:TOF}
 T(P,Q)(\alpha):=
  \inf_{\phi}\Bigl\{
        \beta_{\vp} := \e_{Q}[1-\vp] 
        \,\Bigm\vert\,
        \alpha_{\vp}:= \e_{P}[\vp]\le \alpha,\;
        \vp:\cW\!\to[0,1]\text{ measurable}
      \Bigr\}.
\end{equation}
\textbf{Intuition:} In words, for any given type I error $\alpha$, the trade-off function returns the smallest possible value of type II error $\beta_{\vp}$ over all possible test functions $\vp\colon \cW\to[0,1]$. 
\end{definition}
 
\textbf{From trade-off functions to General certifiability:} Using the notion of trade-off functions, and inspired by the notion of $f$-differential privacy introduced in \cite{dong2022gaussian}, we aim to define a notion of certifiability that generalizes and unifies the existing techniques. Intuitively speaking, once we decide on a trade-off function $f: [0,1] \rightarrow [0,1]$, we can certify $\bar{A}$ if  
\begin{equation}
  T(\cP_{\rm re},\cP_{\rm un})(\alpha)\ge f(\alpha)
  \quad
  \text{for all }\alpha\in [0,1]
\end{equation}
To make our certifiability criteria less restrictive, we require the above definition to be satisfied with high probability over datasets $\cD$. This leads us to the following definition of general \textit{$f-$ certifiability}.
\begin{definition}[$f$-certifiability] \label{def:f-Unlearn} 
    Given $\phi > 0, m \in [n]$ and $\cP_{\rm un}, \cP_{\rm re}$ as define in \eqref{eq:unlearn-testing}, and a trade-off curve $f: [0,1]\rightarrow [0,1]$, we say unlearning algorithm $\bar{A}$ satisfies $\phi$-certifiable with respect to $f$ if
    \begin{equation}  \label{eq:f-PAR}
      \p \left[ \inf_{|\cM| \leq  m}   
      \min(
      T(\cP_{\rm re}, \cP_{\rm un})(\alpha), T(\cP_{\rm un}, \cP_{\rm re})(\alpha))  \geq f(\alpha) \quad
      \text{for all }
      \alpha\in [0,1]
      \right] \geq  1-\phi
    \end{equation}
    where the probability $\p$  is solely over the randomness of the data $\cD$.
\end{definition}

\textbf{Different choices of $f$ yields different notions of certifiability:} Observe that by choosing different trade-off functions, we can cover a wide range of notions of privacy in the literature \cite{pmlr-v119-guo20c}, \cite{sekhari2021remember}, \cite{zou2025certified}. For instance, by setting $f(\alpha)$ in the following way:  
 \begin{equation} \label{eq:e-d DP}
   f_{\ve, \delta}(\alpha):=  \max\{0, 1-\delta - e^{\ve}\alpha, e^{-\ve} (1-\delta- \alpha)\}  
    \end{equation}
we will obtain a notion similar to $(\ve\text{-}\delta)$-certifiability condition discussed in \cite{sekhari2021remember}.

\vspace{-.1 cm}
\subsection{(\texorpdfstring{$\phi$}{phi},\texorpdfstring{$\ve$}{ve})-Gaussian  certfiability- a canonical choice in high dimensions } \label{sec:Gauss_cert}
\vspace{-.1 cm} 

\textbf{Motivation:} We study machine unlearning in high dimensions, where choosing an appropriate $f$ is the key to formalizing certifiability. In this regime, a natural (indeed canonical) choice of 
$f$ emerges from two facts. 1) In practice, privacy is achieved by adding noise, and 2) an intrinsically high-dimensional phenomenon implies that, \textit{with high probability (up to a tolerance $\phi$), most  $1$d projections of an isotropic high-dimensional (log-concave) distribution are approximately Gaussian}\footnote{Think of all the $1$d projections of the uniform distribution on a high dimensional sphere as an example.}  \cite{diaconis1984asymptotics}, \cite{klartag2007cltconvex}. A precise instantiation of this idea appears in the differential privacy literature \cite{dong2021central}, which motivates us to consider the following choice of the trade-off curve $f_{G,\varepsilon}$ in Definition \ref{def:f-Unlearn} in comparison to previously proposed notions such as $f_{\varepsilon, \delta}$.
\begin{equation} \label{eq:GTOF}
  f_{ G,\ve}(\alpha):= T(N(0,1), N(\ve, 1))(\alpha)= \Phi(\Phi^{-1}(1-\alpha) - \ve)
\end{equation}
where $\Phi(\cdot)$ is the cdf of a standard Gaussian variable $\Phi(t)= \mathbb{P}[N(0,1)\leq t]$ for all $t\in \mathbb{R}$.

$f_{G, \varepsilon}(\alpha)$ possesses several attractive properties. The following lemma lists one such property of Gaussian trade-off functions known as \textit{dimension-freeness}. This will be used later in this paper.
\begin{lemma}[Dimension freeness] 
\label{lem:GTOF1} For any  $\bm{\mu}_1, \bm{\mu}_2 \in \R^p$ and $\sigma >0$ let $\ve := \frac{1}{\sigma}\|\bm{\mu}_1- \bm{\mu}_2\|_2$. Then 
\begin{equation} \label{eq:dto1}
    T(\bm{\mu}_1+ \sigma N(0, \II_p), \bm{\mu}_2 + \sigma N(0, \II_p)) \equiv T(N(0,1), N(\ve, 1)).
\end{equation}
 \end{lemma}

We call the criterion of  Definition \ref{def:f-Unlearn}, with $f_{G,\ve}(\cdot)$ replacing $f(\cdot)$  as \textbf{$(\phi,\varepsilon)$-GPAR}, to abbreviate  $(\phi,\varepsilon)$-Gaussian Probabilistically certified Approximate data Removal or $(\phi, \varepsilon)$-Gaussian certifiability. We enlist some more properties of $(\phi,\varepsilon)$-GPAR that make it emerging in high-dimensional settings:
\par
\textbf{Universality of Gaussian noise:} \cite{dong2021central} shows that a broad class of isotropic log-concave noise-adding mechanisms (to achieve privacy) converge in behavior to $(\phi,\varepsilon)$-Gaussian certifiability as $p\uparrow \infty$. This makes the Gaussian mechanism the natural noise-adding mechanism in high dimensions.
\par
\textbf{Gaussian certifiability as the tightest notion:}  Now, once we use the canonical Gaussian noise for randomizing the estimates, a consequence of \textit{Blackwell ordering} says that the Gaussian trade-off curve \eqref{eq:GTOF} is the tightest way to capture the Gaussian mechanism \cite{dong2022gaussian}. 
\par
\textbf{Suboptimality of all other notions of certifiability:} All the previously proposed notions of unlearning, including $\epsilon-\delta$ variations of \cite{pmlr-v119-guo20c}, \cite{sekhari2021remember}, \cite{zou2024theoretical}, Rényi certifiability of \cite{allouah2025the}, are not tightly achievable under the Gaussian mechanism except for Gaussian certifiability  \cite{dong2022gaussian}. This establishes the natural emergence of the $\varepsilon$-GPAR framework with tolerance $\phi$ as superior over any other framework in high dimensions. 
\par
\textbf{Appearance in practice, in related experiments:}
In the context of machine unlearning, Gaussian trade-off curves \eqref{eq:GTOF} (the core of Gaussian certifiability) have been experimentally shown to match empirical trade-off curves remarkably well for the removal of poisoned data \cite{pawelczyk2025machine}.

In Section \ref{sec:theoreticaguarantees}, we will use the $(\phi, \varepsilon)$-GPAR framework to theoretically evaluate the user privacy protection capabilities of machine unlearning algorithms that are based on the Newton method. 

 \vspace{-.1 cm}
\subsection{Generalization Error Divergence: a measure of accuracy}
 \vspace{-.1 cm}
 
In the previous subsection, we introduced a canonical notion of certifiability for unlearning. We now turn to the second criterion, which aims to evaluate model accuracy. Without such a criterion, an unlearning algorithm could output pure noise, achieving perfect user privacy at the cost of severely degraded model performance. To formally evaluate the generalization capabilities of our unlearning procedure $\bar{A}$, we consider the metric Generalization error divergence (GED).

\begin{definition}\label{def: GED} 
    Given a dataset $\cD_n$ of $n$ i.i.d. samples $(\bx_i,y_i)$ as introduced in Section~\ref{sec:erm-intro}, let $(\bx_0,y_0)$ be a new sample i.i.d. with the observations in $\cD_n$. Let $\ell(y|\bx^\top\bbeta)$ be a measure of error between $y$ and $\bx^\top\bbeta$. Then the \textbf{Generalization Error Divergence (GED)}  of the learning-unlearning pair $(A, \bar{A})$ on the dataset $\cD_n$ with a subset $\cM \subset [n]$ of data removal requests is defined as:
    \begin{equation} \label{eq:GED}
        {\rm GED}_{\ell}(A,\bar{A}; \cM, \cD_n):=
        \e\left(
        \left[
        \big\vert
        \ell(y_0|\bx_0^\top A(\cD_{\setminus\cM})) - \ell(y_0|\bx_0^\top \bar{A}(A(\cD_n),\cD_{\cM},T(\cD_n),\bb)) 
        \big\vert\right]
        | \cD_n\right),
    \end{equation}
    where we condition on the randomness of the data set $\cD_n$ and average over the randomness of the unlearning algorithm $\bar{A}$, as well as that of the test data point $(\bx_0, y_0)$.
\end{definition}   

\textbf{Intuition behind:} The above metric compares the generalization error of the unlearning algorithm $\bar{A}$ given $\hat{\bbeta}$ and data removal requests $\cM$ with that of the model $A$ trained from scratch on $\cD_{\setminus \cM}$, on  a fresh sample $(\bx_0, y_0)$. We note that previously \cite{sekhari2021remember} used the notion of excess risk of $\bar{A}$ against that of the true minimizer of population risk as a measure of accuracy of machine unlearning algorithms. But, as we will show in Section ~\ref{sec:comparison} of appendix that this notion does not behave well under the proportional high-dimensional setting. We also refer to \cite{zou2025certified} for more.

\vspace{-.1 cm}
\section{Newton-based machine unlearning: theoretical guarantees}\label{sec:theoreticaguarantees}
\vspace{-.1 cm}

Having settled down on both the evaluation criteria, we study the performance of unlearning procedure $\bar{A}$  based on one Newton step and perturbing the result with Gaussian noise to achieve $(\phi,\varepsilon)$-GPAR.

\vspace{-.1 cm}
\subsection{Newton-based unlearning mechanism}
\vspace{-.1 cm}

Newton method is an iterative method for solving optimization problems based on a second-order Taylor approximation of the given optimizing function \cite{boyd2004convex}. 
 \begin{definition} (Newton-Raphson method)
     \label{def: Newtona}
    Suppose that $h:\R^p\to\R^p$ is a differentiable function with  an invertible Jacobian matrix $G$ anywhere on $\R^p$ and that $h$ has a root $h(\bbeta^*)=0$. Starting from an initial point $\bbeta^{(0)}\in \R^p$, the Newton method is the following iterative procedure: for step $t\geq 1$,
    \begin{equation} \label{eq:Newton}
      \bbeta^{(t)} := \bbeta^{(t-1)} - \bG^{-1}(\bbeta^{(t-1)})h(\bbeta^{(t-1)}).
    \end{equation}
 \end{definition}
Using the Newton method \cite{sekhari2021remember} suggested the following unlearning algorithm:
\textbf{Approximation step :} starting from $\hat{\bbeta}$, we run one  Newton step with $h= \nabla L_{\setminus \cM}$ to obtain:
\begin{equation} \label{eq:Newton_step}
   \hat{\bbeta}^{(1)}_{\setminus\cM} = \hat{\bbeta} - \bG(L_{\setminus \cM})^{-1}(\hat{\bbeta}) \nabla L_{\setminus \cM}(\hat{\bbeta}), \text{ where } \bG(L_{\setminus \cM}) \text{ is the Hessian of } L_{\setminus \cM} \text{  in }\eqref{eq:betahat-relearn}.  
\end{equation}
\textbf{Randomization step:} add a Gaussian noise $\bb \sim  N(\mathbf{0},\sigma^2\II_p)$ to  $\hat{\bbeta}^{(1)}_{\setminus\cM} $ to have the unlearned output
\begin{equation}\label{eq:randomization_step}
    \textbf{Noisy one step Newton output: } \bar{A}(A(\cD_n),\cD_{\cM},T(\cD_n),\bb))= \tilde{\bbeta}_{\setminus\cM} : = \hat{\bbeta}^{(1)}_{\setminus\cM} + \bb.  
\end{equation}
\textbf{Intuition and the goal:} First, the choice of $\sigma$ will be discussed in our theoretical results.  Next, note that since $\hat{\boldsymbol{\beta}}^{(1)}_{\setminus\mathcal{M}}$ differs from $\hat{\boldsymbol{\beta}}_{\setminus\mathcal{M}}$, the difference between the two vectors may reveal information about the data to be removed, $\mathcal{D}_{\mathcal{M}}$. Hence,  a standard practice is to obscure it by adding random noise $\bb$. Finally, the aim in the remaining part of this section is to answer the following two questions:
\par
\textbf{Privacy:} How to choose  $\sigma$ to ensure that $\tilde{\bbeta}_{\setminus\cM}$ satisfies $(\phi, \varepsilon)$-GPAR  when $n,p$ large with $n/p \to \gamma$?

\par
\textbf{Accuracy:} Can we calculate the generalization error divergence of $\tilde{\bbeta}_{\setminus\cM}$ given the choice of $\sigma$ obtained above? Does the GED go to zero as $n,p \rightarrow \infty$ while $n/p \rightarrow \gamma$? 

\vspace{-.1cm}
\subsection{Technical assumptions on the loss functions and the data}
\vspace{-.1cm}

\textbf{Loss functions:} We describe the assumptions on the unregularized individual loss $l$ and the regularizer $r$ \eqref{eq:betahat}. To ensure that the RERM introduced in \eqref{eq:betahat} has a unique solution (from strong convexity arguments) and Newton unlearning makes sense theoretically, the following assumption are made:
\par
\namedlabel{assum:separability}{(A1)} \textbf{Separability of $r$:}
     $r:\mathbb{R}^p \to \mathbb{R}_{+}$ is separable. More precisely, $r(\bbeta) = \sum_{k\in[p]}r_k(\beta_k)$.
    \par
   \namedlabel{assum:smoothness}{(A2)} \textbf{Smoothness of $(l,r)$:}
   $\ell:\RR\times\RR\to\RR_+$ and $r:\RR^p\to\RR_+$ are both thrice differentiable. 
    \par
    \namedlabel{assum:convexity}{(A3)} \textbf{Convexity of $(l,r)$:}  $\ell$ and $r$ are proper convex, and $r$ is $\nu$-strongly convex  for  $\nu=\Theta(1)>0$.
    \par 
    \namedlabel{assum:poly}{(A4)} \textbf{Polynomial growth:} $\exists$ constants $C,s>0$ such that for all  $y,z$ (denote $\ld(y,z) := \partial_y \ell(y,z)$)
    \begin{equation}
        \max\{\ell(y,z), |\ld(y,z)|,|\lddd(y,z)| \}\leq C(1+|y|^s + |z|^s)  \text{ and  } \nabla^2 r(\bbeta) = \diag[\rdd_k(\beta_k)]_{k\in[p]} 
    \end{equation}
    is $C_{rr}(n)$-Lipschitz (in Frobenius norm) in $\bbeta$ for some $C_{rr}(n)=O(\polylog(n))$.
    \par
\textbf{Relaxed assumptions on $(\ell, r)$ and the ridge example:} Note these assumptions on $(\ell, r)$ are  from \cite{zou2025certified}, and they weaken many of the standard optimization assumptions of $\Omega(1)$ strong convexity and $O(1)$ smoothness of per example loss $f$ in \citet{allouah2025the} or variations thereof in \citet{sekhari2021remember}. Convexity of $\ell$ and $\nu$-Strong convexity of $r$ above  actually means that per-example loss $\bbeta \to f(\bbeta,\mathbf{z})= l(y, x^T\bbeta) + \frac{\lambda}{n} r(\bbeta)$ is $\frac{\lambda \nu}{n}$ strongly convex, which is significant relaxation of  $\Omega(1)$ strong convexity of $f$, capturing the concrete case of the ridge regularized $r(\bbeta)= \|\bbeta\|^2$ quadratic loss $l(y,z)= (y-z)^2$ .  \citet{allouah2025the} and \citet{sekhari2021remember} both fail to capture the simplest example of high dimensional statistics in their respective frameworks.

\textbf{Data:} We require the following on the data that generates our features $\bx_i$ and responses $y_i$.
\par 
\namedlabel{assum:x}{(B1)} \textbf{Subgaussian features:} $\bm{x}_i$ are mean zero sub-Gaussian vectors with covariance $\bm{\Sigma}$ denoted as  $\bm{x}_i\overset{iid}{\sim} \mathcal{SG} (\bzero, \bm{\Sigma})$. We further assume that  $ \lambda_{\max}(\bSigma) \leq \frac{C_X}{p}$, for some constant $C_X>0$.  \label{featurescaling}
\par
\namedlabel{assum:y}{(B2)} \textbf{Sub-polylogarithmic responses:} 
   We assume that $\PP(|y_i|>C_y(n))\leq q_y(n)$ and $\EE|y_i|^{2s}\le C_{y,s}$ for some $C_y(n)=O(\polylog(n))$, a constant $C_{y,s}$ and $q_n^{(y)}=o(n^{-1})$.

\textbf{Intuition behind features and responses:} Assumption~\ref{assum:x} on mean-zero sub-Gaussian features is a mild and frequently made in many papers on high dimensional statistics. See for example \cite{miolane2021distribution,rad2018scalable,auddy24a, zheng2017does, donoho2016high, bellec2025corrected}. Assumption ~\ref{assum:y} is not a stringent assumption either, and most popular models such as logistic regression, Poisson regression, and linear regression satisfy this assumption. Using these assumptions, we can finally prove our main theoretical results. 


 \vspace{-.1cm}
\subsection{Theoretical Guarantees: ensuring user privacy as well as model accuracy}
\vspace{-.1cm}

\textbf{One step Newton achieves $(\phi, \varepsilon)$-GPAR:} Our first result shows that for a suitable noise variance, the randomized one-step Newton unlearned estimator defined in \eqref{eq:randomization_step} satisfies $(\phi,\ve)$-GPAR. 

\begin{theorem}\label{thm:Cert}
   Suppose Assumptions~\ref{assum:separability}-\ref{assum:poly} as well as \ref{assum:x}-\ref{assum:y} hold. Then there exist poly-logarithmic  $C_1(n),C_2(n)=O(\polylog(n))$ for which the randomized one-step Newton unlearning \eqref{eq:randomization_step} when used with a perturbation vector $\bb\sim N\left(\mathbf{0}, \frac{r^2}{\ve^2}\II_p\right)$, achieves $(\phi_n,\ve)$-GPAR with 
    \begin{equation} \label{eq:phi_n}
    r = C_1(n) \sqrt{\frac{C_2(n)m^3}{2\lambda \nu n}},
    \quad   \phi_n = nq_n^{(y)} + 8n^{-3} + ne^{-p/2} + 2e^{-p}\to 0.
    \end{equation}
\end{theorem}

The proof of this theorem is presented in Sections ~\ref{sec:Proof_G}, ~\ref{sec:Proof_SG} of the appendix.  After the above theoretical guarantee ensuring user privacy, we now calculate the GED of this machine unlearning algorithm. 

\begin{theorem}\label{thm:GED}
   Suppose Assumptions~\ref{assum:separability}-\ref{assum:poly} as well as \ref{assum:x}-\ref{assum:y} hold. Consider the unlearning estimator defined in \eqref{eq:randomization_step} with the noise variance set according to Theorem \ref{thm:Cert}. Then, with probability at least $1-(n+1)q_n^{(y)}-14n^{-3}-ne^{-p/2}-2e^{-p}-e^{-(1-\log(2))p}$, 
    \begin{align*}
        {\rm GED}(\tbm,\hbm)
        &\leq C_1(n)\sqrt{C_2(n)}\left(\frac{1}{\ve}+\frac{1}{\sqrt{p}}\right)\sqrt{\frac{m^3(m+2)}{\lambda \nu n}}\cdot\polylog(n).
    \end{align*}
\end{theorem}
\textbf{The final technical conclusion of the paper:} Combining the above two theorems we reach the conclusion: If we set the variance of the Gaussian noise as suggested by Theorem \ref{thm:Cert}, the machine unlearning algorithm $\tbm$ that is based on \eqref{eq:randomization_step} offers $(\phi_n, \varepsilon)$-GPAR as well as vanishing GED
\begin{equation}
{\rm GED}(\tbm,\hbm) = O_p\left(\frac{m^2 \polylog (n)}{\sqrt{n}}\right) \text{ as soon as }  \frac{m^2 \polylog (n)}{\sqrt{n}} \to 0
\end{equation}
In particular, if $m=o(n^{\frac14-\alpha})$ for arbitrary $\alpha>0$, we have  ${\rm GED}(\tbm,\hbm)=o_p(1)$. Note that both theorems are valid in high-dimensional settings where $n,p \rightarrow \infty$, while $n/p \to  \gamma$.

\vspace{-.2cm}
\section{Experiments validating our theoretical contributions} \label{sec:experiments}
\vspace{-.2cm}

\textbf{Overview:} Given our theoretical focus, and the abundance of empirical studies on Newton-based unlearning \cite{pmlr-v119-guo20c}, \cite{bui2024newtonsmethodunlearnneural}, we present simulations solely to  validate our results and to compare with the results of \cite{zou2025certified}. The incompatibility  of  $\ve$ certifiability  with noise adding mechanisms requires  disproportionately large amount of isotropic Laplace noise in \cite{zou2025certified}, whereas $\ve$-GPAR criterion optimally captures the  Gaussian mechanism. Our simulations show that it leads to substantial improvement of the unlearned estimator across different  settings.

\textbf{Logistic regression model:} For simplicity, we use ridge penalized $r(\bbeta)=\|\bbeta\|^2$ logistic loss  $\ell(y|\bx^{\top}\bbeta)$,  with random features  $\bx_i\sim N(\mathbf{0},\frac{1}{n}\II_p)$ and true parameter $  \bbeta_* \sim N(\mathbf{0},\II_p)$.  To depict high-dimensionality, we take $n=p$, penalty parameter $\lambda=0.5$, and show behavior of the error metrics by varying the values of $n,p,m,$ and certifiability parameter $\ve$. 1) When $m=1$, we look at all possible removals of size one justifying robustness of our certifiability. 2) When $m>1$, instead of all possible removals, we study unlearning across 1000 randomly chosen subsets of size $m$.  More  details and codes are  in the appendix and at \url{https://anonymous.4open.science/r/unlearning-E14D}.
\begin{figure}[htbp]
    \centering
    \includegraphics[width=0.32\linewidth]{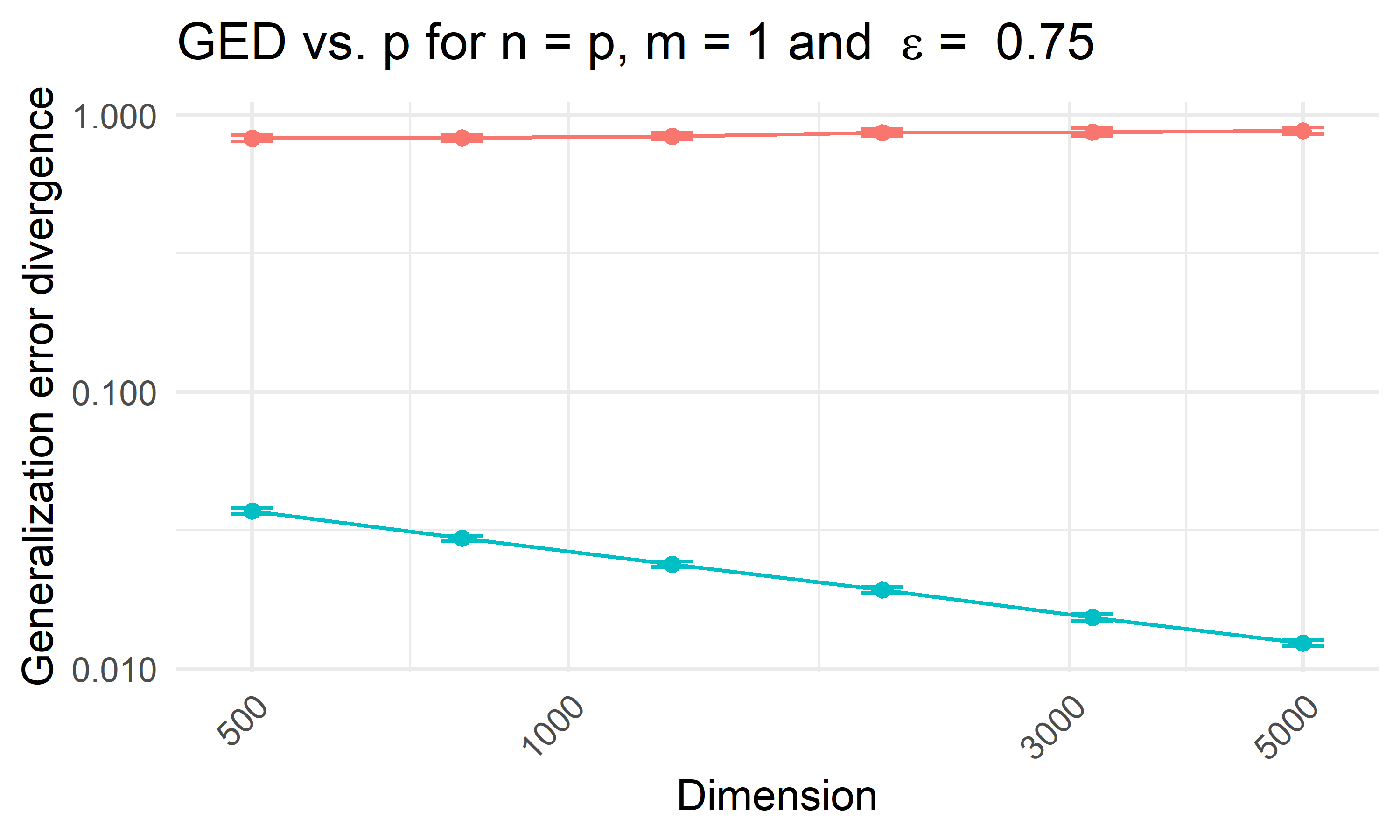}
    \includegraphics[width=0.32\linewidth]{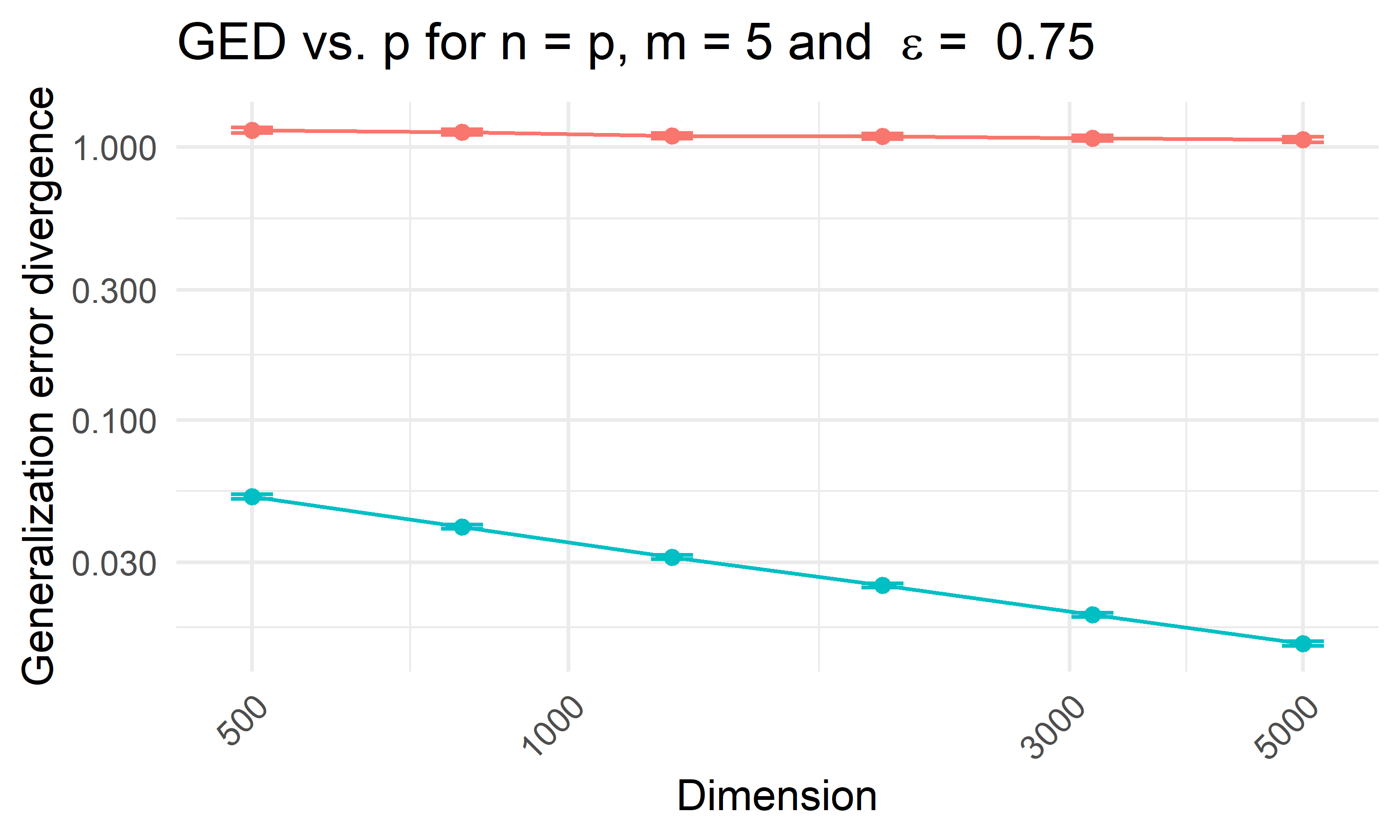}
    \includegraphics[width=0.32\linewidth]{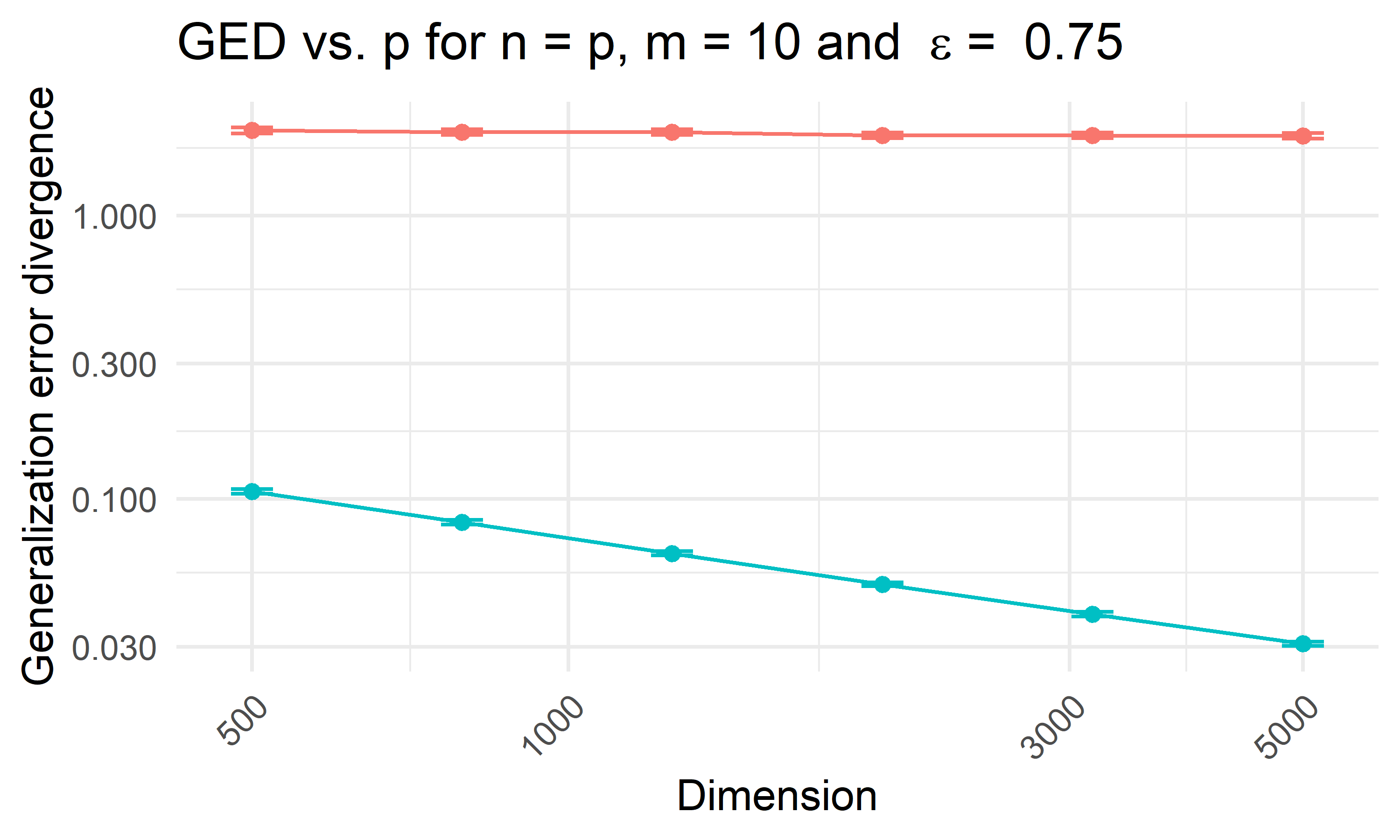}
    \caption{Comparison of unlearned estimators on new test data: mean GED (with 3 SD error bars) across the dimension $p$ (both in $\log$ scale) for Laplace (in red) vs. Gaussian (in cyan). We set $\lambda=0.5$.}
    \label{fig:ged-across-p}
\end{figure}
\begin{figure}[htbp]
   \centering
    \includegraphics[width=0.32\linewidth]{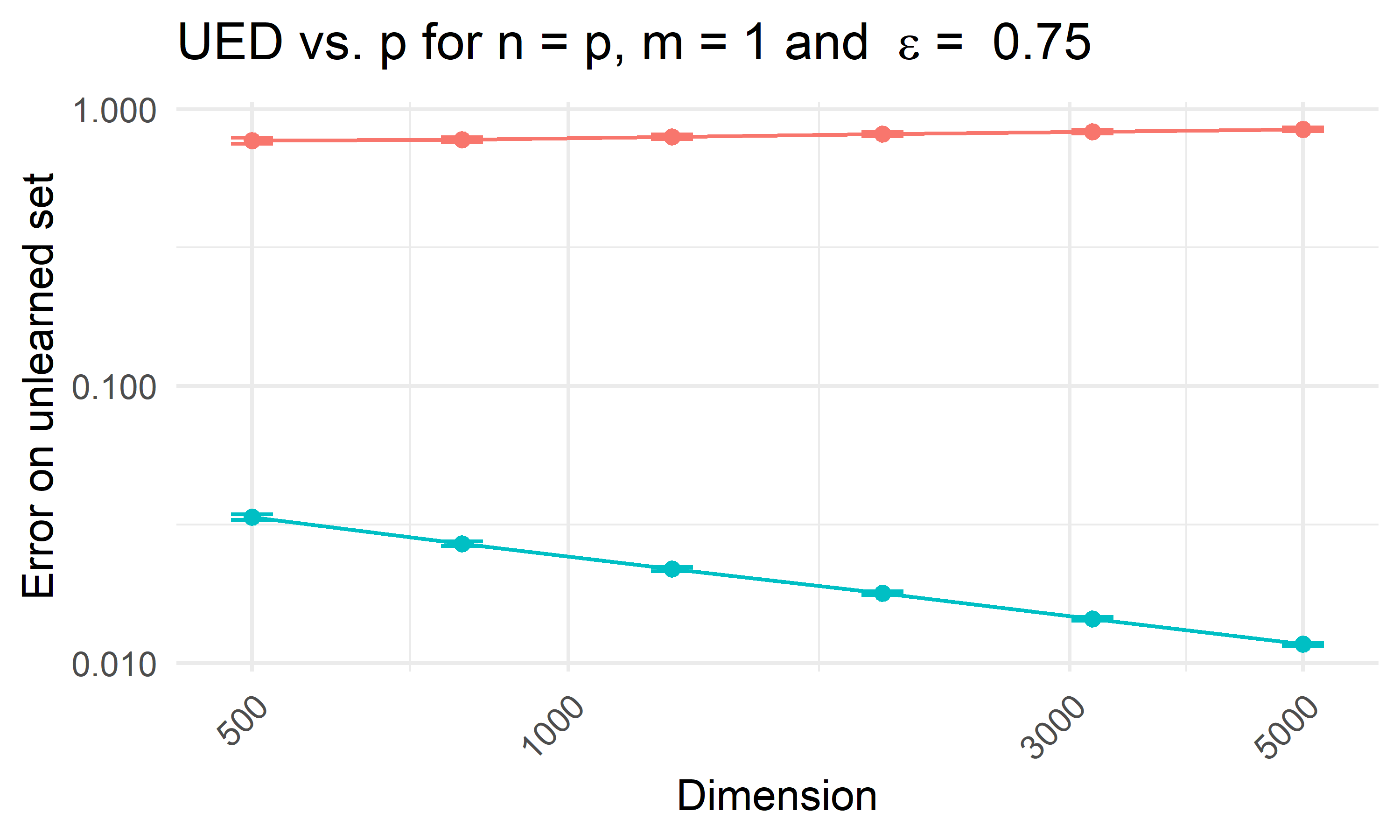}
    \includegraphics[width=0.32\linewidth]{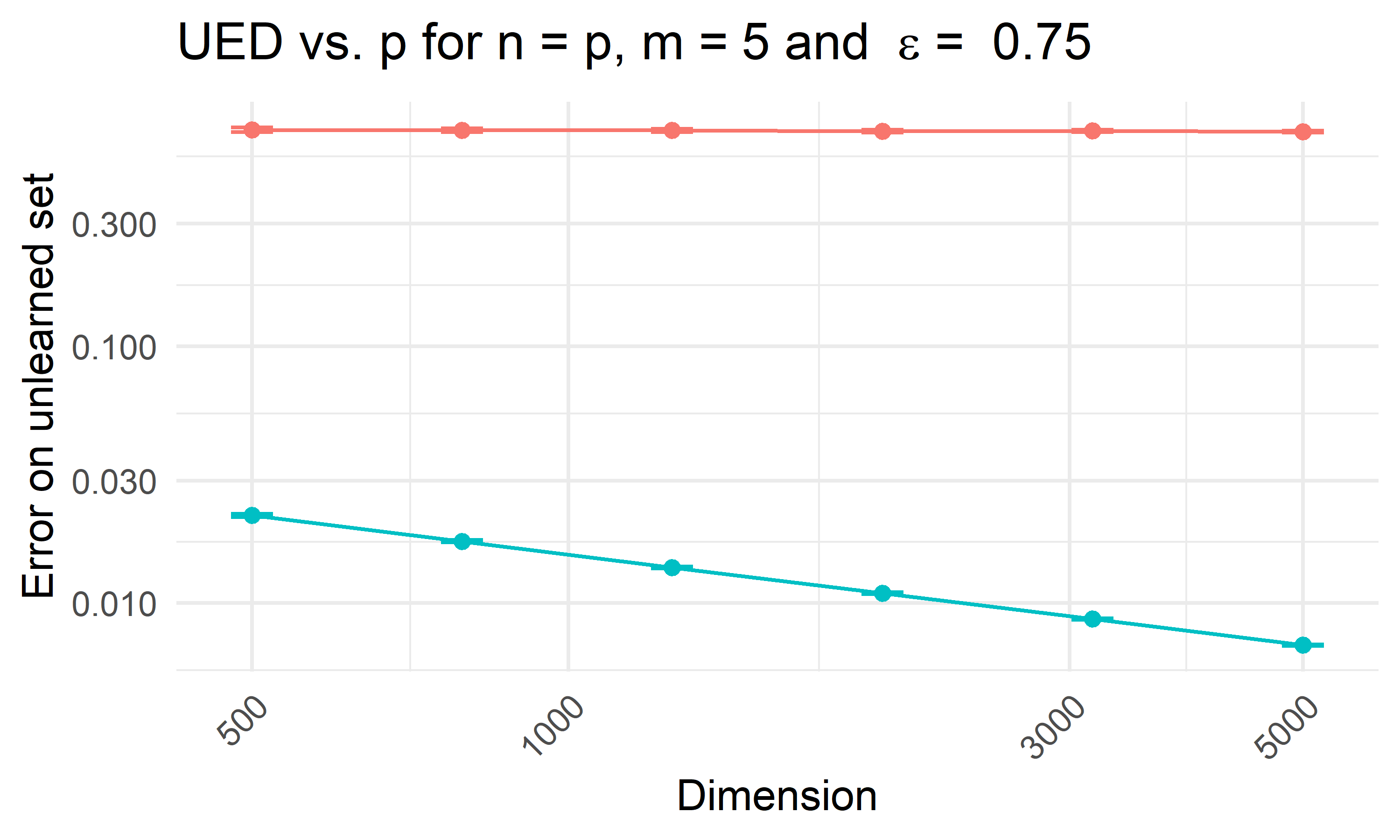}
    \includegraphics[width=0.32\linewidth]{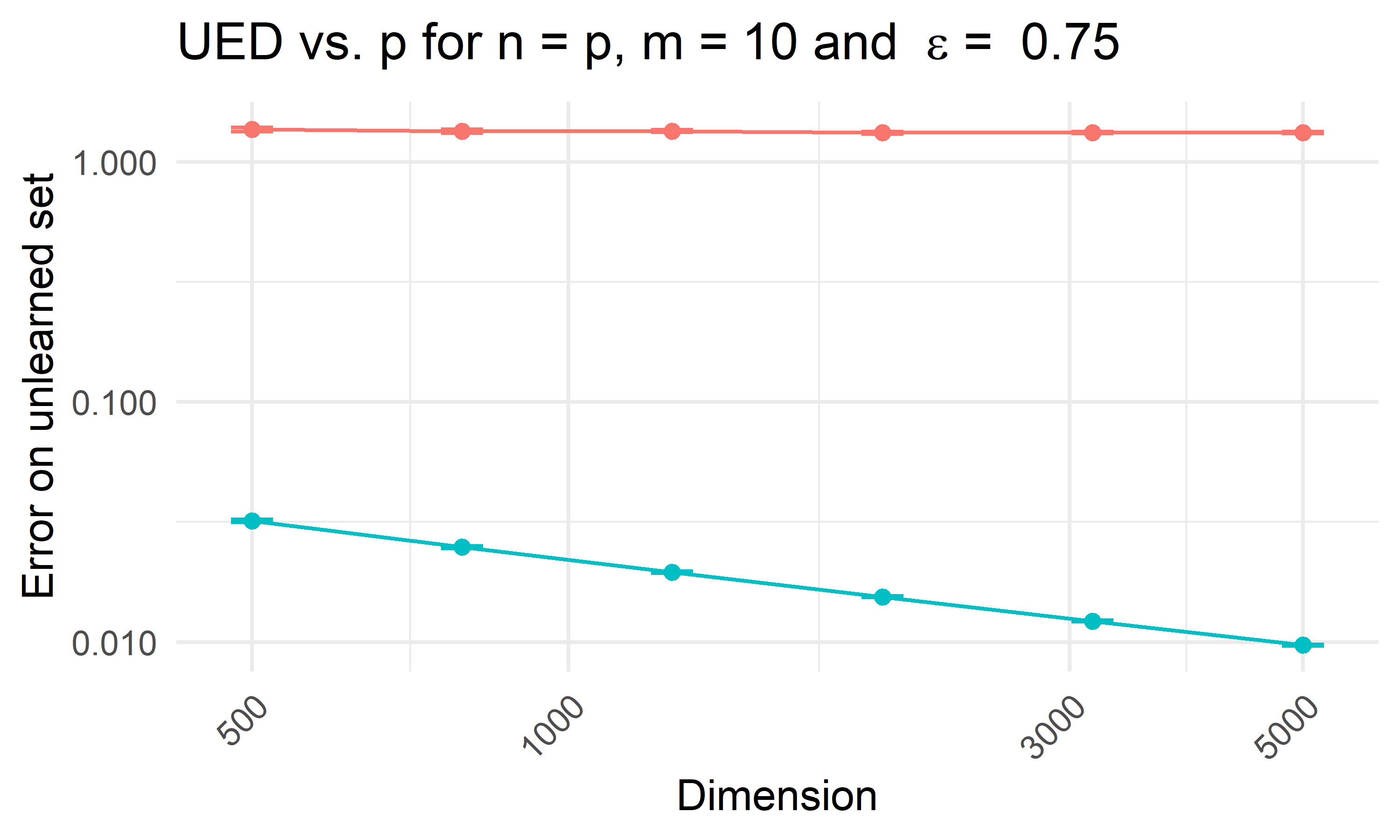}
    \caption{
    Comparison on unlearned data: mean UED (with 3 SD error bars) on the unlearned set across $p$ (both in $\log$ scale) for Laplace  (in red) vs. Gaussian noise (in cyan). We set $\lambda=0.5$.}
    \label{fig:forget-across-p}
\end{figure}
\par
\vspace{-.2 cm}
\textbf{First set (meaning of figure 1):} We fix $\ve=0.75$, and examine via GED \eqref{eq:GED} the generalization performance of unlearning on 100 previously unseen test data. We vary $n=p$ across 6 values from 500 to 5000 equally spaced on a logarithmic scale. Figure~\ref{fig:ged-across-p} plots the average GED values (across 200 replications) against dimension $p$, transforming both X and Y axes to be on a logarithmic scale. The figures are repeated for unlearning sizes $m=1$ (left), $m=5$ (middle), and $m=10$ (right).
\par
\textbf{Conclusions from figure 1:} Observe for $m=1,5,10$, as $p \uparrow$, GED for Laplace perturbed estimator of \cite{zou2025certified} is unchanged, whereas our Gaussian perturbed estimator decreases steadily. A linear regression of $\log({\rm mean(GED)})$ vs. $\log(p)$ confirms this. The lines for Laplace unlearning had slopes of 0.03, -0.03, and -0.01 for $m=1,5,10$ respectively,  implying one step Laplace-Newton estimator has a non-vanishing GED as $p\uparrow$. This shows the sub-optimality of $\varepsilon$-certifiability  by \emph{requiring at least two Newton steps} to reach vanishing GED. In contrast, the Gaussian unlearning lines had slopes of -0.47,-0.54, -0.51, thus GED decaying at an order of  $p^{-0.5}$ supporting  Theorem~\ref{thm:GED}.
\par
\textbf{Figure 2 and conclusions from it:} In figure~\ref{fig:forget-across-p} we observe the same behavior as in figure ~\ref{fig:ged-across-p}  when comparing the performance on the unlearned dataset, where we compute the unlearned error divergence (UED) between the retrained and unlearned estimators (see appendix for more on UED). 
\par
\begin{figure}[htbp]
    \centering
    \includegraphics[width=0.32\linewidth]{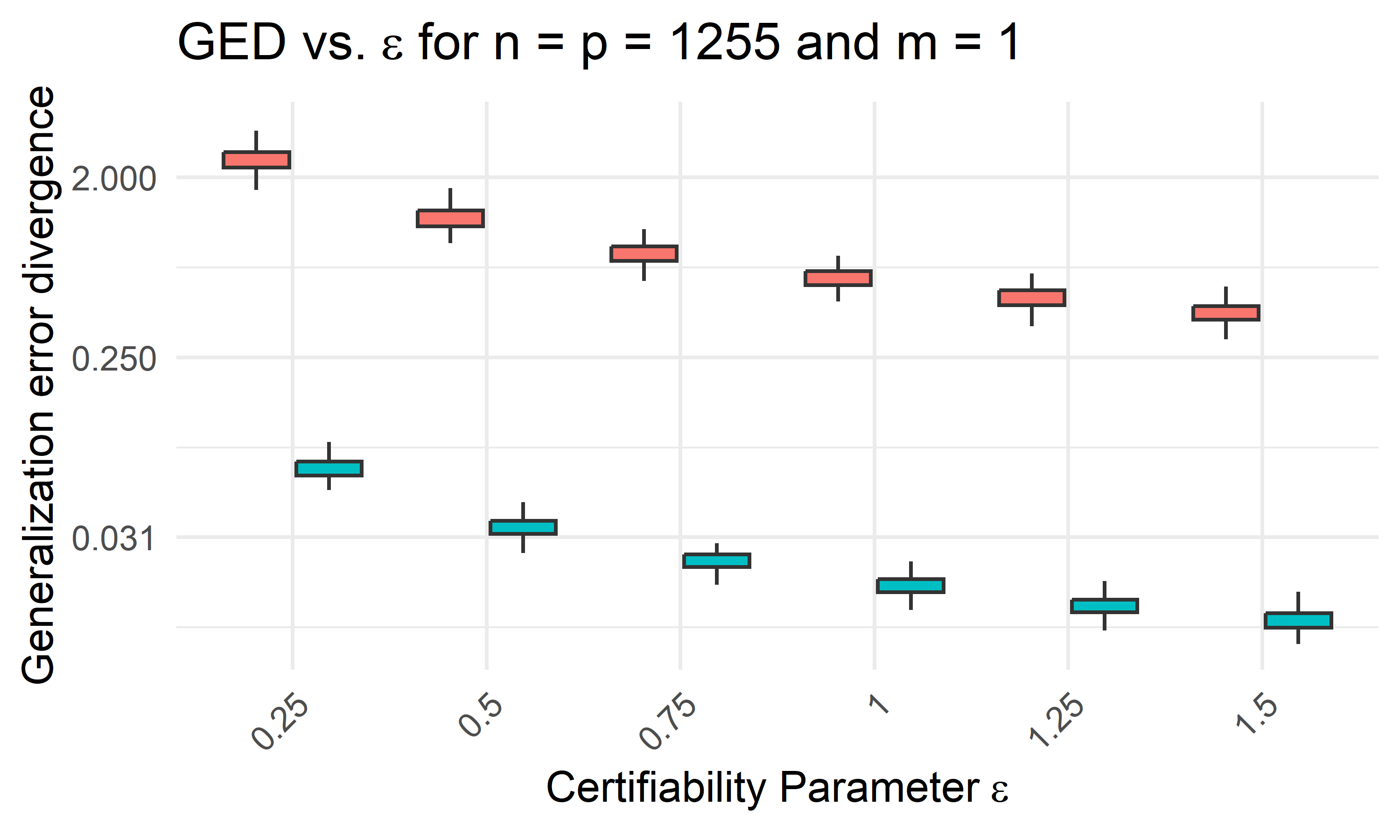}
    \includegraphics[width=0.32\linewidth]{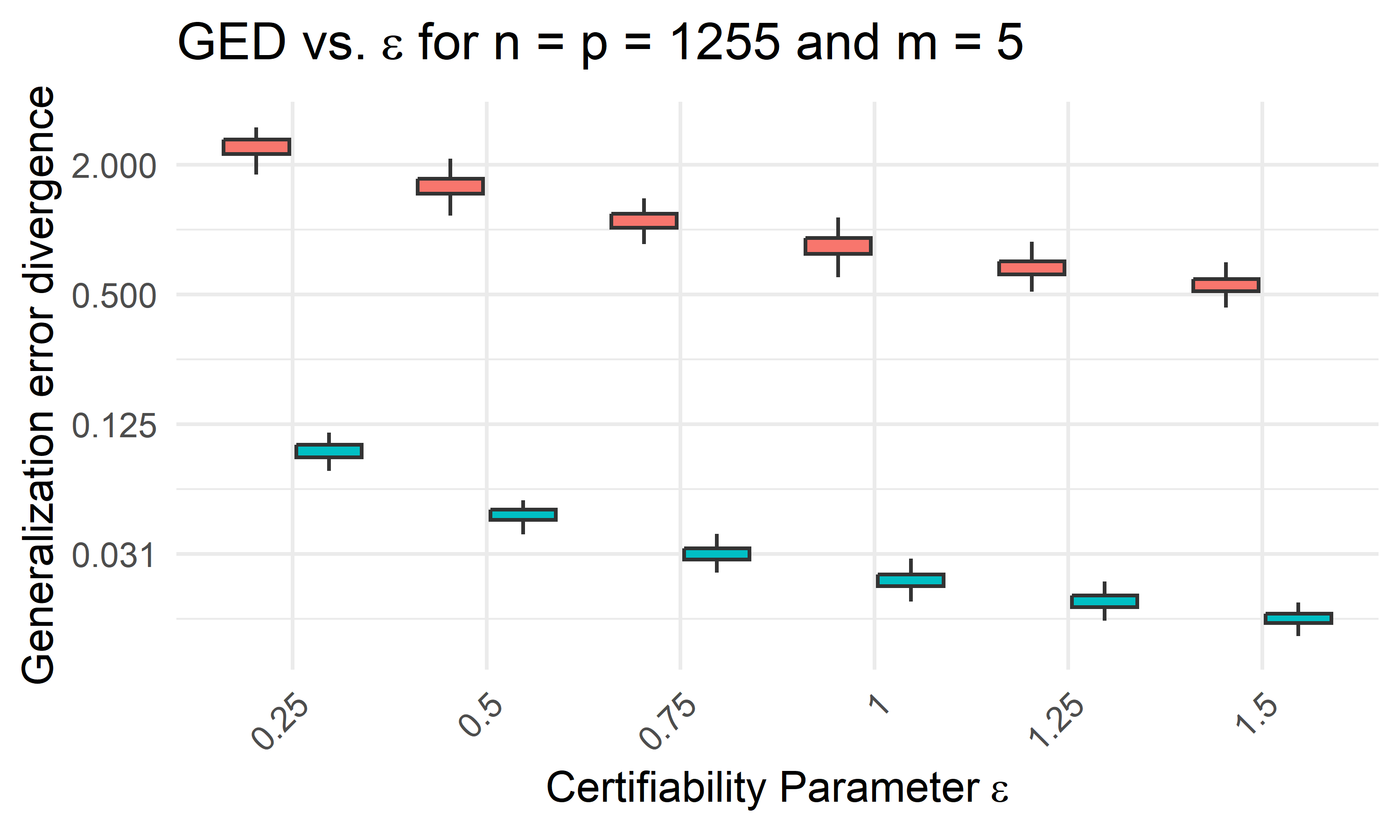}
    \includegraphics[width=0.32\linewidth]{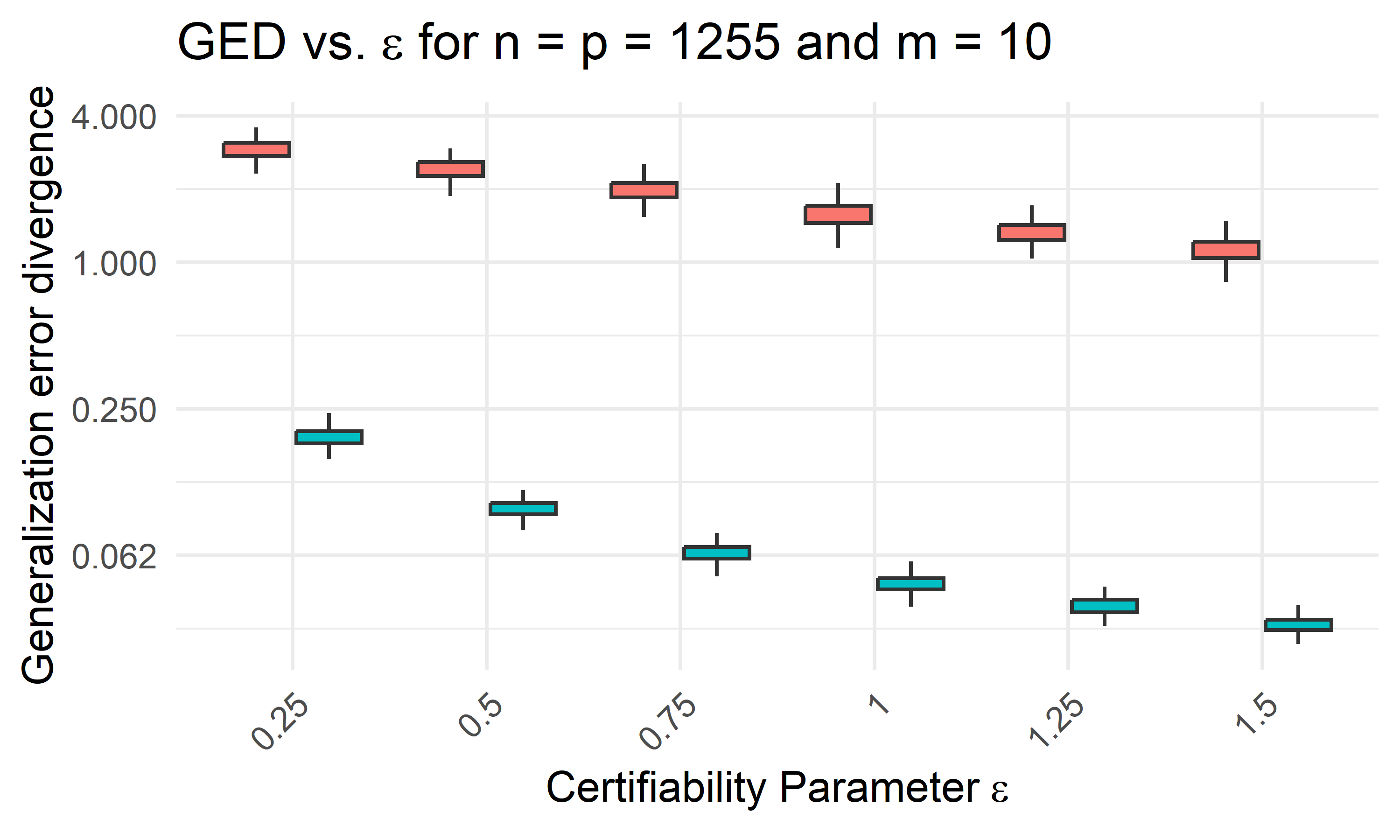}
    \caption{Comparison of GED (plotted in $\log$ scale) across different values of $\ve$ for Laplace noise (in red) vs. Gaussian noise (in cyan). We set $\lambda=0.5$.}
    \label{fig:ged-across-eps}
\end{figure}

\textbf{Second set (meaning of figure 3):} We fix $n=p=1255$ and examine GED with X axis as $\ve$ with  $m=1,5, 10$. Figure~\ref{fig:ged-across-eps} reports boxplots of GED in this setting across 200 Monte Carlo replications.
\par
\textbf{Conclusions from figure 3:} Our Gaussian perturbed estimator performs increasingly closer to that of the ideal retrained estimator as  $\ve \uparrow$. The error for the Laplace-perturbed estimator remains high above its Gaussian counterpart, uniformly over $\ve$, establishing robustness of Gaussian certifiability across all scales $\varepsilon$. Similar behavior is reflected on the unlearned subset (see Section ~\ref{sec:experimenta} in appendix).
\par
\begin{figure}[htbp]
    \centering
    \includegraphics[width=0.32\linewidth]{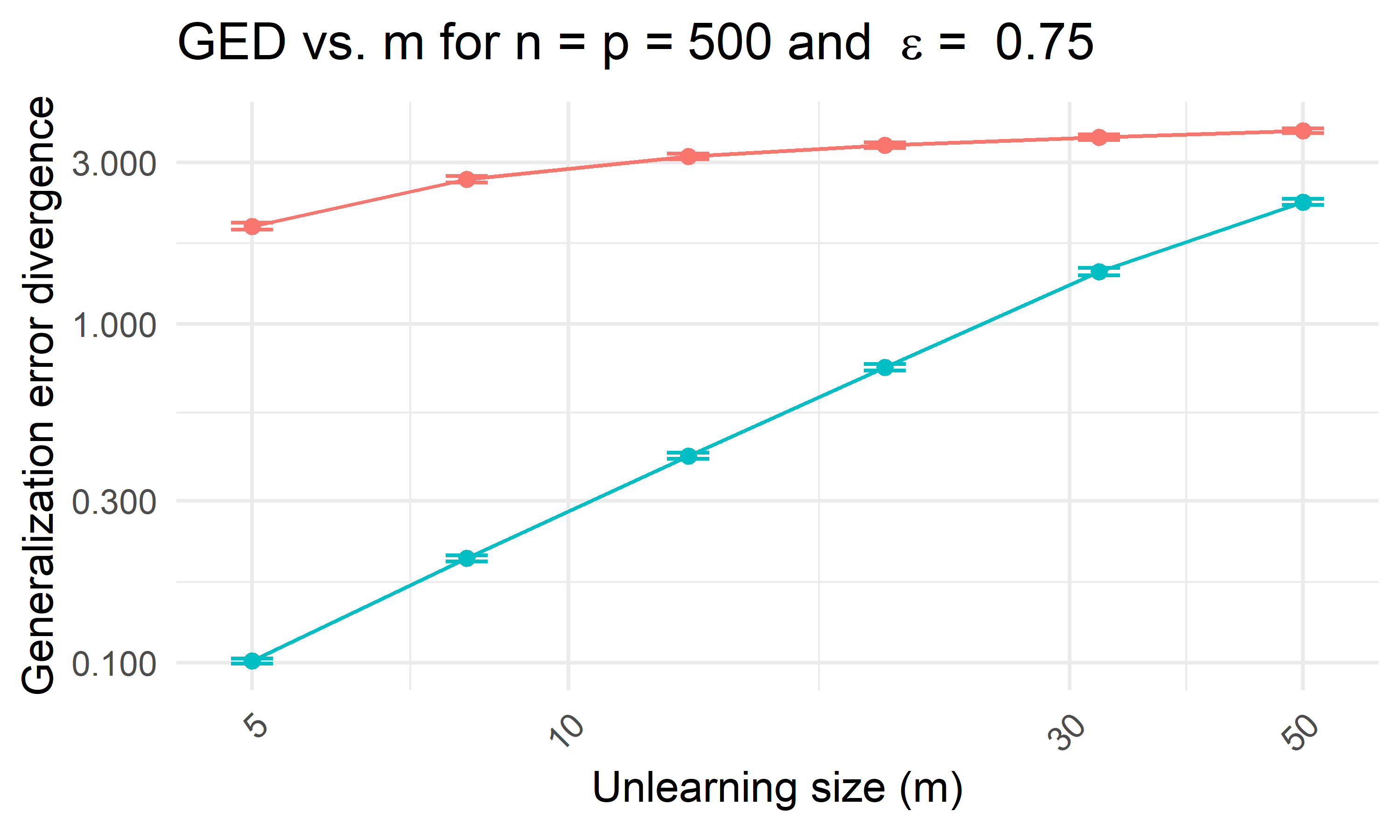}
    \includegraphics[width=0.32\linewidth]{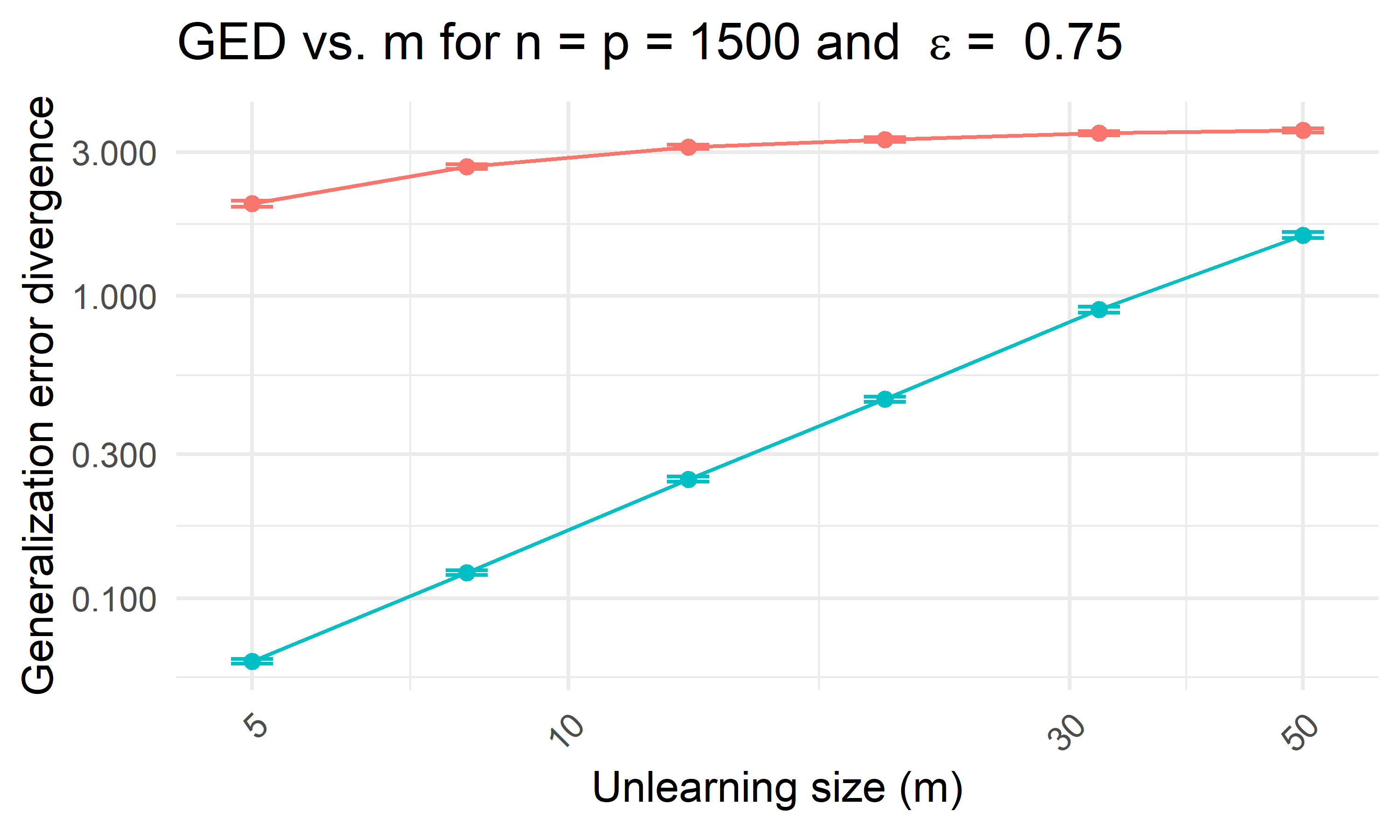}
    \includegraphics[width=0.32\linewidth]{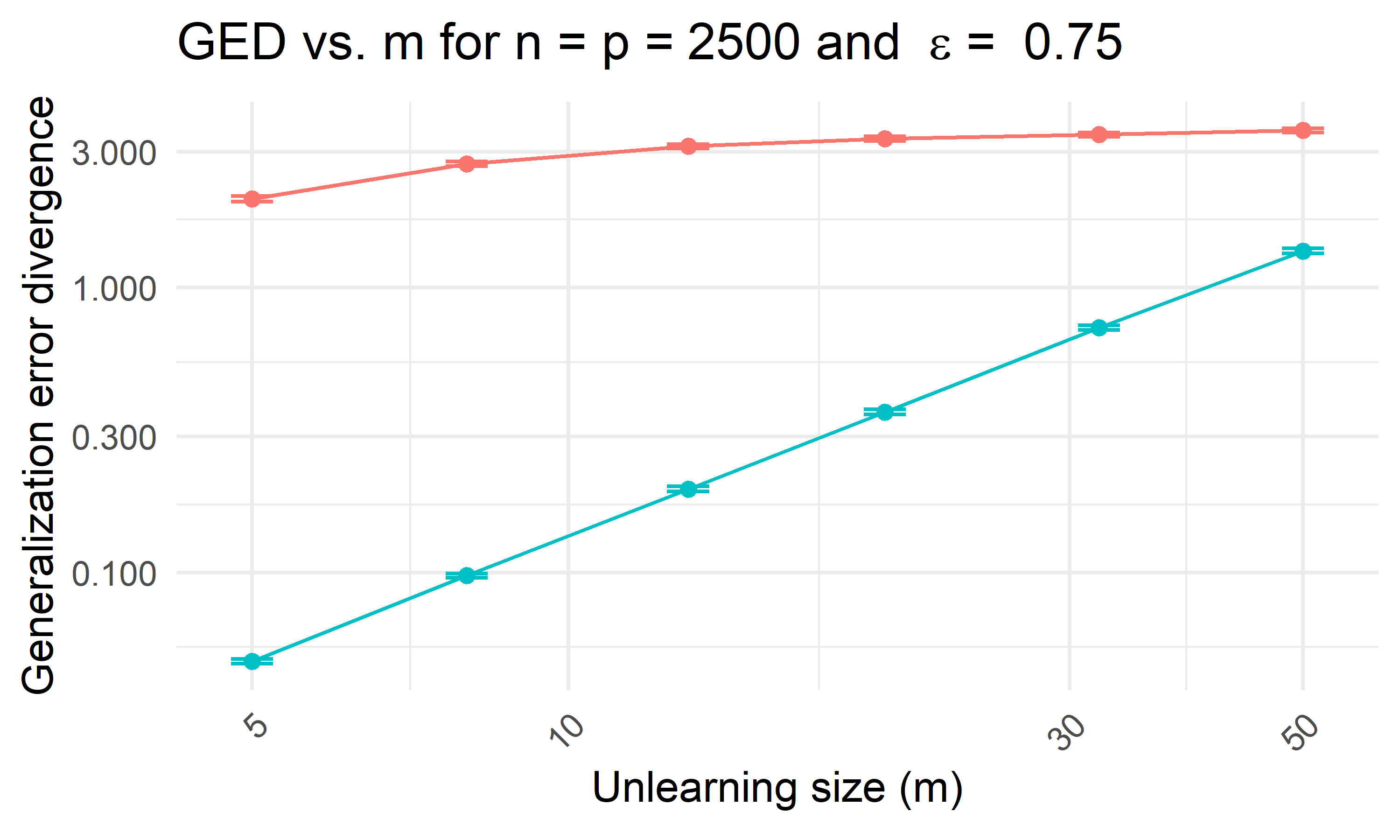}
    \caption{Comparison of mean GED (with 3 SD error bars) across the unlearning size $m$ (both in $\log$ scale) for Laplace noise (in red) vs. Gaussian noise (in cyan). We set $\lambda=0.5$.}
    \label{fig:ged-across-m}
\end{figure}

\textbf{Final set (meaning of figure 4):}, We fix $\ve=0.75$, examine GED  with X axis as $m$ , as $n=p$ varies between $500, 1500, 2500$. We vary $m$ across 6 values from 5 to 50 equally spaced on a log scale. Figure~\ref{fig:ged-across-m} plots the average GED values (200 replications), transforming  X and Y axes on a log scale.
\par
\textbf{Conclusions from figure 4:} For all three plots, as $m\uparrow$, GED increases for both the estimators with Laplace high above Gaussian. A linear regression of $\log({\rm mean(GED)})$ vs. $\log(m)$ confirms this. The lines for Laplace unlearning had slopes of 0.26, 0.22, and 0.24 for $p=500,1500,2500$, whereas Gaussian unlearning had slopes of 1.37,1.42, 1.44, showing that GED values increase at an order of approximately $m^{1.5}$ supporting Theorem~\ref{thm:GED} but suggests a  better dependence on $m$ in practice. 

\vspace{-.1cm}
\section{Conclusion}
\vspace{-.1cm}

We introduce $\varepsilon$-Gaussian certifiability, and show that our noisy one Gaussian-Newton step ensures both privacy and accuracy in the high-dimensional proportional setting. This contrasts with \cite{zou2025certified}, which, under the  $(\phi,\varepsilon)$-certifiability framework, concludes that at least two Newton steps are necessary, even for removing one datapoint. Our results indicate that this gap reflects the suboptimality of their $(\phi,\varepsilon)$-certifiability, which $(\phi,\varepsilon)$-Gaussian certifiability resolves optimally.

\bibliographystyle{iclr2026_conference}
\bibliography{refs}

\appendix
 
\section{Appendix of Gaussian certified unlearning in high dimensions: A Hypothesis Testing approach}

In this supplement, we provide  motivations ~\ref{sec:motivations} for some of the procedures in the paper, more background on Gaussian certifiability ~\ref{sec:Gaussiancertifiability}, discussion and comparison of assumptions on the  per-example loss functions ~\ref{sec:comparison},   proofs of our theoretical results ~\ref{sec:Proof_G}, ~\ref{sec:Proof_SG}, implementation details for our experiments ~\ref{sec:experimenta}, and results from additional simulations. For completeness, we will also restate the proposed Newton-based unlearning algorithm ~\ref{sec:Newtona} and theoretical results ~\ref{sec:resultsa} in the paper.
\section{Motivations behind our approach on a high level} \label{sec:motivations}
In this section, we provide motivations for several aspects of our paper on a high level. First,  we provide intuitions for our approach to describe the problem of machine unlearning. Then we discuss the theoretical assumption of (generalized) linearity of the model of data, and applicability of the framework in general. We end with discussing our experimental metrics and the reasoning behind their focus on prediction-level distributional differences GED and UED.
\par
\subsection{Motivation behind the definition of machine unlearning}
\textbf{On comparing the unlearned $\bar{A}(\hat{\bbeta},\cD_{\cM},T(\cD_n),\bb) $ 
with $\bar{A}(\hbm,\emptyset,T(\cD_{\setminus\cM}),\bb)$   and not with the ideal retrained model $A(\cD_{\setminus \cM})$, for the criteria of $\varepsilon-$ Gaussian certifiability 
.}
\par
\textbf{Condition on $\cD$:} In considering  $\varepsilon-$ Gaussian certifiability,  the following was the reason for comparing the output $\bar{A}(\hat{\bbeta},\cD_{\cM}, T(\cD_n),\bb)$ of `the unlearning algorithm on the full dataset with removed set $\cM$’  and $\bar{A}(\hbm,\emptyset,T(\cD_{\setminus\cM}),\bb)$ `the result of the unlearning algorithm on the dataset without $\cM$ with an empty removed set’.  Our hypothesis testing procedure is conditional on the data $\cD$.
\par
\textbf{Why condition on $\cD$:} This conditioning is required as unlearning happens after learning, so it should be captured almost surely or conditionally on $\cD$. Therefore, the natural base candidate $A(\cD_{\setminus \cM})$, `the learning result on the dataset without $\cM$’ to compare to, becomes deterministic. More precisely, conditionally on $\cD$, the distribution of $A(\cD_{\setminus \cM})$  becomes a Dirac mass as a probability distribution and hence trivially distinguishable from `any other’ probability distribution, unless the learning algorithm is itself randomized, which we do not know a priori (in general).  In our case, the learning algorithm $A$ is just Regularized Empirical Risk Minimization, which is deterministic given the data.
\par
\subsection{Model assumption and applicability of the framework}
 \textbf{On the assumption of (generalized) linearity of the model and applicability of the framework.}
\par
\textbf{GLM in theory:} Within high-dimensional statistics and machine learning, we consider a Generalized linear model $ dP_{\bbeta^*}(\bx,y)= dp(\bx) dq(y|\bx^{\top}\bbeta^*)$ capturing a large class of statistical models, including linear, Poisson, logit, multinomial regression, single index models, etc. This is indeed a large class of models among high-dimensional machine learning models considered in the theoretical literature. 
\par
\textbf{$\varepsilon-$Gaussian certifiability in theory:} Our focus in this paper is on closing the gap by proposing a `universally applicable' definition of certifiability that is well-suited in high dimensions. As argued in the paper, the proposed notion is the most practical, optimally achievable, statistically interpretable, and robust notion of certifiability that optimally captures a broad class of noise adding mechanisms.

\textbf{Practical and theoretical deficiency of previous notions of certifiability:} None of the previously proposed notions of certifiability, including the proposals of \cite{pmlr-v119-guo20c},\cite{sekhari2021remember}, \cite{zou2025certified} on variations of $\varepsilon-$ certifiability, and the proposal of \cite{allouah2025the} on a divergence-based certifiability, have all these nice theoretical properties of $\varepsilon-$Gaussian certifiability simultaneously. This essentially stems from the fact that all these previous notions are incompatible (in achieving optimality) with noise adding mechanisms, the practical way of achieving privacy. We refer the reader to \cite{dong2022gaussian} for elaborate discussions on these deficiencies.

\textbf{$\varepsilon-$Gaussian certifiability in practice:} \textit{Gaussian trade-off curves} are the core of our proposed Gaussian certifiability framework, and they appear in practice, \cite{pawelczyk2025machine}. Here, these authors allow different choices of learning and unlearning algorithms of their choosing than the ones used to establish the theoretical results of the paper.

\textbf{Newton-based unlearning in theory:} Given the superiority of our proposed $\varepsilon-$ Gaussian certifiability framework in high dimensions, we restrict ourselves to showing its utility for a particular noisy Newton-based algorithm \cite{sekhari2021remember} in the Generalized linear model as a foundational step  .

\textbf{Newton-based unlearning in practice: (from GLM to LLM)} Newton-based unlearning is popular in practice and has been used for neural networks in existing papers \cite{pmlr-v119-guo20c}, \cite{bui2024newtonsmethodunlearnneural}. We believe that the theoretical insights gained from analyzing these Generalized linear models can help us better understand the high-dimensional behavior of more sophisticated models, including going beyond proportional regimes, and Neural Networks used in LLMs and vision.

\subsection{Experimental metrics and prediction level evalutations}
\textbf{On the accuracy metrics, and their focus on prediction-level distributional differences}
\par
We do not rule out other accuracy criteria for machine unlearning problems. However, most of the existing literature has also focused on prediction-level accuracy. This emphasis is because such a criterion is more closely aligned with many AI applications, including image classification and predictive modeling. Secondly, a good prediction accuracy usually implies good ‘upper-stream’ accuracies, such as in the training level. This is indeed true in our study, as Theorem \ref{thm:haolinpaper} in the supplement material indicates.  Hence, to make our results comparable with the existing literature and clarify the benefits, we also adopted a notion of prediction level accuracy. For example, see our definition of Generalization Error Divergence in the paper for measuring accuracy. Below, we explain the reason behind picking the experimental metrics GED and UED.
\par
\textbf{GED:} The notion of Generalization Error Divergence focuses on the distributional level generalization error in comparison to previous notions like excess risk of \cite{sekhari2021remember}. It has been shown in \cite{zou2025certified} that such previous notions are not well-suited in high dimensions, which is based on the fact that we can allow unbounded Lipschitz bounds on the loss $\ell$ (indeed with low tolerance) compared to the worst-case bounds, which very much restricts the class of loss functions, particularly in high dimensions, including the ridge regularized empirical risk minimization. 

\textbf{UED:} In the experimental section, we evaluated the prediction accuracy on the Unlearned (Error Divergence) set as well, with the motivation that the unlearned set should not have an undue advantage over new data in terms of prediction. Indeed, the error levels are comparable on the unlearned dataset as well as the new test data, thus validating that the unlearning algorithm has successfully removed the influence of the unlearned set from the training data. 

\section{Gaussian Certifiability and connections with related notions} \label{sec:Gaussiancertifiability}
Having outlined the motivations for our approach on a high level, we now provide intuitions for some of the more technical aspects of the assumptions made in the paper. First, we describe in detail the relations between our Gaussian certifiability framework and other previously proposed frameworks. Second, we describe from a dual perspective the emergence of Gaussian certifiability over any other framework of certifiability in high dimensions. 

\subsection{The Gaussian certifiability framework: relations with other notions}
\textbf{On comparison of Gaussian certifiability with variations of $\varepsilon$-certifiability }
 \par
In short, our definition of Gaussian certifiability (with tolerance $\phi$) implies (for a certain collection of pairs of $\epsilon-\delta$ better known as `privacy profile’), a collection of $\epsilon-\delta$ extension of the unlearning framework proposed in \cite{zou2025certified} with tolerance $\phi$ as well as a collection of the certifiability definition of \cite{sekhari2021remember} with tolerance level $\phi=0$. This follows immediately from Proposition 6 and Corollary 1 of \cite{dong2022gaussian}.

\textbf{On comparison of Gaussian certifiability with Rényi divergence based certifiability}
\par
In short, our definition of Gaussian certifiability (with tolerance $\phi$) also implies the Rényi certifiability framework of \cite{allouah2025the} of all orders `$\alpha$’ (with tolerance $\phi$) for a certain range of the other parameter. This follows immediately from Corollary B.6 in the appendix of \cite{dong2022gaussian}. So, the Gaussian certifiability framework also subsumes the Rényi divergence-based frameworks. 

\textbf{Intricate zoo of different notions of certifiability and  Gaussian certifiability at its center}
\par
Having discussed that Gaussian certifiability (with tolerance $\phi$) implies all previously proposed such notions, including their privacy profile versions, we would like to highlight three key points here. 

\textbf{Lossless conversions:} First, from the \cite{dong2022gaussian} results, Gaussian certifiability with tolerance $\phi$ is in fact equivalent to a collection of $\epsilon-\delta$ unlearning extension of the definition proposed in \cite{zou2025certified} with tolerance $\phi$. Therefore, the conversion is lossless, which is not the case for conversions from Rényi certifiability to $\epsilon-\delta$ certifiability and vice versa.

\textbf{Blackwell ordering:} Secondly, Rényi divergence (even including all orders `$\alpha$’) does not determine Blackwell ordering (see Proposition B.7 in the appendix of \cite{dong2022gaussian}). For us, it means that Rényi divergence-based certifiability does not behave tightly under post-processing of the output of the unlearning algorithm.  This is a necessary requirement for an unlearning algorithm.  In contrast, trade-off functions exactly capture Blackwell ordering (see Theorem 2 of \cite{dong2022gaussian}) and is the core of our definition of Gaussian certifiability, thereby confirming optimal behavior of our certifiability under any arbitrary post-processing.

\textbf{About tolerance $\phi$:} Finally, as mentioned in the paper, we allow the tolerance $\phi$ to be positive, and it is of utmost importance, especially in high dimensions, and is absent in \cite{sekhari2021remember}, \cite{dong2022gaussian}, \cite{allouah2025the}. This allows us derivatives of the unregularized individual loss  $l$ (as well as the regularizer $r$)  to have polynomial growth, in particular, unbounded (albeit with small tolerance $\phi$), whereas \cite{sekhari2021remember}, \cite{pmlr-v119-guo20c}, \cite{allouah2025the} requires worst-case uniform bounds on them (simultaneously with a $\Omega(1)$ strong convexity assumption), which break down in high dimensions even for ridge-regularized empirical risk minimization.

\subsection{Gaussian certifiability as the canonical framework of certifiability}
As mentioned above, the theoretical as well as practical implications of the $(\phi, \varepsilon)$-GPAR framework are significant compared to all previously proposed notions of certifiability, especially in high dimensions. In summary, in (practice) every real-world scenario, where we achieve privacy by adding `isotropic log-concave’ noise, GPAR beats every other notion of privacy, in high dimensions.

\textbf{On Gaussian certifiability as the canonical framework of certifiability in high dimensions}

\textbf{Dual perspective:} A detailed answer to the above question lies in an interesting dual perspective. It is the realization that the starting point in the theoretical investigation of unlearning (privacy) is not to begin with a criterion of privacy, but with how we achieve it in practice. 
\par
\textbf{Step 1:} The most natural and sought after way to achieve unlearning (privacy/certifiability) is to add noise to the (deterministic) algorithm output, and Gaussian noise appears as a natural candidate in high dimensions because of the following intrinsically high dimensional fact that \textit{with high probability (up to a tolerance $\phi$), most  $1$d projections of an isotropic high-dimensional (log-concave) distribution are approximately Gaussian} \cite{diaconis1984asymptotics}, \cite{klartag2007cltconvex}[Theorem 1.1].  
\par
\textbf{Step 2:} In the context of differential privacy, this translates into the fact that up to a tolerance $\phi$, in high dimensions, a large collection of (log-concave) noise adding mechanisms becomes comparable to the Gaussian noise addition mechanism \cite{dong2021central}[Theorem 3.1]. This makes the Gaussian perturbation mechanism the canonical noise adding mechanism in high dimensions.
\par
\textbf{Step 3:} So, up to a tolerance $\phi$, the question reduces to devising an unlearning framework that can tightly capture the Gaussian mechanism. It turns out trade-off functions capture it in the tightest way possible through Blackwell’s theorem \cite{dong2022gaussian}[Theorem 2].
\par
\textbf{Step 4:} More precisely, all the previously proposed notions of unlearning, including $\epsilon-\delta$ variations of \cite{pmlr-v119-guo20c}, \cite{sekhari2021remember}, \cite{zou2024theoretical}, Rényi unlearning of \cite{allouah2025the}, are not tightly achievable under the Gaussian mechanism except for Gaussian certifiability \cite{dong2022gaussian}[Theorem 1, Equation 6]. So, this establishes the natural emergence of the $(\phi, \varepsilon)$-GPAR framework with tolerance $\phi$ as superior over any other framework in high dimensions. 
\section{Assumptions on the unregularized individual loss l, regularizer r and comparison with previous works} \label{sec:comparison}
Before delving into precise comparisons with previous works such as \cite{sekhari2021remember}, \cite{allouah2025the}, we address the motivations behind some of the more generic assumptions on our unregularized individual loss $l$ and the regularizer $r$ in the proportional regime.
\begin{equation} \label{eq:betahata}
\hat{{\bbeta}} =
A(\mathcal{D}_n) :=  \underset{{{\bbeta}} \in \mathbb{R}^p}{\arg\min} \quad  L({\bbeta}) := \underset{{{\bbeta}} \in \mathbb{R}^p}{\arg\min }\quad
\lambda r({\bbeta}) +\sum_{i=1}^n  \ell ( y_i \mid \bx_i^\top {\bbeta} )  ,
\end{equation}
\subsection{On convexity and smoothness of unregularized individual loss l and the regularizer r}
Convexity of the pair of losses $(\ell, r)$ and strong convexity of the regularizer $r$ are assumed to ensure that there exists a unique minimum of the total loss function $L$ above \eqref{eq:betahata}. Smoothness of the pair $(\ell, r)$ is assumed so that the Newton method proposed in the paper makes sense theoretically. 
\par
However, we expect the theorems presented in this paper to extend to certain non-convex loss functions, potentially under stronger constraints on the size of the removal set. More specifically, as long as the local minimizer of the total loss function after data removal remains within the basin of attraction of the local minimizer for the full dataset, and the total loss function is strongly convex within that basin, a similar proof strategy should be applicable. The main remaining challenge, however, is to verify whether these assumptions hold for widely used AI models such as transformers or multilayer perceptrons (MLPs). This is certainly an interesting question for future research and one of the directions we plan to explore further. We also note that all prior theoretical works on certified unlearning cited in our paper, including  \cite{pmlr-v119-guo20c}, \cite{sekhari2021remember}, \cite{allouah2025the}, \cite{zou2025certified} assume convexity of the pair $(\ell,r)$.

\subsection{On the regularization parameter being O(1)} \label{ssec:regularize}
\textbf{Step 1- Natural growth of norm $\|\bbeta^*\|_2 = B= \Omega(\sqrt{p})$ of the true parameter $\bbeta^*$:} In the high-dimensional proportional regime $p\sim n$, it is natural to assume that the $l_2$ norm  $\|\bbeta^*\|_2$ of the true underlying parameter $\bbeta^*$ is atleast of order $\sqrt{p}$\footnote{This is exactly assumption 3 of \cite{sekhari2021remember} where the parameter $B$  should be treated as of order $\Omega(\sqrt{p})$. This is one such quantity that contributes to the blowup in all estimates of their paper.} One reason for such an assumption comes from a standard volume estimate of the ball of radius $B$ in dimension $p$ \cite{vershynin2018hdp}[Exercise 4.29]    
\begin{equation}
\operatorname{Vol}_p(B_p(B))
=\frac{\pi^{p/2}}{\Gamma\!\left(\tfrac p2+1\right)}\,B^{p}
\;\sim\;
\frac{1}{\sqrt{\pi p}}\left(\frac{2\pi e\,B^{2}}{p}\right)^{\!p/2},
\end{equation}
This immediately implies that unless $B > \sqrt{\frac{p}{2\pi e}}$, volume of the ball goes to zero. This translates to the fact that unless $B > \sqrt{\frac{p}{2\pi e}}$, volume of our possibilities for $\bbeta^*$ shrinks to nothing as $p \to \infty$, and therefore not an interesting regime. Moreover, a stricter but more intuitive assumption on $\bbeta^*$  causing this $l_2$ growth is that the coordinates of the underlying parameter $\bbeta^*$ are of order one individually, as is often assumed for precise high-dimensional asymptotics \cite{miolane2021distribution}
\begin{equation}
\text{weak convergence }\frac{1}{p} \sum_{i=1}^p \delta_{\bbeta^*_i} \overset{d}{\to} \mu \text{ and } \lim_{p \to \infty} \frac{1}{p} \sum_{i=1}^p \bbeta_{i}^{*2} = \int x^2 d\mu(x) = \Theta(1)   
\end{equation} for some Borel probability measure $\mu$ on $\mathbb{R}$ with finite and positive second moment. 
\par
\textbf{Step 2- balance between noise and regularizer:} Consider the simplest case of ridge regularized $r(\beta)= \|\beta\|^2$ quadratic loss  $l(y, x^T\beta)= (y-x^T\beta)^2$\footnote{The same computation holds more generally in our setting with the same scaling for $n$ and $p$ upto constants.}. Observe that the total loss has the form
\begin{equation}
    L(\beta)= \|y-X\beta\|^2+ \lambda \|\beta\|^2
\end{equation}
Now, given that the true minimizer $\bbeta^*$ has order one coordinates, we would also want to have order one coordinates for the minimizer $\hat{\bbeta}$ of the empirical total loss $L$. In this case we even have
\begin{equation}
    \hat{\bbeta}= (X^TX+\lambda I_p)^{-1} X^Ty.
\end{equation}
Now, under our normalization of each row $x_i$ of $X$  is  i.i.d. (sub)-Gaussian $\mathcal{SN}(0,\bm{\Sigma})$ with $\lambda_{\text{max}}(\bm{\Sigma}) \leq \frac{C}{p}$, we immediately have that $\|X^TX\| = \Theta(1)$ with  high probability \cite{vershynin2018hdp}[Exercise 4.43]. Therefore, one would want $\lambda = \Theta(1)$ to not develop significant bias either towards the Hessian $X^TX$ of the actual loss, nor towards the Hessian $\lambda I_p$ of the regularization term\footnote{This example makes it clear that according to the normalization in \cite{sekhari2021remember}, the regularization parameter also called $\lambda$ should be taken to be of order $\Theta\left(\frac{1}{n}\right)$ throughout their paper. This is another such quantity that contributes to the blowup of all their results.}.
\subsection{On comparison with Sekhari et al. (2021)}
In this section, we show that the results of \cite{sekhari2021remember} blow up in our high-dimensional proportional regime \eqref{eq:betahata}, and motivates our definition of Generalization Error Divergence. First, observe that  \cite{sekhari2021remember} studies a one-step Newton method with Gaussian perturbation as a machine unlearning algorithm that is similar to our work.

\textbf{Step 1- mapping the optimization problem and loss functions:} To map their result to our setting, we first set the common ground for  the optimization problem of the estimator $\hat{\bbeta}$ and define the per-example loss $f$ in terms of the unregularized individual loss $\ell$ and regularizer $r$ 
\begin{equation}
\hat{\beta}= \arg\min_{\beta}  \frac{1}{n} \sum_{i=1}^n f(\beta, \mathbf{z}_i)= \arg\min_{\beta} \sum_{i=1}^n f(\beta, \mathbf{z}_i), \text{ where } 
\end{equation}
   
\begin{equation} \label{eq:perexmapleloss}
f(\beta, \mathbf{z}) := \ell(y; \bx^\top \bbeta) + \frac{\lambda}{n} r(\bbeta), \quad \text{and} \quad F(\bbeta) := \mathbb{E}[f(\bbeta)].
\end{equation}
\par
 \textbf{Step 2- mapping the assumptions:} \cite{sekhari2021remember}[Theorem 4, Second equation 24] assumes that $f$ is worst case  $L= O(1)$-Lipschitz, worst case  $\lambda= \Theta\left(\frac{1}{n}\right)$-strongly convex, and worst case $M =O(1)$-Hessian Lipschitz, and show the following analogue of Generalization error divergence result of ours (Theorem \ref{thm:GED}) after one noisy-Newton step  (substitute $\|\bbeta^*\|_2=B= \Omega(\sqrt{p})$).
\begin{align}\label{eq:sekhari:bound}
\mathbb{E} \left| F(\tilde{\bbeta}_{\backslash \cM}) - \min_{\bbeta} F(\bbeta) \right| 
&
= O\left( \frac{\lambda B^2}{2}+ \frac{\sqrt{p} M m^2 L^3}{\lambda^3 n^2 \varepsilon} \sqrt{\log \left(\frac{1}{\varepsilon} \right)} + \frac{mL^2}{\lambda n} \right)
\\
&
= 
O\left(1 + \frac{n\sqrt{n} M m^2 L^3}{ \varepsilon} \sqrt{\log \left(\frac{1}{\varepsilon} \right)} + mL^2 \right)
\end{align}

Under our high-dimensional proportional regime $p\sim n$, and from the arguments in ~\ref{ssec:regularize} it follows immediately that $\lambda = \Theta\left(\frac{1}{n}\right)$, $M = \Theta(1)$, and $L = \Theta(1)$. This simplifies \eqref{eq:sekhari:bound} to:
\begin{align}\label{eq:sekhari:bound1}
\mathbb{E} \left| F(\tilde{\bbeta}_{\backslash \cM}) - \min_{\bbeta} F(\bbeta) \right| 
= 
O\left(1 + \frac{n\sqrt{n} M m^2 L^3}{ \varepsilon} \sqrt{\log \left(\frac{1}{\varepsilon} \right)} + mL^2 \right)
\end{align}

which diverges as $n, p \to \infty$ with $n/p \to \gamma$, making the result inapplicable in high-dimensions.
\par
\textbf{Summary of the failure of \cite{sekhari2021remember} results and motivation for GED:} To summarize, the bound above \eqref{eq:sekhari:bound1} essentially concludes that one-step noisy Newton method of \cite{sekhari2021remember} under the notion of $(\epsilon, \delta)$-unlearning does not achieve vanishing change in generalization capabilities (in fact it diverges), and therefore remains inconclusive in our high dimensional setting.
\par
We emphasize that the failure of the above measure of accuracy (called  \textit{excess risk}) \eqref{eq:sekhari:bound} in high dimensions motivates us to consider a different way to measure accuracy, namely through the Generalization error divergence. which we show does go zero even in the high dimensional proportional regime.
\par
\textbf{Remark 1- On the regularization factor $\lambda$ (an important normalization):} Observe the $\frac{\lambda}{n}$ is used in \eqref{eq:perexmapleloss} instead of $\lambda$, since in \cite{sekhari2021remember} $\lambda\|\beta\|^2$ has been added $n$ times in total and once to every example \cite{sekhari2021remember}[Algorithm 4], whereas we only add it once to the entire pure loss part $\sum_{i=1}^n l(y_i,x_i^T\beta)$. According to the arguments of ~\ref{ssec:regularize} $\lambda= \Theta(1)$ in our normalization of $\|X^TX\| = \Theta(1)$. This translates to $\lambda = \Theta\left(\frac{1}{n}\right)$ in their normalization, causing blowup in \eqref{eq:sekhari:bound}.
\par
These changes in normalization of $\lambda$, although ineffective in low-dimension, turn out to be quite important in high-dimension. Finally, if some readers are still not convinced by our arguments in ~\ref{ssec:regularize} in favor of $\lambda = \Theta(1)$ in our normalization, observe that \cite{sekhari2021remember}[Corollary 2] also finds the `optimal choice' of $\lambda$ is of constant order in our normalization or of order $\frac{1}{n}$ in their normalization.
\par
\textbf{Remark 2- On the worst case Lipschitz constant $L$ of loss $f$ (a weakening):} This worst case $L= O(1)$ Lipschitzness of the per-example loss $f$ fails even for the simplest ridge regularized $r(\beta)= \|\beta\|^2$ quadratic case $ \beta \to l(y|x^T\beta)= (y-x^T\beta)^2$, since quadratic functions are never worst case Lipschitz. A weakening of this stringent assumption is given in our paper ~\ref{assum:poly} by allowing the Lipschitz constant `$\ld$' of the individual loss $\ell$ to have polynomial growth, which we will prove happens with low tolerance probability $\phi$. The key difference is that our proofs will be intricate and different in comparison to the standard proofs of such results using usual optimization assumptions.
\par
\textbf{Remark 3- On the worst case Hessian Lipschitz constant $M$ of loss $f$ (another weakening):} $M$ does not appear in the simplest case of ridge regularized least squares, but, we weaken this stringent worst case Hessian Lipschitzness assumption by allowing polynomial growth of `$\lddd$' in ~\ref{assum:poly}.
\subsection{On comparison with Allouah et al. (2025)}
In this section, we show in detail that the standard optimization assumptions of $\Omega(1)$ strong convexity and $O(1)$ smoothness of the per-example loss $f$ as made in  \cite{allouah2025the}, fails completely in high dimensional proportional setting of ours. This in fact true in the simplest case of ridge regularized $r(\bbeta)= \|\bbeta\|^2$ quadratic (unregularized individual) loss $l(y|x^T\bbeta)= (y-x^T\bbeta)^2$ case.
\par
\textbf{Step 1-mapping the optimization problem and loss functions: } 
 To map their result to our setting, we first set the common ground for  the optimization problem of the estimator $\hat{\bbeta}$ and define the per-example loss $f$ in terms of the unregularized individual loss $\ell$ and regularizer $r$ 
\begin{equation}
\hat{\beta}= \arg\min_{\beta}  \frac{1}{n} \sum_{i=1}^n f(\beta, \mathbf{z}_i)= \arg\min_{\beta} \sum_{i=1}^n f(\beta, \mathbf{z}_i), \text{ where } 
\end{equation}
   
\begin{equation} \label{eq:perexmapleloss1}
f(\beta, \mathbf{z}) := \ell(y; \bx^\top \bbeta) + \frac{\lambda}{n} r(\bbeta).
\end{equation}

\textbf{Step 2-- Ridge plus quadratic loss:} Denote $\mathbf{z}_i = (x_i,y_i)\in \mathbb{R}^{p+1}$.  \cite{allouah2025the} assumes that per-example loss  $f$ satisfy standard optimization assumptions:  $\bbeta \to f(\bbeta, \mathbf{z}_i)$  is simultaneously
\begin{equation} \label{eq:stconv}
    \text{ worst case }\mu=\Omega(1) \text{ strongly convex } \iff \nabla^2 f(\bbeta, \mathbf{z}_i)\succeq \mu I_p  \text{ for all }\mathbf{z}_i, 
\end{equation}
\begin{equation} \label{Lsmooth}
    \text{ and worse case } L=O(1) \text{ smooth } \iff \nabla^2 f(\bbeta, \mathbf{z}_i) \preceq LI_p \text{ for all }\mathbf{z}_i.
\end{equation}
Now, consider the ridge regularized  case  $r(\bbeta)= \|\bbeta\|^2$ with (unregularized individual) quadratic loss $l(y| x^T\bbeta)= (y-x^T\bbeta)^2$ of our high dimensional setting  $p\sim n$ \eqref{eq:betahata}. Then we have
\begin{equation}
\hat{\bbeta}= \arg\min_{\bbeta}  \frac{1}{n} \sum_{i=1}^n f(\bbeta, \mathbf{z}_i)= \arg\min_{\bbeta} \sum_{i=1}^n f(\bbeta, \mathbf{z}_i), \text{ with } f(\bbeta, \mathbf{z}_i)=  (y_i-x_i^T\bbeta)^2 + \frac{\lambda}{n}\|\bbeta\|^2 
\end{equation}
\begin{equation} 
\nabla^2 f(\bbeta, \mathbf{z}_i)= 2x_ix_i^T + 2 \frac{\lambda I_p}{n}, \text{ and }
\nabla^2 \left( \sum_{i=1}^n f(\bbeta, \mathbf{z}_i)\right)= 2X^TX + 2\lambda I_p.
\end{equation}
\textbf{Step 3- Consequence of Hessian computation:} 
Since all the (distinct) eigenvalues of the Hessian of $f$ are $\{2\frac{\lambda}{n},2\frac{\lambda}{n} + 2\|x_i\|_2\}$, in the high dimensional proportional regime $p \sim n$,  $L=O(1)$ smoothness constant of the per-example loss $f$ inevitably requires the scaling  $\|x_i\|_2 \sim 1$ (with high probability) for all $i \in [n]$ \ref{featurescaling}. This in turn implies $\|X^TX\| = O(1)$ (with high probability) \cite{vershynin2018hdp}[Exercise 4.43] and requires a comparable $\lambda = O(1)$ ~\ref{ssec:regularize}. But, this completely destroys the $\mu=\Omega(1)$ strong convexity of the individual loss $f$, as $\lambda_{min} \left(\nabla^2 f(\bbeta, \mathbf{z}_i)\right) = 2\frac{\lambda}{n} \neq \Omega(1)$. 
\par
\textbf{Summary of the failure of assumptions made in \cite{allouah2025the} in high dimensions:} The above bounds essentially convey that the standard optimization assumptions made about the per-exmaple loss $f$ in \cite{allouah2025the} do not capture the simplest model of ridge regularized least squares in high dimensions, in the context of machine unlearning in high dimensions.
\par
\textbf{Remark 1- Our relaxed set of assumptions:} The failure of the assumptions made in \cite{allouah2025the} demands a new set of relaxed assumptions for the per-example loss $f$ that are well-suited in high dimensions. In this regard, we work with the weaker set of assumptions ~\ref{ssec:lossfunctions}. Most importantly,  our assumptions on  (only requiring) convexity of $\ell$ and $\nu$-Strong convexity of $r$ above  actually means that per-example loss $\bbeta \to f(\bbeta,\mathbf{z})= l(y, x^T\bbeta) + \frac{\lambda}{n} r(\bbeta)$ is only required to be $\frac{\lambda \nu}{n}$ strongly convex, which is significant relaxation of  the requirement of $\Omega(1)$ strong convexity of $f$, capturing the concrete case of the ridge regularized $r(\bbeta)= \|\bbeta\|^2$ quadratic loss $l(y,z)= (y-z)^2$.
\par
We end with the observation that on the algorithmic side \cite{allouah2025the}, \cite{neel2021descent} focuses on proposing first-order methods of machine unlearning based on \textit{gradient descent} and \textit{stochastic gradient descent}, whereas we focus on a second-order Newton-based method.
\section{Noisy Newton-based unlearning mechanism} \label{sec:Newtona}

Newton method is an iterative method for solving optimization problems based on a second-order Taylor approximation of the given optimizing function \cite{boyd2004convex}. 
 \begin{definition} (Newton-Raphson method)
     \label{def: Newton}
    Suppose that $h:\R^p\to\R^p$ is a differentiable function with  an invertible Jacobian matrix $G$ anywhere on $\R^p$ and that $h$ has a root $h(\bbeta^*)=0$. Starting from an initial point $\bbeta^{(0)}\in \R^p$, the Newton method is the following iterative procedure: for step $t\geq 1$,
    \begin{equation} \label{eq:Newtona}
      \bbeta^{(t)} := \bbeta^{(t-1)} - \bG^{-1}(\bbeta^{(t-1)})h(\bbeta^{(t-1)}).
    \end{equation}
 \end{definition}
Using the Newton method \cite{sekhari2021remember} suggested the following unlearning algorithm:
\textbf{Approximation step :} starting from $\hat{\bbeta}$, we run one  Newton step with $h= \nabla L_{\setminus \cM}$ to obtain:
\begin{equation} \label{eq:Newton_stepa}
   \hat{\bbeta}^{(1)}_{\setminus\cM} = \hat{\bbeta} - \bG(L_{\setminus \cM})^{-1}(\hat{\beta}) \nabla L_{\setminus \cM}(\hat{\bbeta}), \text{ where } \bG(L_{\setminus \cM}) \text{ is the Hessian of } L_{\setminus \cM}.  
\end{equation}
\textbf{Randomization step:} add a Gaussian noise $\bb \sim  N(\mathbf{0},\sigma^2\II_p)$ to  $\hat{\bbeta}^{(1)}_{\setminus\cM} $ to have the unlearned output
\begin{equation}\label{eq:randomization_stepa}
    \textbf{Noisy one step Newton output: } \bar{A}(A(\cD_n),\cD_{\cM},T(\cD_n),\bb))= \tilde{\bbeta}_{\setminus\cM} : = \hat{\bbeta}^{(1)}_{\setminus\cM} + \bb.  
\end{equation}
\textbf{Intuition and the goal:} First, the choice of $\sigma$ will be discussed in our theoretical results.  Next, note that since $\hat{\boldsymbol{\beta}}^{(1)}_{\setminus\mathcal{M}}$ differs from $\hat{\boldsymbol{\beta}}_{\setminus\mathcal{M}}$, the difference between the two vectors may reveal information about the data to be removed, $\mathcal{D}_{\mathcal{M}}$. Hence,  a standard practice is to obscure it by adding noise $\bb$.

\section{Main theoretical result: assumptions and conclusions} \label{sec:resultsa}

\vspace{-.2cm}
\subsection{Technical assumptions on the loss functions and the data} \label{ssec:lossfunctions}
\vspace{-.2cm}
\textbf{Loss functions:} We describe the assumptions on the unregularized individual loss $l$ and the regularizer $r$ \eqref{eq:betahata}. To ensure that the RERM introduced in \eqref{eq:betahata} has a unique solution (from strong convexity arguments) and Newton unlearning makes sense theoretically, the following assumption are made:
\par
\namedlabel{assum:separabilitya}{(A1)} \textbf{Separability of $r$:}
     $r:\mathbb{R}^p \to \mathbb{R}_{+}$ is separable. More precisely, $r(\bbeta) = \sum_{k\in[p]}r_k(\beta_k)$.
    \par
   \namedlabel{assum:smoothnessa}{(A2)} \textbf{Smoothness of $(l,r)$:}
   $\ell:\RR\times\RR\to\RR_+$ and $r:\RR^p\to\RR_+$ are both thrice differentiable. 
    \par
    \namedlabel{assum:convexitya}{(A3)} \textbf{Convexity of $(l,r)$:}  $\ell$ and $r$ are proper convex, and $r$ is $\nu$-strongly convex  for  $\nu=\Theta(1)>0$.
    \par 
    \namedlabel{assum:polya}{(A4)} \textbf{Polynomial growth:} $\exists$ constants $C,s>0$ such that for all  $y,z$ (denote $\ld(y,z) := \partial_y \ell(y,z)$)
    \begin{equation}
        \max\{\ell(y,z), |\ld(y,z)|,|\lddd(y,z)| \}\leq C(1+|y|^s + |z|^s)  \text{ and  } \nabla^2 r(\bbeta) = \diag[\rdd_k(\beta_k)]_{k\in[p]} 
    \end{equation}
    
    is $C_{rr}(n)$-Lipschitz (in Frobenius norm) in $\bbeta$ for some $C_{rr}(n)=O(\polylog(n))$.
    
\textbf{Relaxed assumptions on $(\ell, r)$ and the ridge example:} First, note that these assumptions on $(\ell, r)$ are inspired from \cite{zou2025certified}, and they weaken many of the standard optimization assumptions of $\Omega(1)$ strong convexity and $O(1)$ smoothness of per example loss $f$ in \cite{allouah2025the} or variations thereof in \cite{sekhari2021remember}. Concretely, the ridge regularized $r(\beta)= \|\beta\|^2$ quadratic loss $l(y,z)= (y-z)^2$  satisfies all the assumptions above.  But  \cite{allouah2025the} and \cite{sekhari2021remember} both fail to capture this example in their respective frameworks, as was detailed earlier.

\textbf{Data:} We require the following on the data that generates our features $\bx_i$ and responses $y_i$.
\par 
\namedlabel{assum:xa}{(B1)} \textbf{Subgaussian features:} $\bm{x}_i$ are mean zero sub-Gaussian vectors with covariance $\bm{\Sigma}$ denoted as  $\bm{x}_i\overset{iid}{\sim} \mathcal{SG} (\bzero, \bm{\Sigma})$. We further assume that  $ \lambda_{\max}(\bSigma) \leq \frac{C_X}{p}$, for some constant $C_X>0$.  \label{featurescalinga}
\par
\namedlabel{assum:ya}{(B2)} \textbf{Sub-polylogarithmic responses:} 
   We assume that $\PP(|y_i|>C_y(n))\leq q_y(n)$ and $\EE|y_i|^{2s}\le C_{y,s}$ for some $C_y(n)=O(\polylog(n))$, a constant $C_{y,s}$ and $q_n^{(y)}=o(n^{-1})$.

\textbf{Intuition behind features and responses:} Assumption~\ref{assum:x} on mean-zero sub-Gaussian features is a mild and frequently made in many papers on high dimensional statistics. See for example \cite{miolane2021distribution,rad2018scalable,auddy24a, zheng2017does, donoho2016high, bellec2025corrected}. Assumption ~\ref{assum:y} is not a stringent assumption either, and most popular models such as logistic regression, Poisson regression, and linear regression satisfy this assumption. Using these assumptions we can finally prove our main theoretical results. 

\subsection{Theoretical Guarantees: user privacy and model accuracy}\label{ssec:theorya}

\textbf{One step Newton achieves $(\phi, \varepsilon)$-GPAR:} Our first result shows that for a suitable noise variance, the randomized one-step Newton unlearned estimator defined in \eqref{eq:randomization_step} satisfies $(\phi,\ve)$-GPAR. 

\begin{theorem}\label{thm:Certa}
   Suppose Assumptions~\ref{assum:separability}-\ref{assum:poly} as well as \ref{assum:x}-\ref{assum:y} hold. Then there exist poly-logarithmic  $C_1(n),C_2(n)=O(\polylog(n))$ for which the randomized one-step Newton unlearning \eqref{eq:randomization_step} when used with a perturbation vector $\bb\sim N\left(\mathbf{0}, \frac{r^2}{\ve^2}\II_p\right)$, achieves $(\phi_n,\ve)$-GPAR with 
    \begin{equation} \label{eq:phi_na}
    r = C_1(n) \sqrt{\frac{C_2(n)m^3}{2\lambda \nu n}},
    \quad   \phi_n = nq_n^{(y)} + 8n^{-3} + ne^{-p/2} + 2e^{-p}\to 0.
    \end{equation}
\end{theorem}

The proof of this theorem is presented in the next section.  After the above theoretical guarantee ensuring user privacy, we now calculate the GED of this machine unlearning algorithm. 

\begin{theorem}\label{thm:GEDa}
   Suppose Assumptions~\ref{assum:separability}-\ref{assum:poly} as well as \ref{assum:x}-\ref{assum:y} hold. Consider the unlearning estimator defined in \eqref{eq:randomization_step} with the noise variance set according to Theorem \ref{thm:Cert}. Then, with probability at least $1-(n+1)q_n^{(y)}-14n^{-3}-ne^{-p/2}-2e^{-p}-e^{-(1-\log(2))p}$, 
    \begin{align*}
        {\rm GED}(\tbm,\hbm)
        &\leq C_1(n)\sqrt{C_2(n)}\left(\frac{1}{\ve}+\frac{1}{\sqrt{p}}\right)\sqrt{\frac{m^3(m+2)}{\lambda \nu n}}\cdot\polylog(n).
    \end{align*}
\end{theorem}
\section{Proofs in the case of Gaussian features} \label{sec:Proof_G}

The main goal of this section is to present the proofs of Theorem \ref{thm:Cert} and Theorem \ref{thm:GED} assuming when $\bm{x}_i \sim N(0, \bm{\Sigma})$ with $\lambda_{\max}(\bm{\Sigma}) \leq \frac{C_X}{p}$ for some constant $C_X>0$. After reviewing some preliminary results that will be used in the proof in Section \ref{sec:prelim} we present the proof of Theorem \ref{thm:Cert} in Section \ref{sec:proof:them:cert} and the proof of Theorem \ref{thm:GED} in Section \ref{sec:proof:them:ged}. 

\subsection{Preliminaries}\label{sec:prelim}
Our \textbf{first} lemma summarizes some of the basic but extremely fruitful properties of the Gaussian trade-off function defined in the main text. 

\begin{lemma} (\cite{dong2022gaussian})\label{lem: GTOF} The Gaussian trade off functions satisfy the following properties:
\begin{itemize}
\item Monotonicity:  For any pair  $\ve_1, \ve_2 \in \R$, $\ve_1 \leq \ve_2$ if and only if  
\begin{equation} \label{eq:GMontone}
   f_{ G,\ve_1}(\alpha)  \geq f_{G,\ve_2}(\alpha) \text{ for all } \alpha \in [0,1]. 
\end{equation}
\item Closure under suprema:  For any collection of $(\ve_i)_{i\in I} \subset 
\R$ with index set $I$
\begin{equation} \label{eq:Gsuprema}
   \inf_{i\in I} f_{ G,\ve_i} = f_{G,\sup_{i\in I}\ve_i} \text{ for all } \alpha \in [0,1]
\end{equation}
\item Symmetry: For any  $\mu_1, \mu_2 \in \R^p$ and $\sigma >0$ 
\begin{equation} \label{eq:symmetry}
   T(\mu_2 + \sigma N(0, \II_p), \mu_1+ \sigma N(0, \II_p)) =   T(\mu_1+ \sigma N(0, I_p), \mu_2 + \sigma N(0, I_p))
\end{equation}
\item Dimension freeness: For any  $\mu_1, \mu_2 \in \R^p$ and $\sigma >0$ let  $\ve := \frac{1}{\sigma}\|\mu_1- \mu_2\|_2$. Then
\begin{equation} \label{eq:dto1a}
    T(\mu_1+ \sigma N(0, \II_p), \mu_2 + \sigma N(0, \II_p)) \equiv T(N(0,1), N(\ve, 1)), 
\end{equation}
\end{itemize}
 \end{lemma}
\textbf{Intuition and importance:} The proof of this lemma is immediate from the explicit description of $f_{G, \ve}(\alpha)=\Phi(\Phi^{-1}(1-\alpha) - \ve)$ where $\Phi(\cdot)$ is the cdf of a standard Gaussian variable $\Phi(t)= \mathbb{P}[N(0,1)\leq t]$ for all $t\in \mathbb{R}$. It also requires applying the Neyman-Pearson lemma or the likelihood ratio test \cite{polyanskiy_wu_2025}. But, the conclusions that they imply are extremely powerful. 
\par
\textbf{Monotonicity:} The \textit{monotonicity} condition \eqref{eq:GMontone} reduces an apriori difficult functional comparison between two functions $f$ and $g$ at uncountably many points to a comparison of just one parameter $\varepsilon$.
\par
\textbf{Closure under suprema:} The closure under suprema property \eqref{eq:Gsuprema} essentially says two very important things. First, it makes it easy to identify what the suprema of an apriori arbitrary collection of functions $\{f_i\}_i$ is (in fact explicitly). Second, the limiting object is a function of the same kind: it is again a Gaussian trade-off function with a different choice of parameter $\ve$.
\par
\textbf{Symmetry:} The symmetry property  \eqref{eq:symmetry}  (requires Neyman-Pearson lemma) is very interesting because the definition of a trade-off function $T(P,Q)$ ~\ref{def:TOF} is asymmetric in general, between its first and second arguments.  But, for a pair of shifted isotropic Gaussians, they match because of the spherical symmetry (orthogonal invariance) of the standard Gaussian density in any dimension.
\par
\textbf{Dimension freeness:} The dimension freeness property \eqref{eq:dto1a}  (requires Neyman-Pearson lemma)  makes the case for Gaussian certifiability in high dimensions stronger than any other notion of certifiability. This is because, almost all the results of classical statistics that are true in low dimensions, fail to hold in high dimensions, because many of the quantities involved in controlling the errors are dimension dependent and blows up when $p \uparrow \infty$. It is often the case that finding a dimension-free quantity or even an inequality that \textit{tensorizes} \footnote{We are referring dimension free Poincare and Logarithmic Sobolev inequalities of high-dimensional statistics that are extremely important in obtaining concentration bounds in high dimensions \cite{vanhandelAPC550}.} help us resolve high dimensional issues.
\par
The \textbf{second} preliminary work we will be using in our proof is about the $\ell_2$ difference of the a single step Newton approximation. Define the event $F$ in the following way:
\begin{equation}\label{eq:def-event-F}
 F=
 \left\{
 \max_{|\cM|\le m}
 \|\hat{\bbeta}^{(1)}_{\setminus\cM}-\hat{\bbeta}\|_2
 \ge C_1(n) \sqrt{\frac{C_2(n)m^3}{2\lambda \nu n}}
 \right\}.
\end{equation}
Now, Theorem 3.4 of \cite{zou2025certified} in the notation and setting of this paper becomes:
\begin{theorem}\cite{zou2025certified}\label{thm:haolinpaper}
If the event $F$ is as defined in \eqref{eq:def-event-F}, then
\[
\p(F) \leq \phi_n,
\]
where $\phi_n:= nq_n^{(y)} + 8n^{1-c} + ne^{-p/2} + 2e^{-p}$. 
\end{theorem}

Based on these two results we can now work on the proofs of our main results. 

\subsection{Proof of Theorem \ref{thm:Cert} in the Gaussian case}\label{sec:proof:them:cert}
Recall that we are supposed to show the following with the prescribed choice of $\phi=\phi_n$ as in \eqref{eq:phi_n}.
\begin{equation}  \label{eq:f-PAR1}
      \p \left[ \inf_{|\cM| \leq  m}   
      \min(
      T(\cP_{\rm re}, \cP_{\rm un})(\alpha), T(\cP_{\rm un}, \cP_{\rm re})(\alpha)  \geq f_{G,\ve}(\alpha) \quad
      \text{for all }
      \alpha\in [0,1]
      \right] \geq  1-\phi. 
    \end{equation}
Let  $\bb \sim \sigma N(0,  I_p)$ (independent of everthything else) with $\sigma =\frac{r}{\ve}$. Then, observe  that 
\begin{eqnarray} \label{eq: GPM}
    \cP_{\rm re}
    &\overset{d}{=}&
    \bar{A}(\hbm,\phi,T(\cD_{\setminus\cM}),\bb)
     = 
     \hbm + \bb, \nonumber \\
    \cP_{\rm un}
    &\overset{d}{=}&
    \bar{A}(\hat{\bbeta},\cD_{\cM},T(\cD),\bb)
    =
     \hat{\bbeta}^N_{\cM}+ \bb,
\end{eqnarray}
 Note that since the distributions $\cP_{\rm re}$ and $\cP_{\rm un}$ are conditional distributions given the dataset $\cD$, we can use the dimension freeness property introduced in Lemma \ref{lem: GTOF}, and conclude that 
\begin{equation} \label{eq:dimensionless1}
T( \hbm + \bb, \hat{\bbeta}^{N}_{\cM} + \bb)
=
T(N(0, 1), N(\|\hat{\bbeta}^{N}_{\cM} -  \hbm\|_2/\sigma, 1)),
\end{equation}
\begin{equation}\label{eq:dimensionless2}
T(\hat{\bbeta}^{N}_{\cM} + \bb, \hbm + \bb)
=
T(N(0, 1), N(\|\hat{\bbeta}^{N}_{\cM} -  \hbm\|_2/\sigma, 1)). 
\end{equation}
Using these two equations we conclude the following chain of equalities, all follow from Lemma ~\ref{lem: GTOF}.
\begin{eqnarray} 
      \lefteqn{\p \left[ \inf_{|\cM| \leq  m}   
      \min(
      T(\cP_{\rm re}, \cP_{\rm un})(\alpha), T(\cP_{\rm un}, \cP_{\rm re})(\alpha))  \geq f_{G, \epsilon}(\alpha) \quad
      \text{for all }
      \alpha\in [0,1]
      \right]} \nonumber \\
      &\overset{(a)}{=}& \p \left[ \inf_{|\cM| \leq  m} 
      T(N(0, 1), N(\|\hat{\bbeta}^{N}_{\cM} -  \hbm\|_2/\sigma, 1)) \geq f_{G, \epsilon}(\alpha) \quad
      \text{for all }
      \alpha\in [0,1]
      \right] \nonumber \\
&\overset{(b)}{=}& \p \left[ 
      T(N(0, 1), N(\sup_{|\cM| \leq  m} \|\hat{\bbeta}^{N}_{\cM} -  \hbm\|_2/\sigma, 1)) \geq f_{G, \epsilon}(\alpha) \quad
      \text{for all }
      \alpha\in [0,1]
      \right]      \nonumber \\
&\overset{(c)}{=}& \p \left[\sup_{|\cM| \leq  m} \|\hat{\bbeta}^{N}_{\cM} -  \hbm\|_2/\sigma \leq \epsilon \right] \nonumber \\
&\overset{(d)}{=}& \p \left[\sup_{|\cM| \leq  m} \|\hat{\bbeta}^{N}_{\cM} -  \hbm\|_2/ \leq r \right].
\end{eqnarray}
The above inequalities have been obtained through the following arguments:
\begin{itemize}
\item Equality (a): To obtain this equality we have used \eqref{eq:dimensionless1} and \eqref{eq:dimensionless2}. 
\item Equality (b):  We  used the monotonicity property of $f_{G, \epsilon}(\alpha)$ mentioned in Lemma \ref{lem: GTOF}.  
\item Equality (c): We have used the monotonicity property of $f_{G, \epsilon}(\alpha)$ of Lemma \ref{lem: GTOF} again. 
\item Equality (d): To obtain equality (d) we have used the fact that $\sigma = r/\ve$. 

\end{itemize}

So, it remains to prove the following with $\phi_n, r$ as prescribed in \ref{thm:Cert}.
    \begin{equation}  \label{eq:PCAGR}
        \p \left[ \sup_{|\cM| \leq  m} \|\hat{\beta}^{N}_{\cM} -  \hbm\|_2 \leq r  \right] \geq  1-\phi.
    \end{equation}
Using Theorem \ref{thm:haolinpaper} establishes this result. 

\textbf{Remark on Gaussianity assumption on features:} Observe that our entire computation above was conditional on the data $\cD$, and therefore do not depend on the Gaussianity of the features at all, except while invoking Theorem ~\ref{thm:haolinpaper} to establish \eqref{eq:PCAGR}. We will address this issue later in Section ~\ref{sec:Proof_SG}.

\subsection{Proof of Theorem~\ref{thm:GED} in the Gaussian case}\label{sec:proof:them:ged}

Similar to \cite{zou2025certified} we define the following two events:
    \begin{align*}
        E_5&:=\left\{ |\bx_0^\top\hb| \leq \sqrt{(\lambda\nu)^{-1}C_X\gamma_0(1+C_y^s(n)2c\log(n))} \right\}\\
        E_6&:=\left\{  
        \max_{|\cM|\leq m}
        |\bx_0^\top(\hb-\hbm)|\leq\frac{2C_X}{\lambda\nu}(\sqrt{\gamma_0}+3)\sqrt{\frac{m(2m+c)}{p}\log(n)}\polylog_4(n)  
        \right\}.
    \end{align*}
    We use the following result from \cite{zou2025certified} to establish some bounds on the above events.
\begin{lemma} \label{lem:event56}
    Then for any $\cD\in F^c$, we have 
    \begin{align*}
        \PP(E_5^c|\cD)&\leq \PP\left(|\bx_0^\top\hb|\leq \sqrt{\frac{C_X}{p}}\Vert\hb \Vert\sqrt{2c\log(n)}|\cD\right)\leq 2n^{-c}\\
        \PP(E_6^c |\cD)&\leq \PP\left(\left. 
 \exists|\cM|\leq m, |\bx_0^\top(\hb-\hbm)|>\sqrt{\frac{C_X}{p}}\Vert\hb-\hbm \Vert\cdot 2\sqrt{\log(N)+c\log(n)} \right|\cD  \right)\\
    &\leq 2n^{-c},
    \end{align*}
    where $N=\sum_{s=0}^m{\binom{n}{s}}\leq 2{ \binom{n}{m}}\leq {\rm e}^{2m\log(n)}.$
\end{lemma}
Define the following two events 
    \begin{align*}
        E_7&:= \bigg\{ \forall|\cM|\leq m, |\bx_0^\top(\tbm^{(t)}-\hbm+\bb)|\\
        &\quad 
        \leq 2\sqrt{C_X}
        \left(
        \frac{2}{\ve}
        +\frac{1}{\sqrt{p}}\right)r_{t,n}\cdot \sqrt{(2m+c)\log(n)} \bigg\},\\
        E_8&:=\{ \Vert\bb\Vert\leq \frac{2r_{t,n}\sqrt{2p}}{\ve} ,|y_0|\leq C_y(n)\}.
    \end{align*} 
    Define $\gamma := 2\sqrt{C_X}
        \left(
        \frac{2}{\ve}
        +\frac{1}{\sqrt{p}}\right)r_{t,n}\cdot \sqrt{(2m+c)\log(n)}$. 
Note that using the union bound we have
\begin{eqnarray}
&&\p(E_7^c \cap F^c \cap E_8 | \cD, \bm{b}) \leq {\rm e}^{2m \log n} \p \bigg\{  |\bx_0^\top(\tbm^{(t)}-\hbm+\bb) | \geq \gamma | \cD, \bm{b} \bigg\} \nonumber  \\
&& \leq 2  {\rm e}^{2m \log n} {\rm e}^{- \frac{\gamma^2 }{2 \Vert\tbm-\hbm+\bb \Vert^2}}.
\end{eqnarray}
    By definition of $F$ from \eqref{eq:def-event-F}, under $F^c\cap E_8$,
    \[
        \Vert\tbm-\hbm+\bb \Vert\leq \Vert\tbm-\hbm \Vert+\Vert\bb\Vert\leq 
        \left(\frac{2\sqrt{2p}}{\ve}+1\right)r_{t,n}.
\]

    Then, following the proof of Theorem 3.2 in \cite{zou2025certified} we would have
    \begin{align*}
        \PP((E_7')^c\cap F^c\cap E_8')
        &\leq 2n^{-c}.
    \end{align*}
    Under $(\cap_{i=5}^6 E_i)\cap E_7\cap E_8\cap F^c$, w.p.$\geq 1-(n+1)q_n^{(y)}-14n^{1-c}-ne^{-p/2}-2e^{-p}-e^{-(1-\log(2))p}$, $\forall a\in[0,1]$,
    \begin{align*}
        &~|\bx_0^\top[a(\tbm+\bb)+(1-a)\hbm]|\\
        \leq&~ |\bx_0^\top\hbm| +a|\bx_0^\top(\tbm^{(t)}-\hbm+\bb)|\\
        \leq&~ \polylog_{1}(n)+2\sqrt{C_X}\left(
                \frac{2\sqrt{2}}{\ve}
        +\frac{1}{\sqrt{p}}\right)r\cdot \sqrt{(2m+c)\log(n)}\\
        \leq&~ \polylog_{2}(n),
    \end{align*}
    provided $r_{t,n}=o\left(\frac{\ve}{\sqrt{m}\polylog(n)}\right)$ so that the second term is $O(\polylog(n))$. So we have
    \begin{align*}
        &~|\ld_0((\tbm+\bb)+(1-a)\hbm)|\\
        \leq&~
        1+C_y^s(n)+|\bx_0^\top[a(\tbm+\bb)+(1-a)\hbm]|^s\\
        \leq&~ \polylog_{3}(n),
    \end{align*}
    and
    \[
        |\bar{\ld}_0|=\left|\int_0^1 \ld_0((\tbm^{(t)}+\bb)+(1-a)\hbm) da \right|\leq \polylog_{3}(n).
    \]
    Finally we have that, under $(\cap_{i=5}^6 E_i)\cap E_7'\cap E_8'\cap F^c$,
    \begin{align*}
        &~{\rm GED}(\tbm,\hbm)\\
        =
        &~|\ell_0(\tbm+\bb)-\ell_0(\hbm)|\\
        \leq& ~|\bar{\ld}_0\Vert\bx_0^\top(\tbm-\hbm+\bb)|\\
        \leq&~ \polylog_{3}(n)2\sqrt{C_X}\left(
        \frac{2}{\ve}
        +\frac{1}{\sqrt{p}}\right)r\cdot \sqrt{(2m+c)\log(n)}\\
        \le 
        &~
        \polylog_{4}(n)2\sqrt{C_X}\left(
        \frac{2}{\ve}
        +\frac{1}{\sqrt{p}}\right)r\cdot \sqrt{(2m+c)\log(n)}
        .
    \end{align*}
    Plugging in the value of $r$ from Theorem~\ref{thm:Cert} finishes the proof. 

    \textbf{Remark on Gaussianity assumption on features:} Observe that the entire computation above was reduced to getting the correct orderwise bound on the lablled set of the events $E_5,E_6, E_7, E_8,E_7',E_8'$ above and event $F$ ~\ref{eq:def-event-F}. Now, all the labelled events  $E_5,E_6,E_7,E_8, E_7', E_8'$ involves a fresh feature data $\bm{x}_0$ essentially in its linear component $\bm{x}_0^Tv$ for some vector $v$. Now, it is well known \cite{vershynin2018hdp} that for sub-Gaussian vector $\bm{x}_0 \sim \mathcal{SG}(0, \Sigma)$, the linear projection $\bm{x}_0^Tv$ has the same (orderwise) Gaussian tail beahvior as if $\bm{x}_0$ was $\mathcal{N}(0,\Sigma)$, since in all these cases $v$ is independent of the fresh sample $\bm{x}_0$ and therefore, the same orderwise bounds of $E_5,E_6,E_7,E_8, E_7',E_8'$ continue to hold except while invoking Theorem ~\ref{thm:haolinpaper} to establish \eqref{eq:PCAGR} or equivalently the tail of ~\ref{eq:def-event-F}. We will address this issue next.
\section{An extension of the proof in the case of sub-Gaussian features} \label{sec:Proof_SG}
Given the arguments already presented in the proofs of Theorems ~\ref{thm:Cert} and ~\ref{thm:GED} in section ~\ref{sec:Proof_G}, it is enough to establish that the Theorem  ~\ref{thm:haolinpaper} of \cite{zou2025certified} can be extended from $x \sim \mathcal{N}(0, \bm{\Sigma})$ to $\bm{x} \sim \mathcal{SG}(0, \bm{\Sigma})$ with the same assumptions on $\bm{\Sigma}$. We now describe why is this possible. An inspection of the proof of   Theorem ~\ref{thm:haolinpaper}  of \cite{zou2025certified} reveals that their entire proof never uses the following potentially Gaussian facts that fails when one goes from Gaussian to sub-Gaussian vectors. 1) It never uses orthogonal invariance of the feature vectors in the proof. 2) It never uses Gaussian conditioning technique: ($X|Y$ is Gaussian if $(X,Y)$ is jointly Gaussian). Rather it relies on standard concentration bounds collected in Lemma B.7 –B.11 in the appendix of \cite{zou2025certified}[arxiv version May 12] which can be all be generalized from Gaussian to sub-Gaussian vectors with exact same (mean, covariance) via a standard computation of its moment-generating functions.\footnote{A sub-Gaussian vector  $x\sim \mathcal{SG}(0, \bm{\Sigma})$ by definition satisfies  $\mathbb{E}[e^{v^T x}] \leq e^{ \frac{1}{2} v^T\bm{\Sigma}v}$ $\forall v\in \mathbb{R}^p$   \cite{vershynin2018hdp}.} Morevoer, their proofs uses the well-known fact that an $m$ by $n$ matrix with centered, normalized independent columns has a bounded operator norm (and its Wishart version) \cite{vershynin2018hdp}[Exercise 4.43], be it sub-Gaussian or Gaussian.  More precisely, each of the results in  has an analogous sub-Gaussian extension (with potentially worse constants, which we disregard anyway) and therefore extends verbatim. 

\section{Additional Numerical Experiments}    \label{sec:experimenta}
In this section, we provide results from additional numerical experiments under the ridge penalized $r(\bbeta)=\|\bbeta\|^2$ logistic regression model  $\ell(y|\bx^{\top}\bbeta)$,  with random features  $\bx_i\sim N(\mathbf{0},\frac{1}{n}\II_p)$ and true parameter $  \bbeta_* \sim N(\mathbf{0},\II_p)$, penalty parameter $\lambda=0.5$ as in the main paper ~\ref{sec:experiments}.  This further supports our theoretical results. These were not kept in the main paper for lack of space.

\subsection{Additional plots for n=p}

\begin{figure}[htbp]
    \centering
    \includegraphics[width=0.48\linewidth]{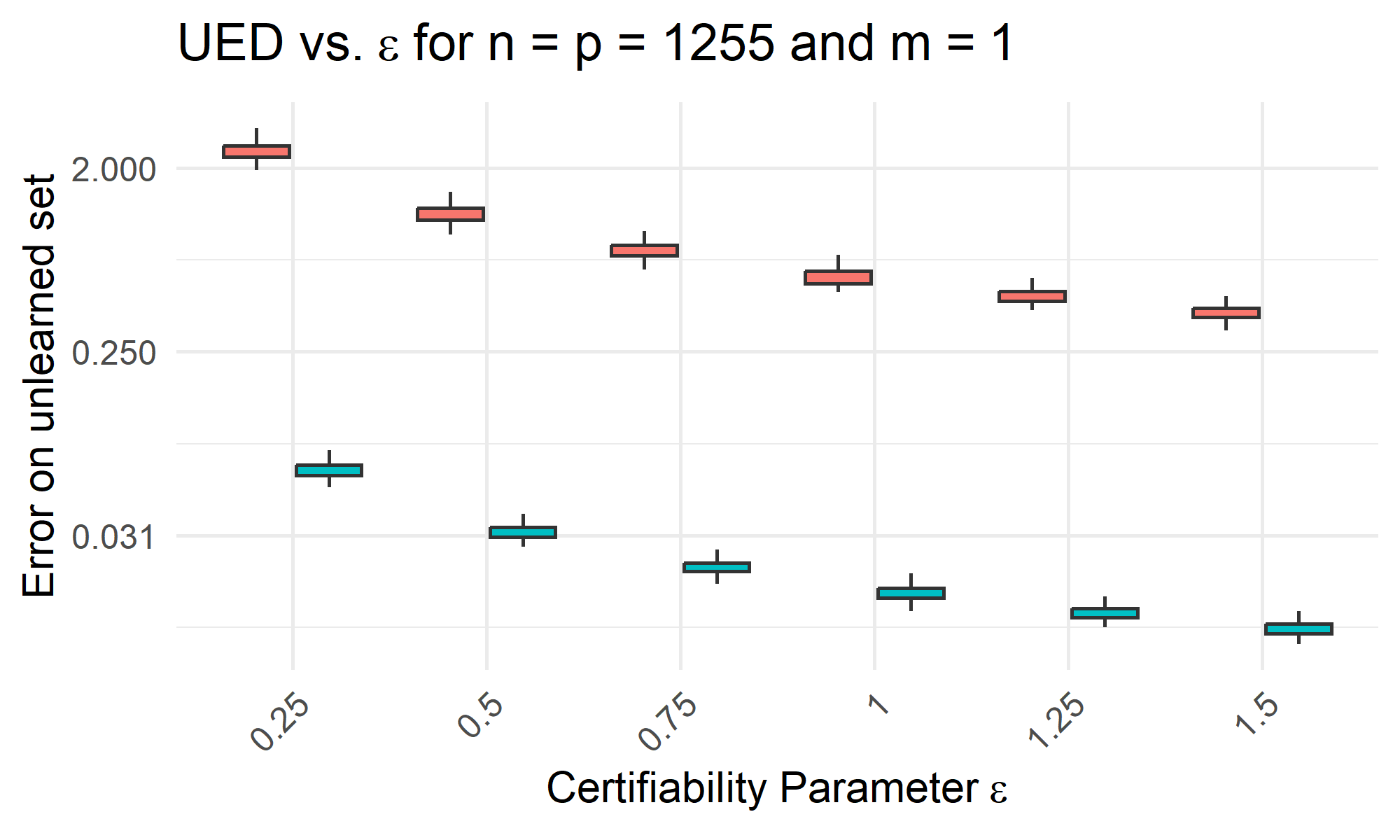}
    \includegraphics[width=0.48\linewidth]{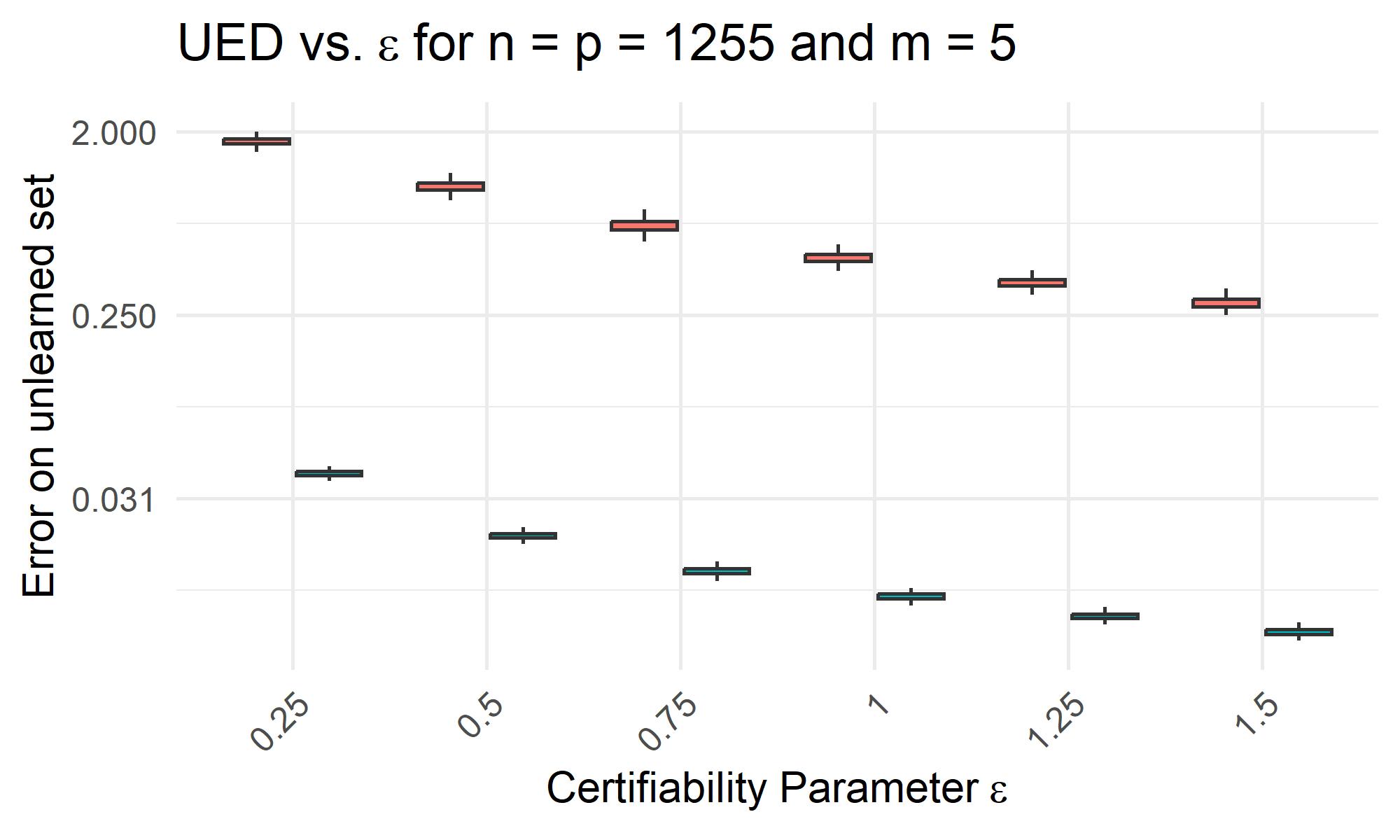}
    \includegraphics[width=0.48\linewidth]{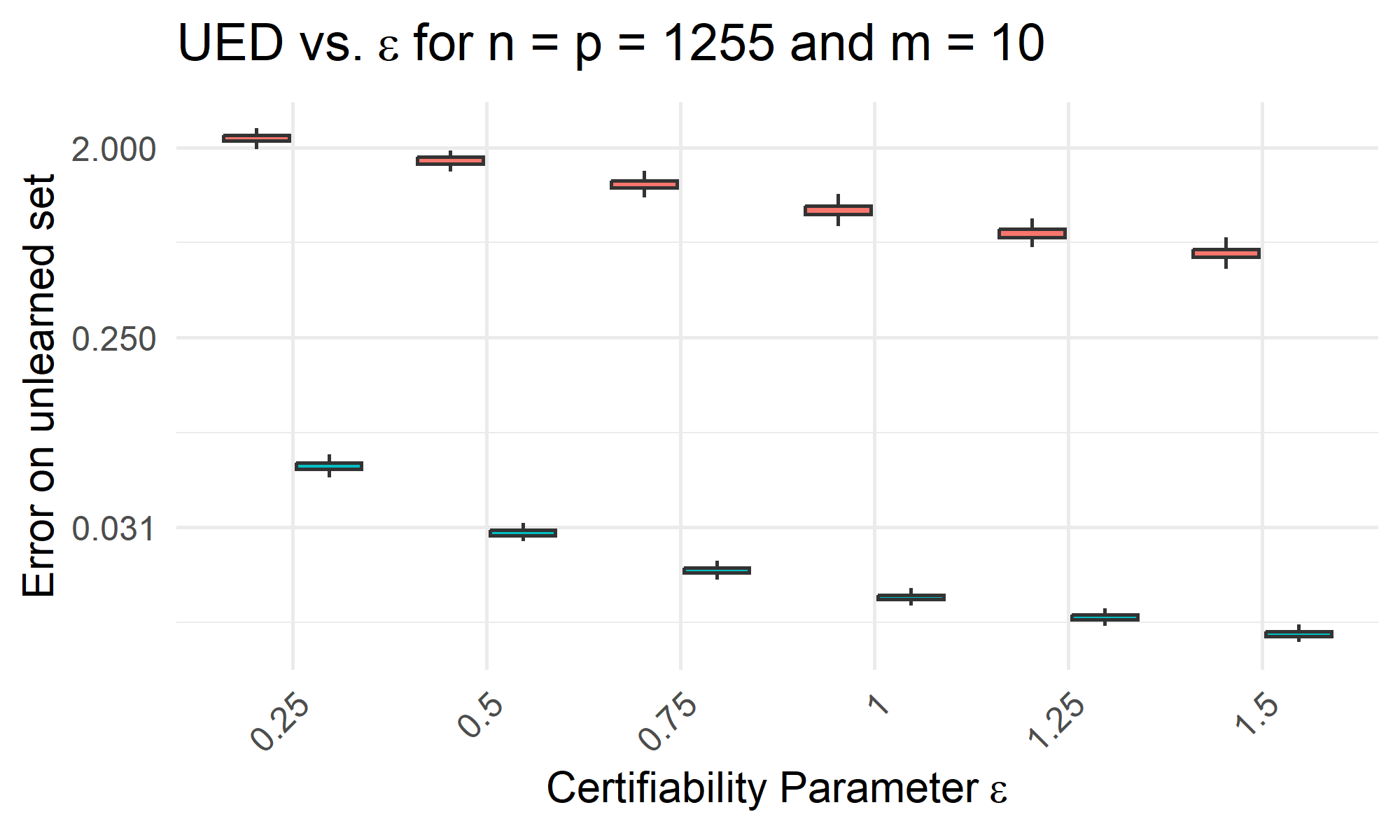}
    \caption{Comparison of the difference in negative log likelihood on the requested removal set (plotted in $\log$ scale) among the retrained estimator, with unlearned estimator with Laplace noise (in red) and Gaussian noise (in cyan). The left, middle, and right figures plot unlearning for removal sizes $m=1$, $m=5$, and  $m=10$ respectively. Here $n=p=1255$.}
    \label{fig:forget-across-eps-del1}
\end{figure}

\textbf{First set (meaning of figure 5) :} We   plot UED: the error divergence of the unlearned estimator on unlearned data, with $n=p=1255$, as we vary $\ve$ on X axis across 6 different values, and take $m=1,5,10$. The mean UED over 200 replications with standard error bars are presented in Figure~\ref{fig:forget-across-eps-del1}.

\begin{figure}[htbp]
    \centering
    \includegraphics[width=0.48\linewidth]{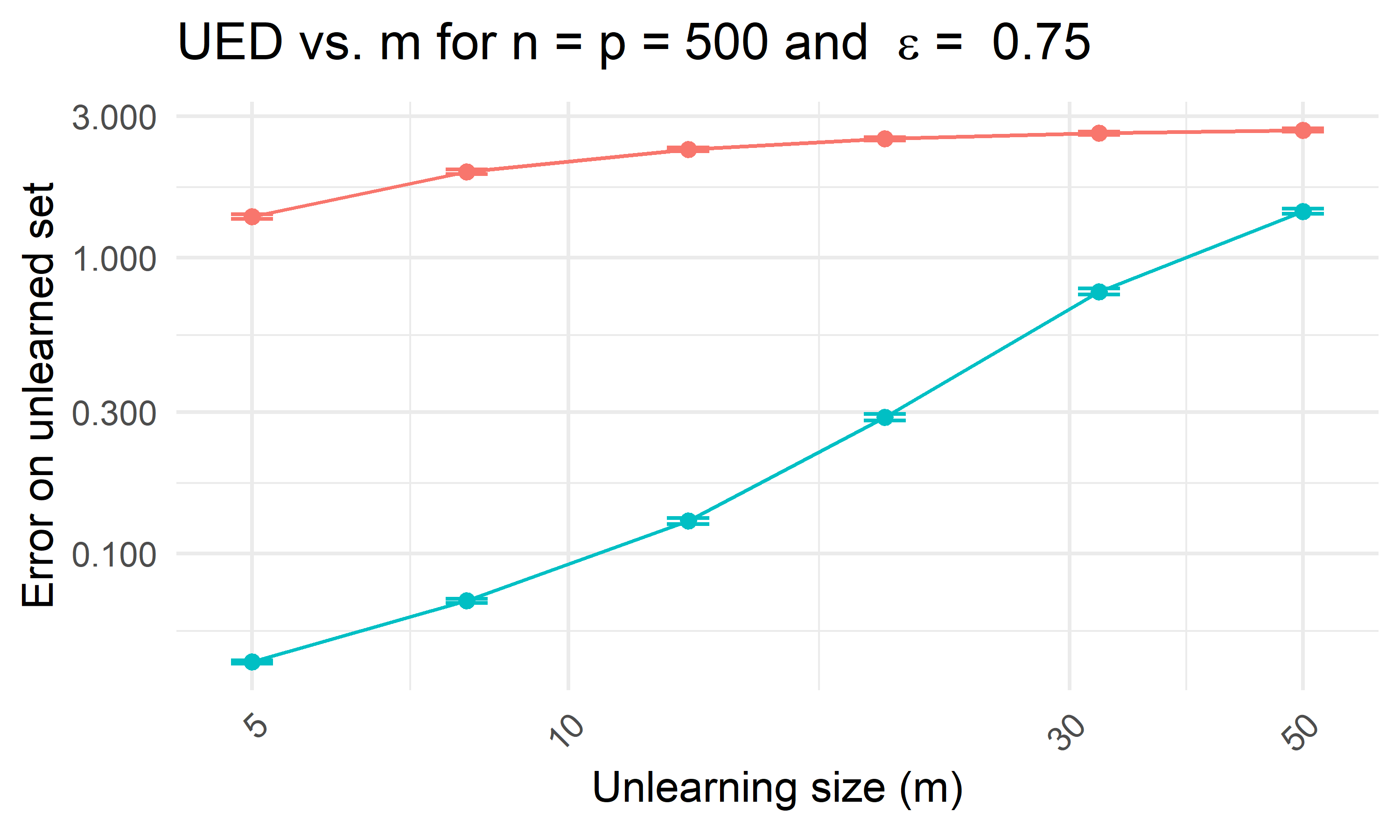}
    \includegraphics[width=0.48\linewidth]{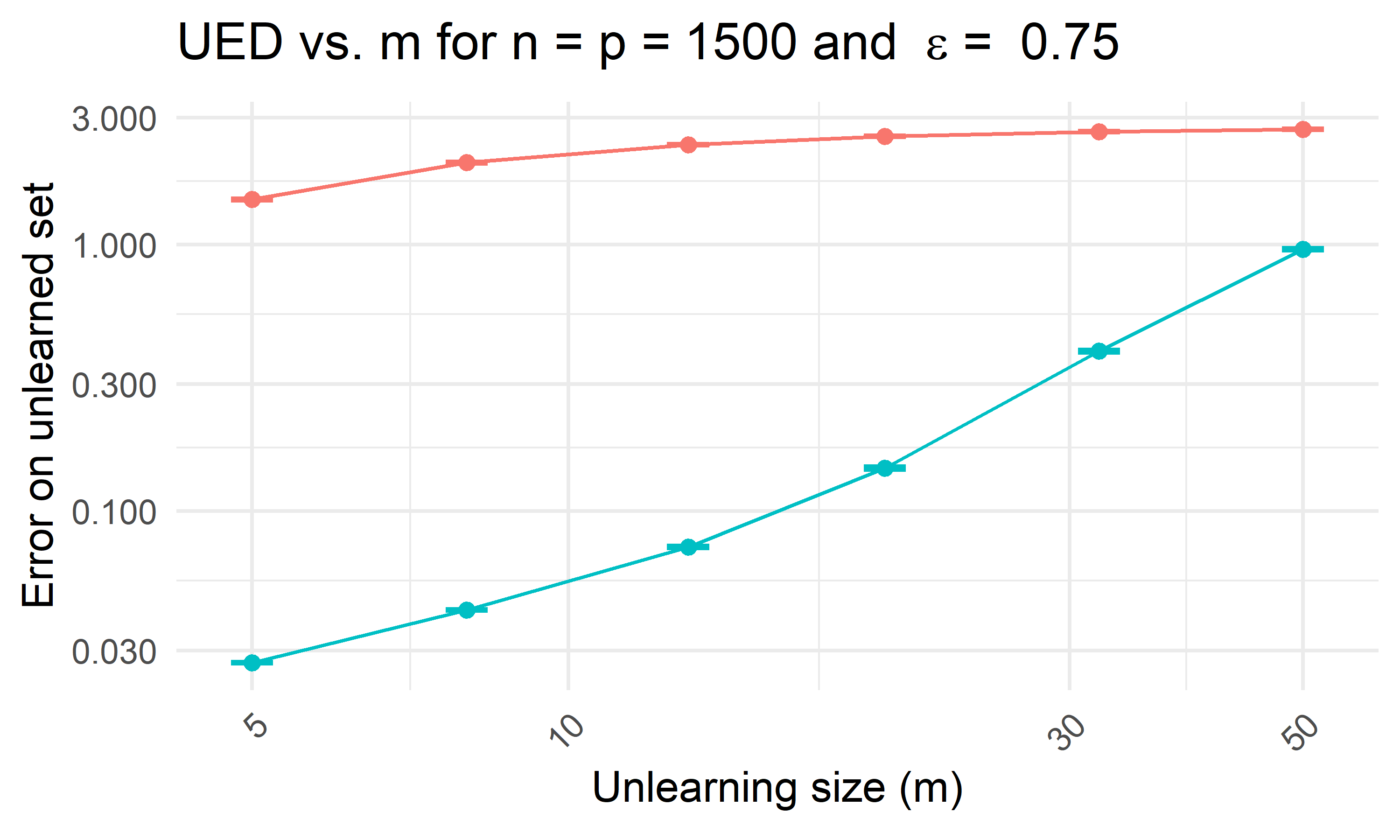}
    \includegraphics[width=0.48\linewidth]{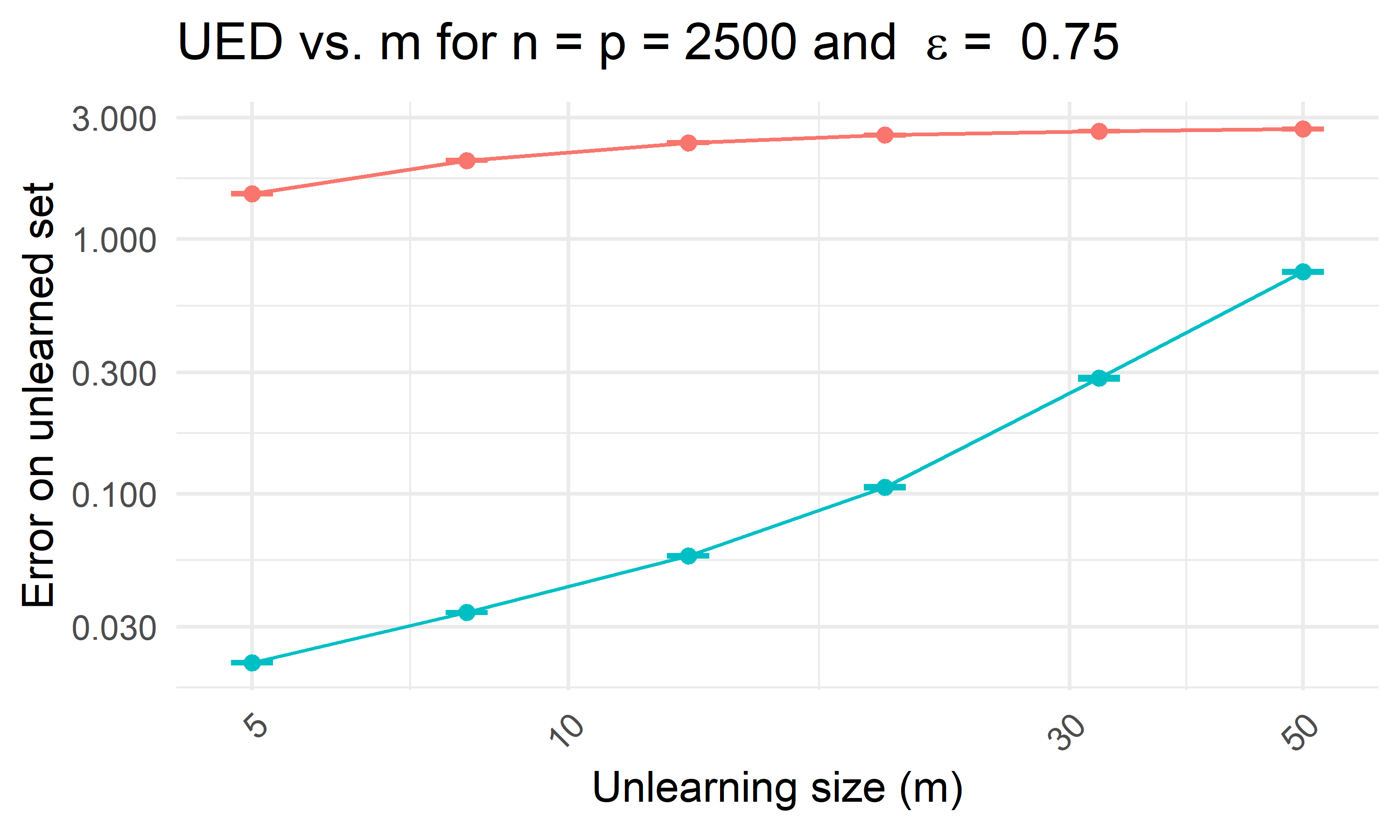}
    \caption{Comparison of UED (plotted in $\log$ scale) across the unlearning size $m$ (plotted in $\log$ scale) for Laplace noise (in red) vs. Gaussian noise (in cyan).}
    \label{fig:forget-across-m-del1}
\end{figure}

\textbf{Second set (meaning of figure 6):}  Now, we fix $\ve=0.75$ and study the change in UED as the unlearning size $m\uparrow$, varying $m$ across 6 values spaced equally between 5 and 50 on the logarithmic scale. The mean UED over 200 replications along with standard error bars are presented in Figure~\ref{fig:forget-across-m-del1}.

\subsection{Plots for n=p/2}
In this subsection we provide results for the analogous experiments in the high dimensional regime where $n=p/2$. It is evident from our results that the conclusions remain the same.

\begin{figure}[htbp]
    \centering
    \includegraphics[width=0.48\linewidth]{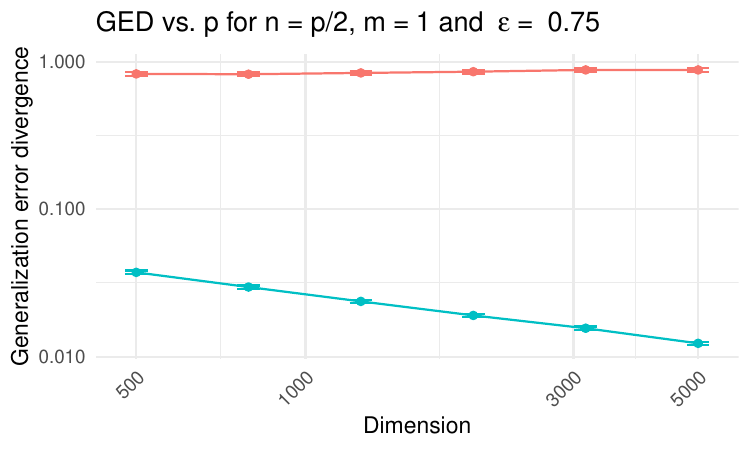}
    \includegraphics[width=0.48\linewidth]{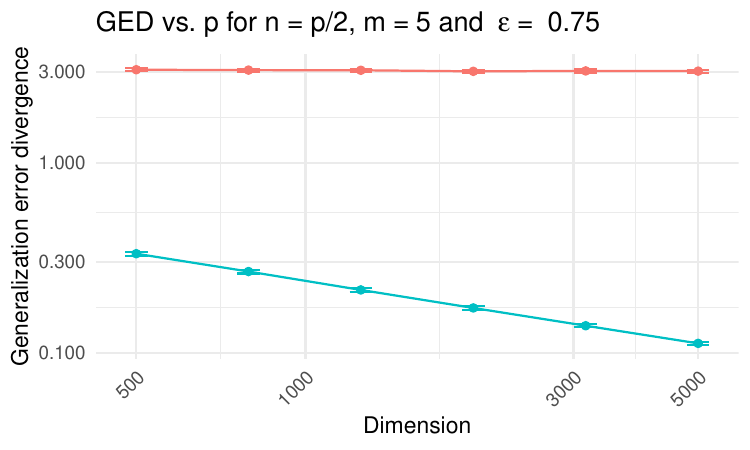}
    \includegraphics[width=0.48\linewidth]{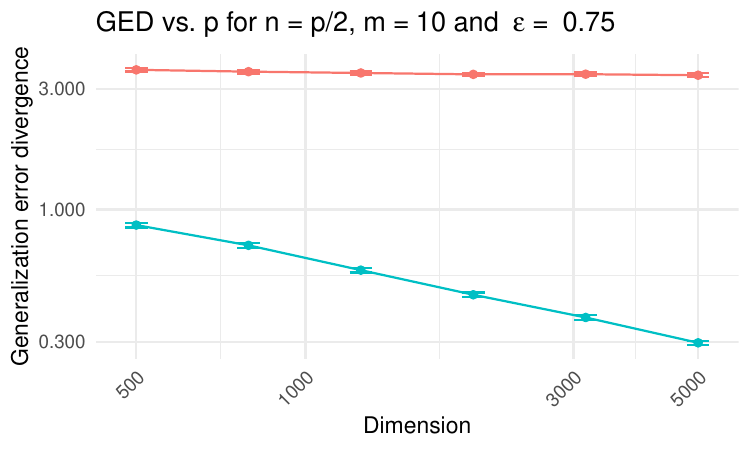}
    \caption{Comparison of unlearned estimators on new test data: mean GED (with 3 SD error bars) across the dimension $p$ (both in $\log$ scale) for Laplace (in red) vs. Gaussian (in cyan). We set $\lambda=0.2$.}
    \label{fig:ged-across-p-del0_5}
\end{figure}

\begin{figure}[htbp]
   \centering
    \includegraphics[width=0.48\linewidth]{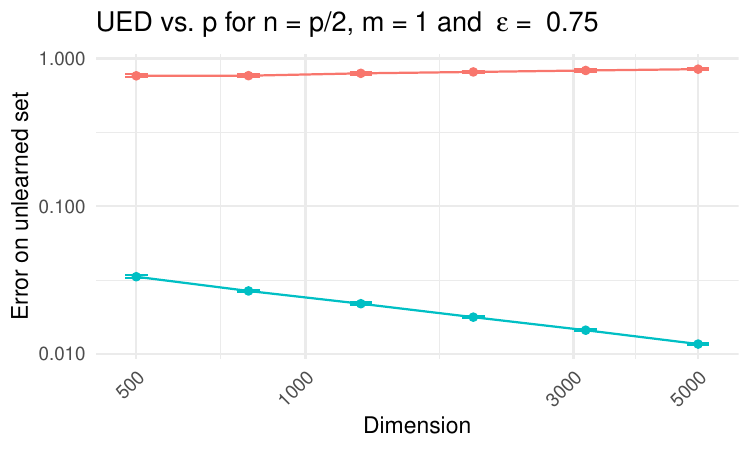}
    \includegraphics[width=0.48\linewidth]{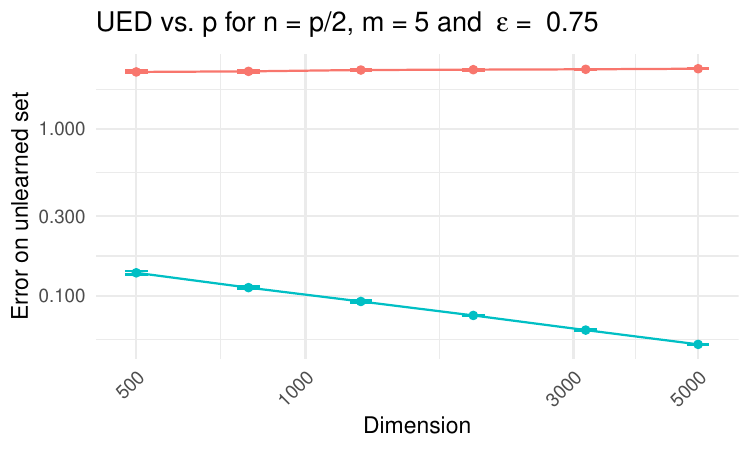}
    \includegraphics[width=0.48\linewidth]{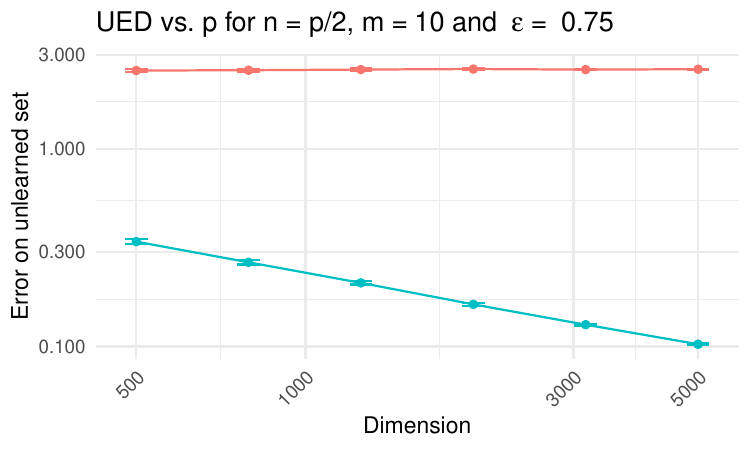}
    \caption{
    Comparison on unlearned data: mean UED (with 3 SD error bars) on the unlearned set across $p$ (both in $\log$ scale) for Laplace  (in red) vs. Gaussian noise (in cyan). We set $\lambda=0.2$.}
    \label{fig:forget-across-p-del0_5}
\end{figure}

\textbf{First set (meaning of figures 7 and 8):} We plot GED as we vary $p$ on the X axis and $n=p/2$ on a logarithmic scale, while keeping $m$ fixed at $1,5,10$ across the three plots. We find the mean GED values across 200 replications decrease with $p$ for Gaussian noise, but stay more or less the same for Laplace noise. Results are plotted in Figure~\ref{fig:ged-across-p-del0_5} and Figure~\ref{fig:forget-across-p-del0_5}.

\begin{figure}[htbp]
    \centering
    \includegraphics[width=0.48\linewidth]{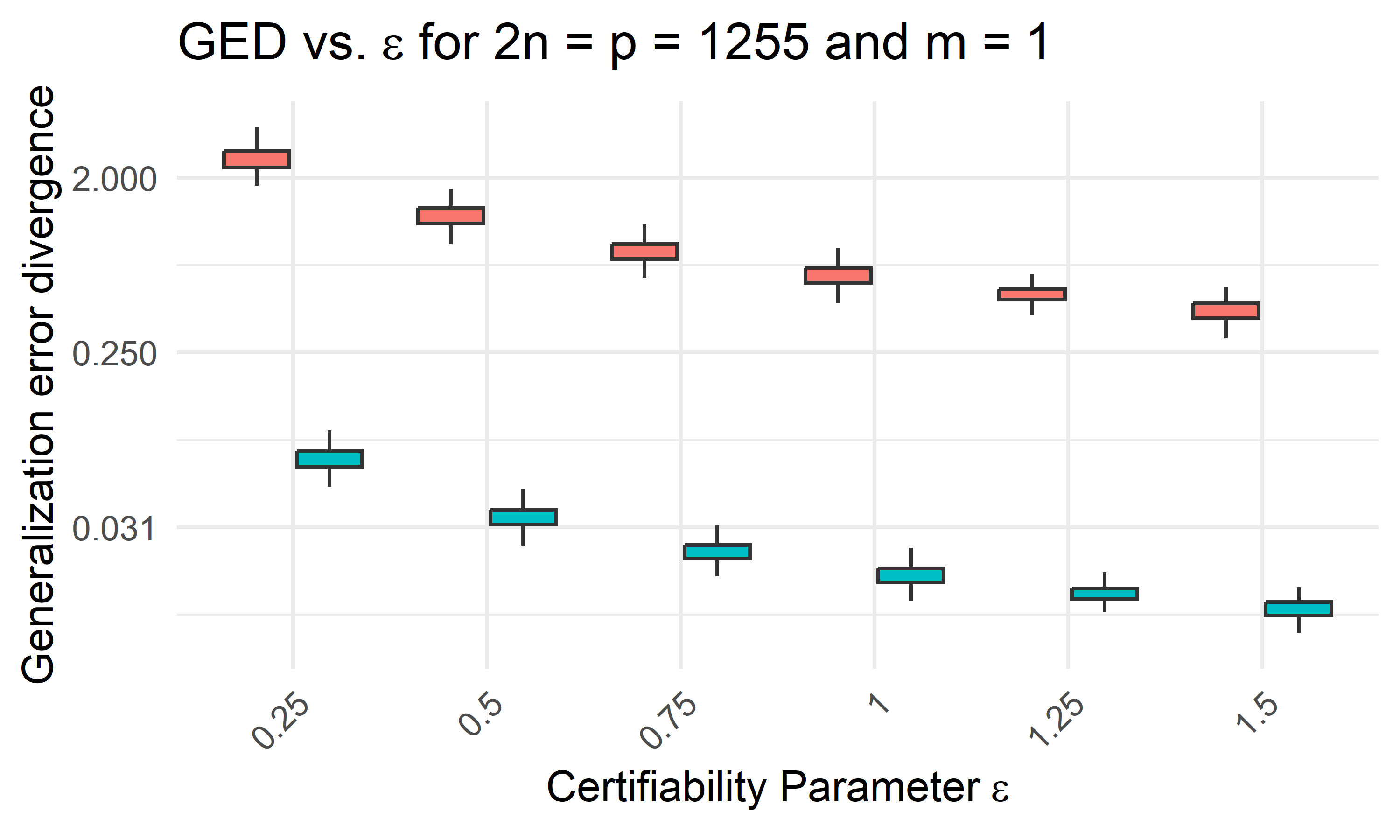}
    \includegraphics[width=0.48\linewidth]{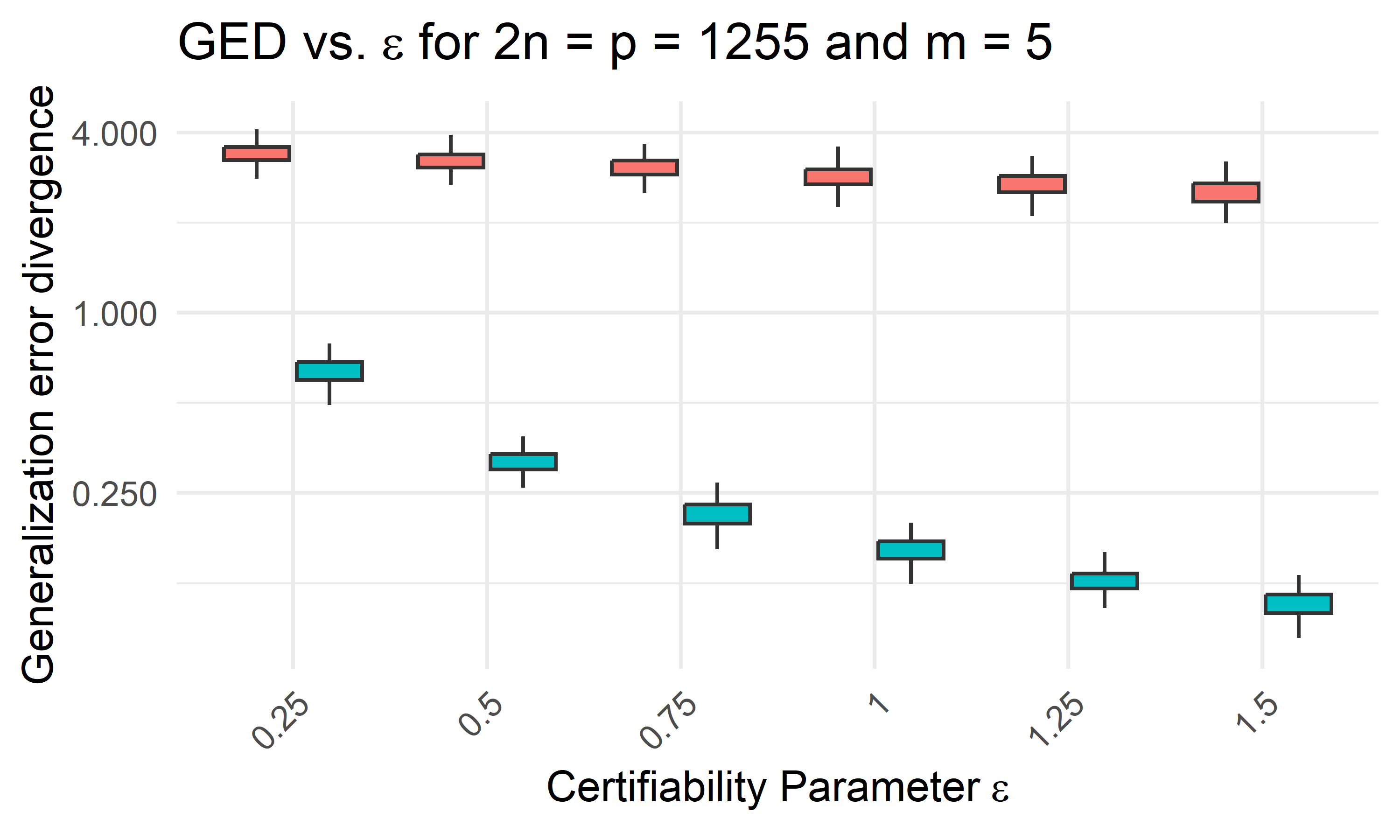}
    \includegraphics[width=0.48\linewidth]{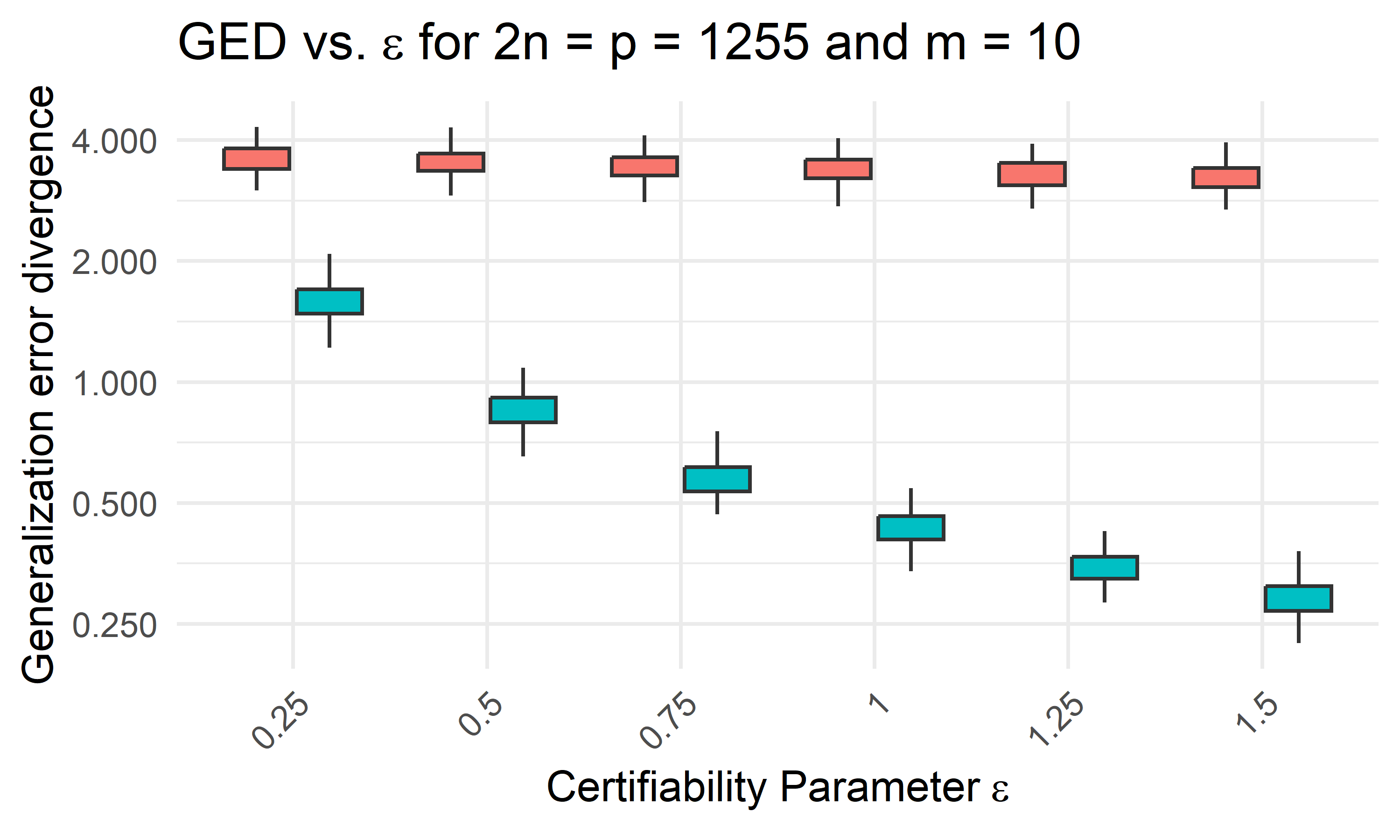}
    \caption{Comparison of GED (plotted in $\log$ scale) across different values of $\ve$ for Laplace noise (in red) vs. Gaussian noise (in cyan). Here $2n=p=1255$. We set $\lambda=0.2$.}
    \label{fig:ged-across-eps-del0_5}
\end{figure}

\begin{figure}[htbp]
    \centering
    \includegraphics[width=0.48\linewidth]{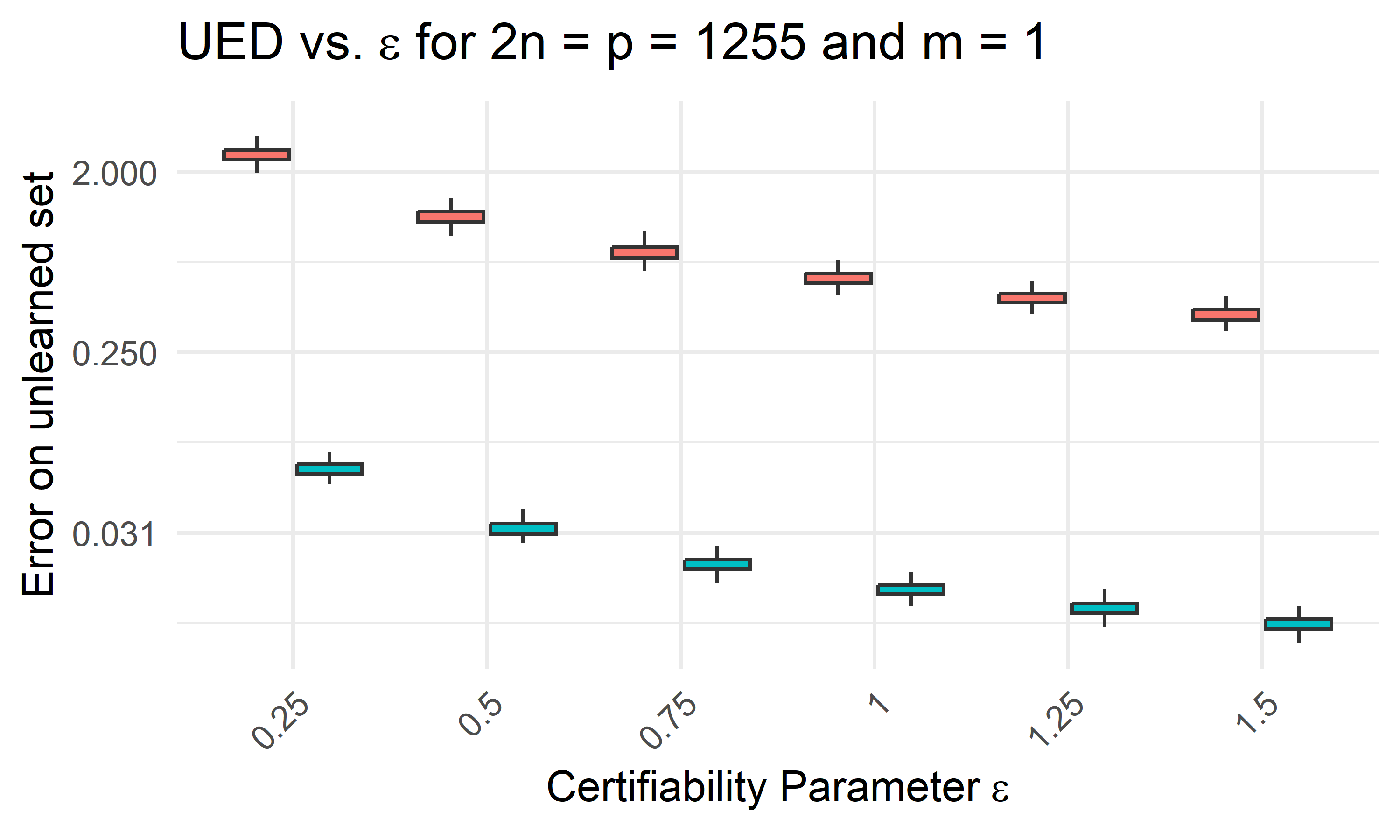}
    \includegraphics[width=0.48\linewidth]{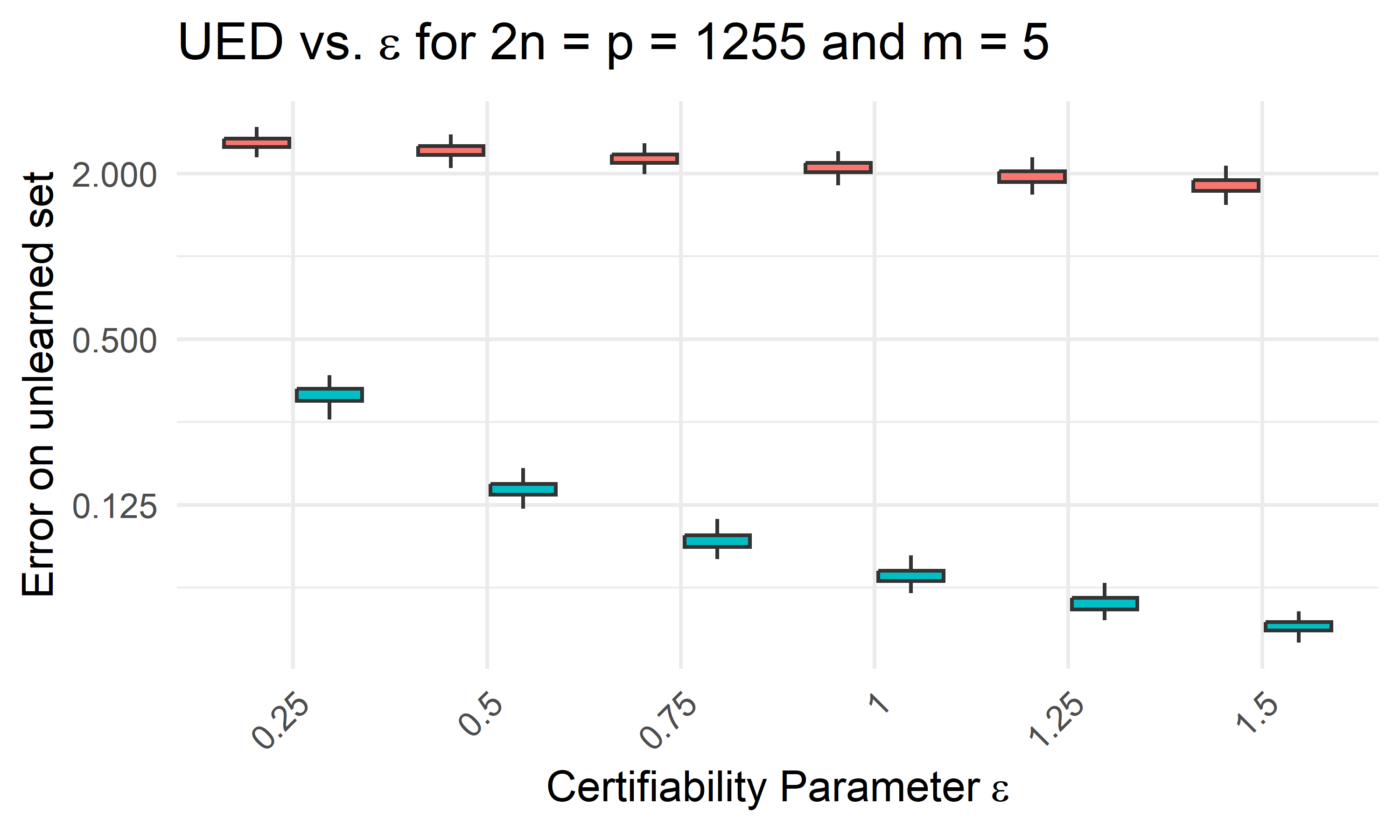}
    \includegraphics[width=0.48\linewidth]{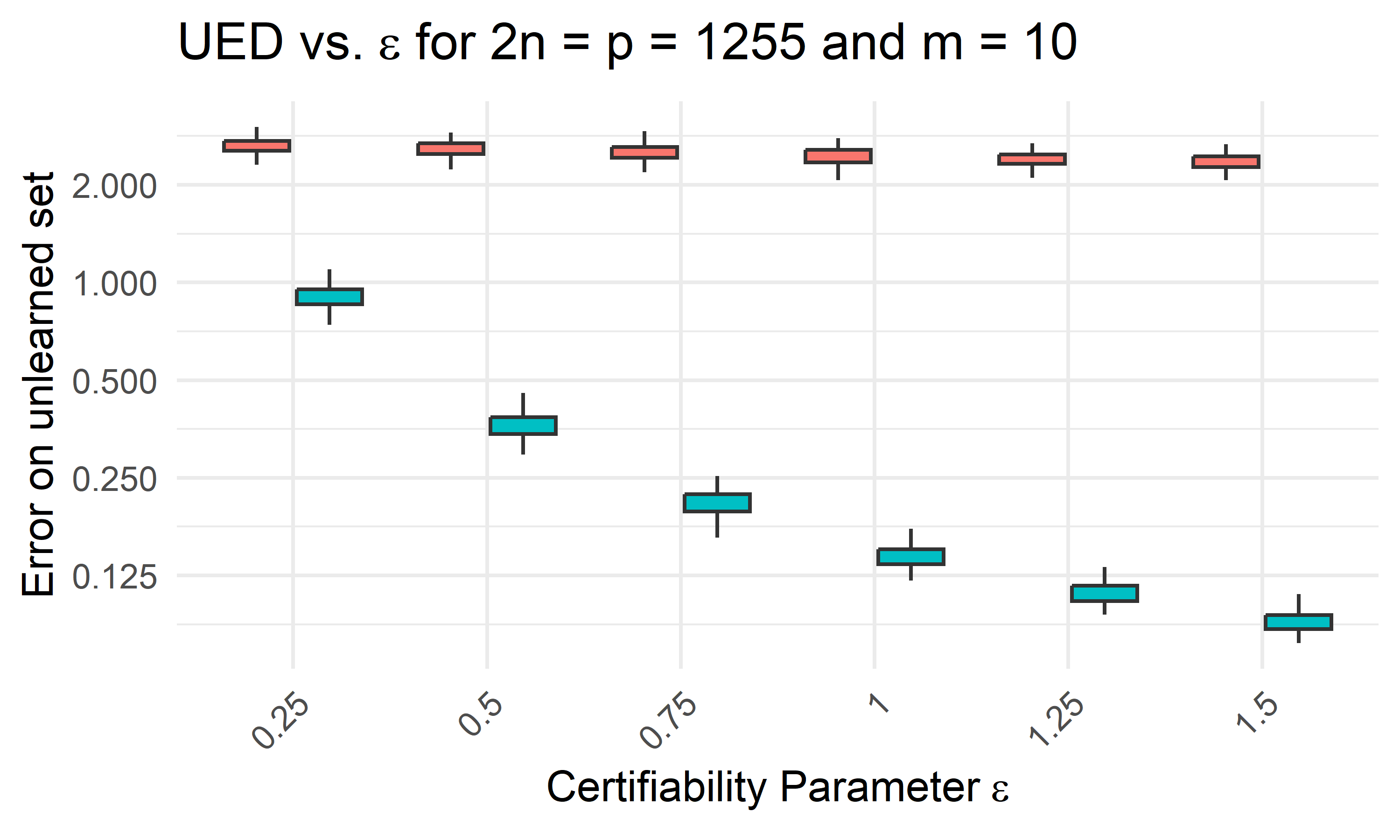}
    \caption{Comparison of the difference in negative log likelihood on the requested removal set (plotted in $\log$ scale) among the retrained estimator, with unlearned estimator with Laplace noise (in red) and Gaussian noise (in cyan). The left, middle, and right figures plot unlearning for removal sizes $m=1$, $m=5$, and  $m=10$ respectively. Here $2n=p=1255$.}
    \label{fig:forget-across-eps-del0_5}
\end{figure}

\textbf{Second set (meaning of figures 9 and 10):} Now, we fix $2n=p=1255$ and examine GED as we vary $\ve$ on the $X$ axis, while $m$ varies between $m=1$ (left), $m=5$ (middle), and  $m=10$ (right). Figure~\ref{fig:ged-across-eps-del0_5} and Figure~\ref{fig:forget-across-eps-del0_5} report boxplots of GED in this setting across 200 Monte Carlo replications.

\textbf{Conclusions from figures 9 and 10:} The performance of Gaussian perturbed estimator becomes increasingly closer to that of the ideal retrained estimator as $\ve \uparrow$ on the X axis. The error for the Laplace perturbed estimator remains above its Gaussian counterpart, uniformly over $\ve$. Similar behavior is reflected on the unlearned subset as well. Results are in Figure~\ref{fig:ged-across-eps-del0_5} and Figure~\ref{fig:forget-across-eps-del0_5}.

\begin{figure}[htbp]
    \centering
    \includegraphics[width=0.48\linewidth]{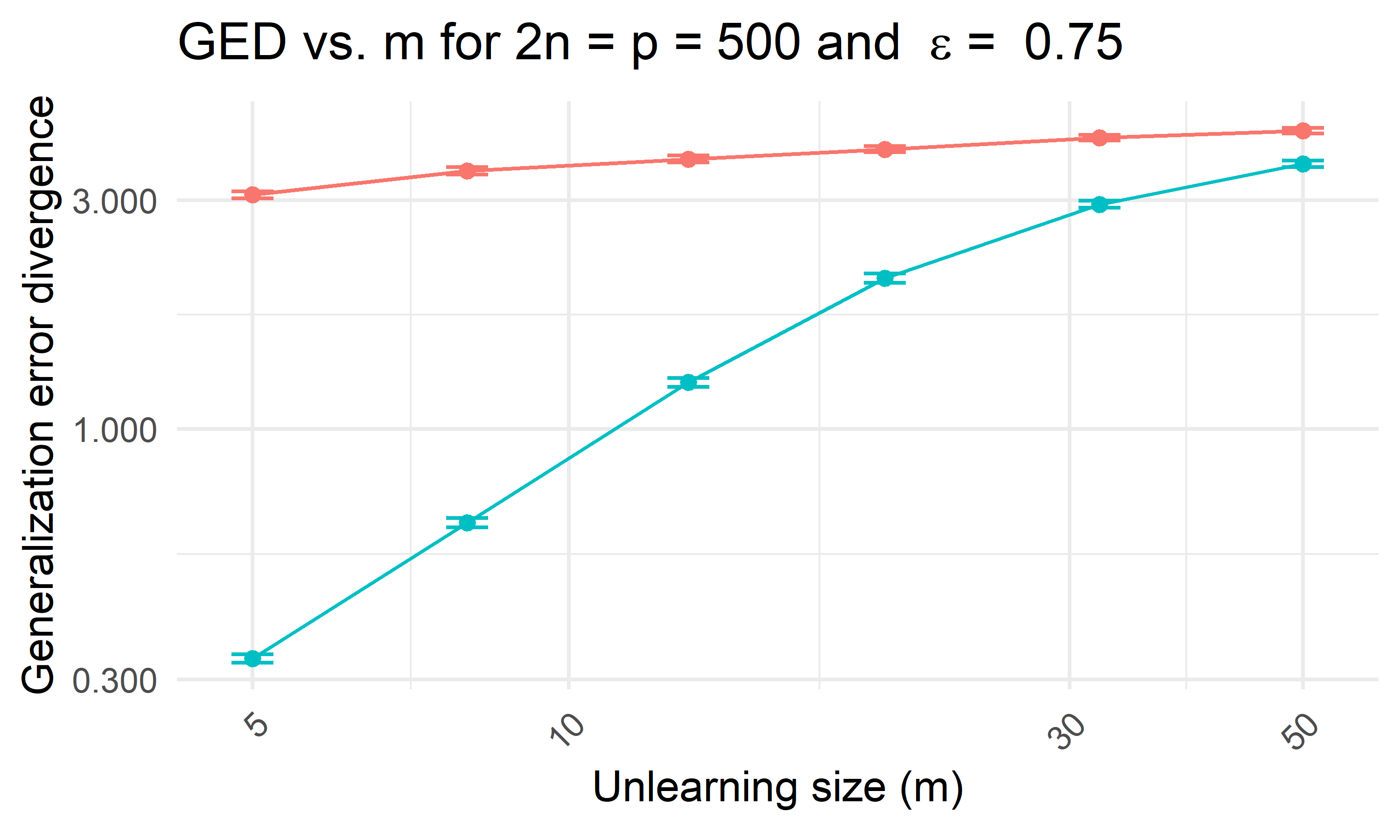}
    \includegraphics[width=0.48\linewidth]{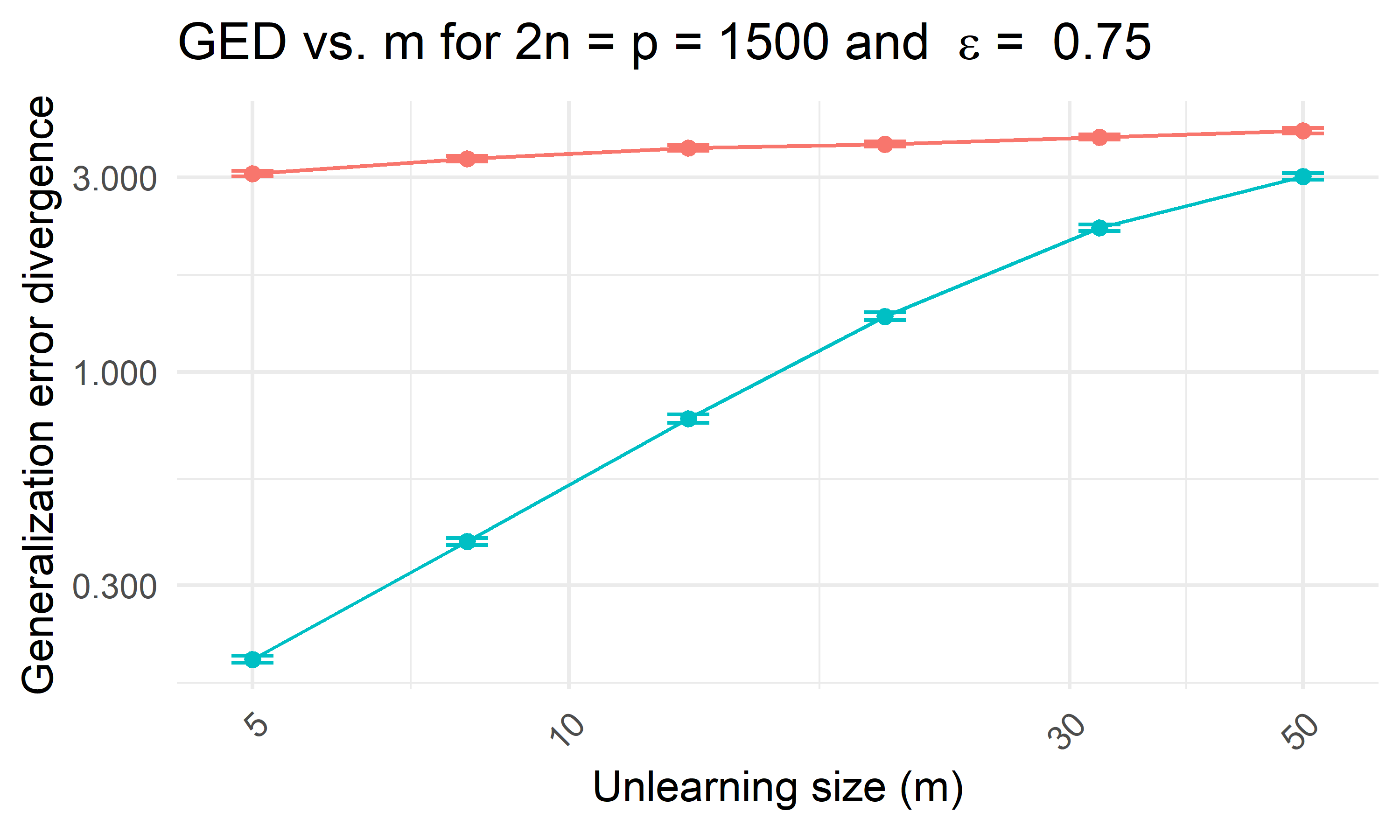}
    \includegraphics[width=0.48\linewidth]{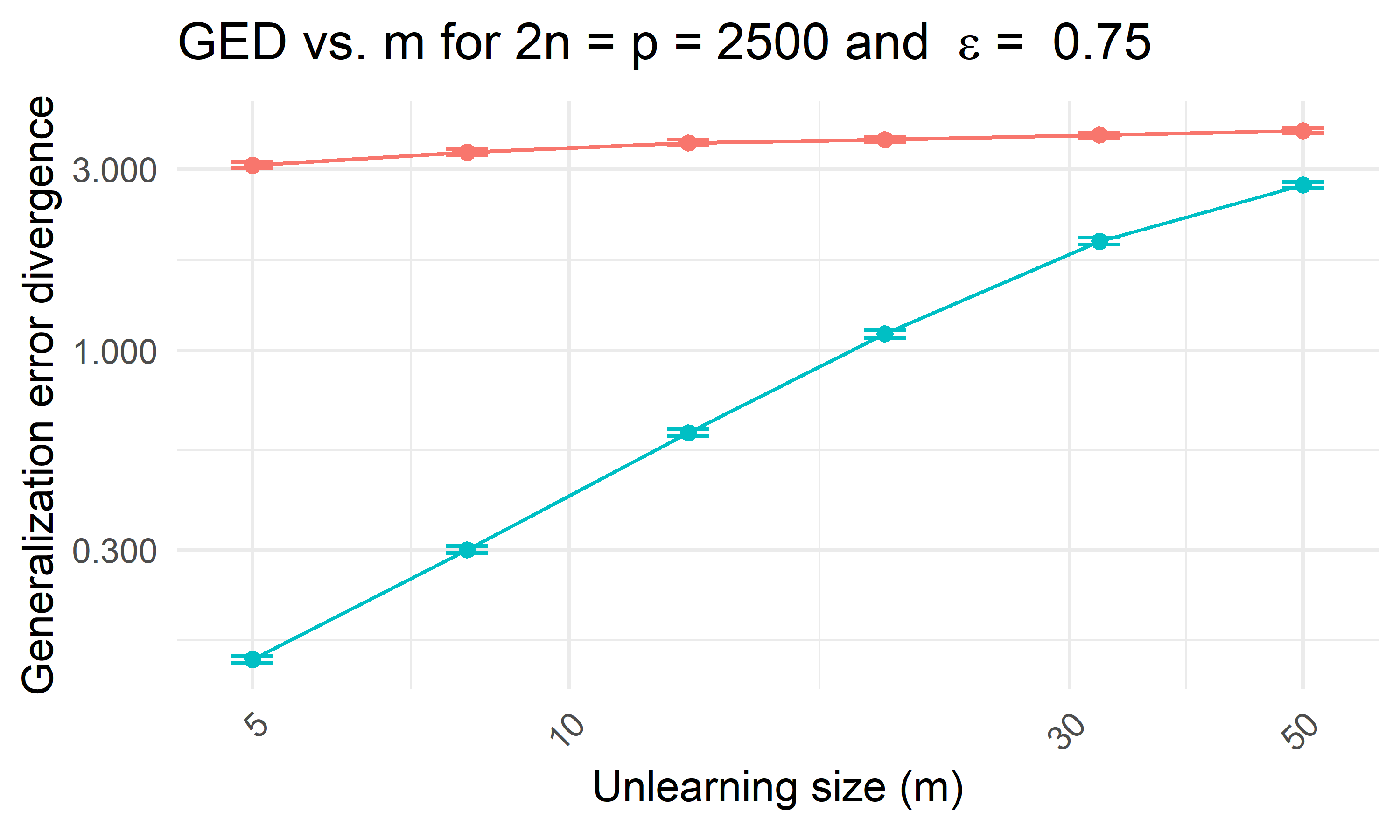}
    \caption{Comparison of mean GED (with 3 SD error bars) across the unlearning size $m$ (both in $\log$ scale) for Laplace noise (in red) vs. Gaussian noise (in cyan). We set $\lambda=0.2$.}
    \label{fig:ged-across-m-del0_5}
\end{figure}

\textbf{Final set (meaning of figures 11 and 12):} In our final set of experiments, we fix the certifiability parameter $\ve=0.75$, and examine GED  as the unlearning size $m$ varies on the X axis, while $n,p$ varies between $2n=p=500$ (left), $2n=p=1500$ (middle), and  $2n=p=2500$ (right). We vary $m$ across 6 values from 5 to 50 equally spaced on a logarithmic scale. Figure~\ref{fig:ged-across-m-del0_5} plots the average GED values (averaged across 200 replications) against dimension $m$, transforming both X and Y axes to be on a logarithmic scale.

\begin{figure}[htbp]
    \centering
    \includegraphics[width=0.48\linewidth]{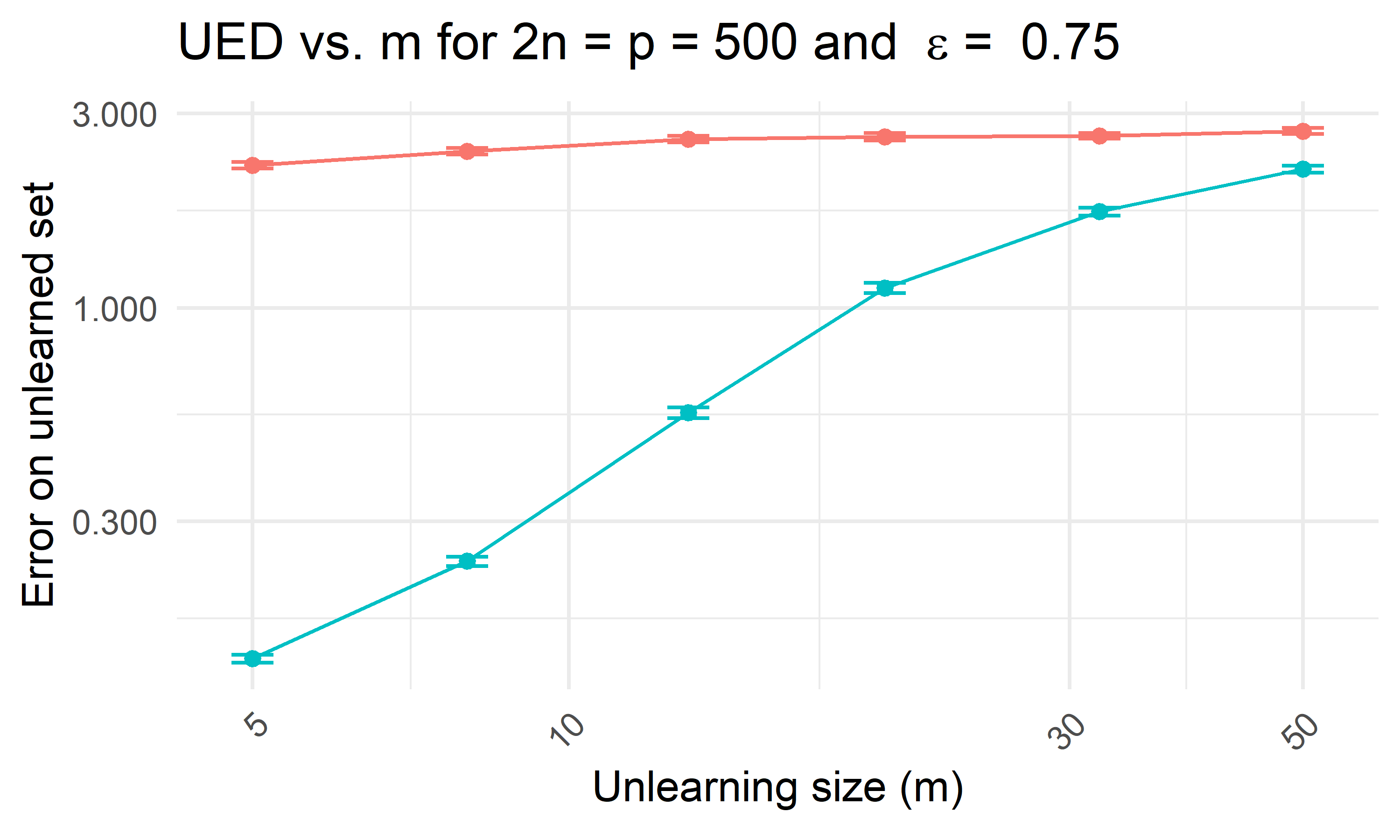}
    \includegraphics[width=0.48\linewidth]{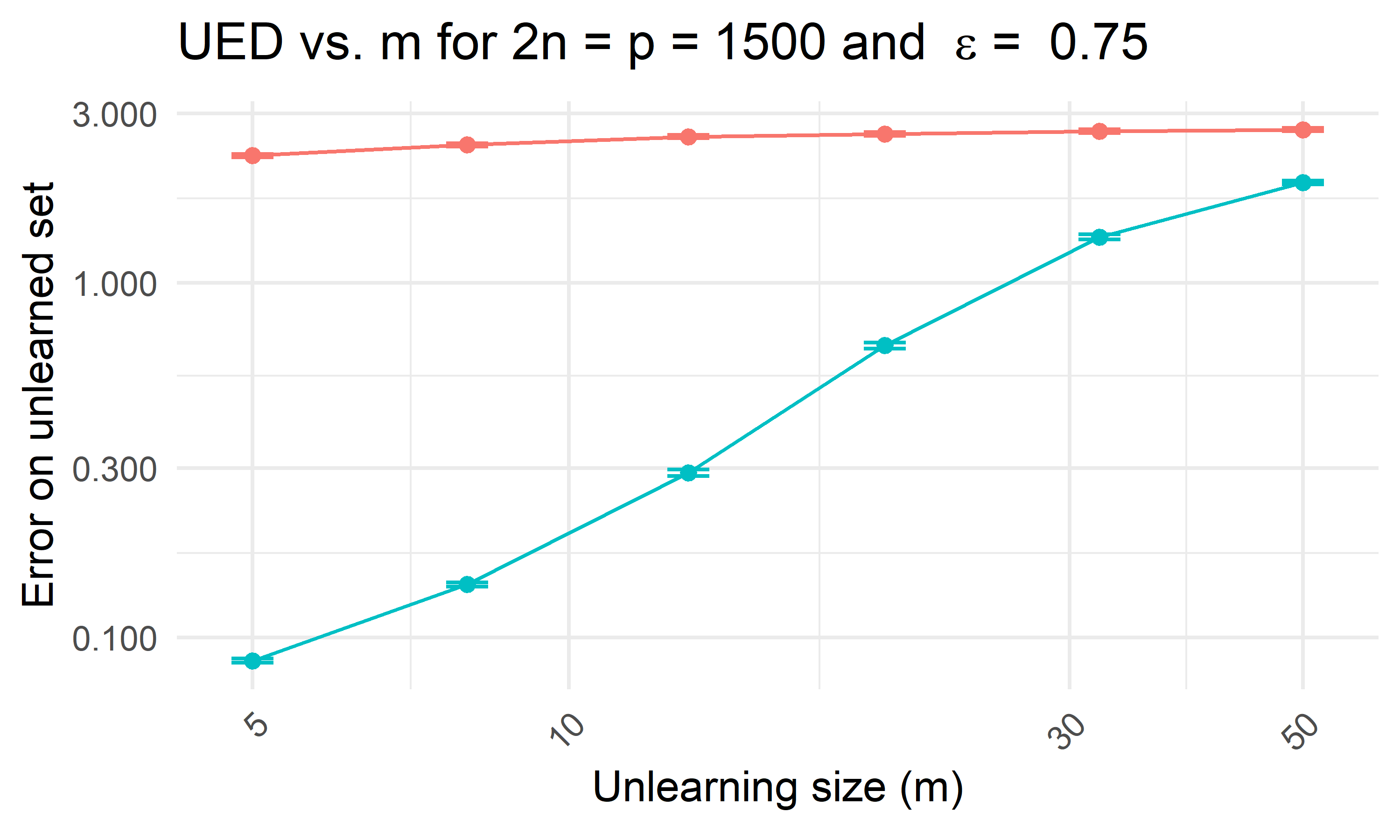}
    \includegraphics[width=0.48\linewidth]{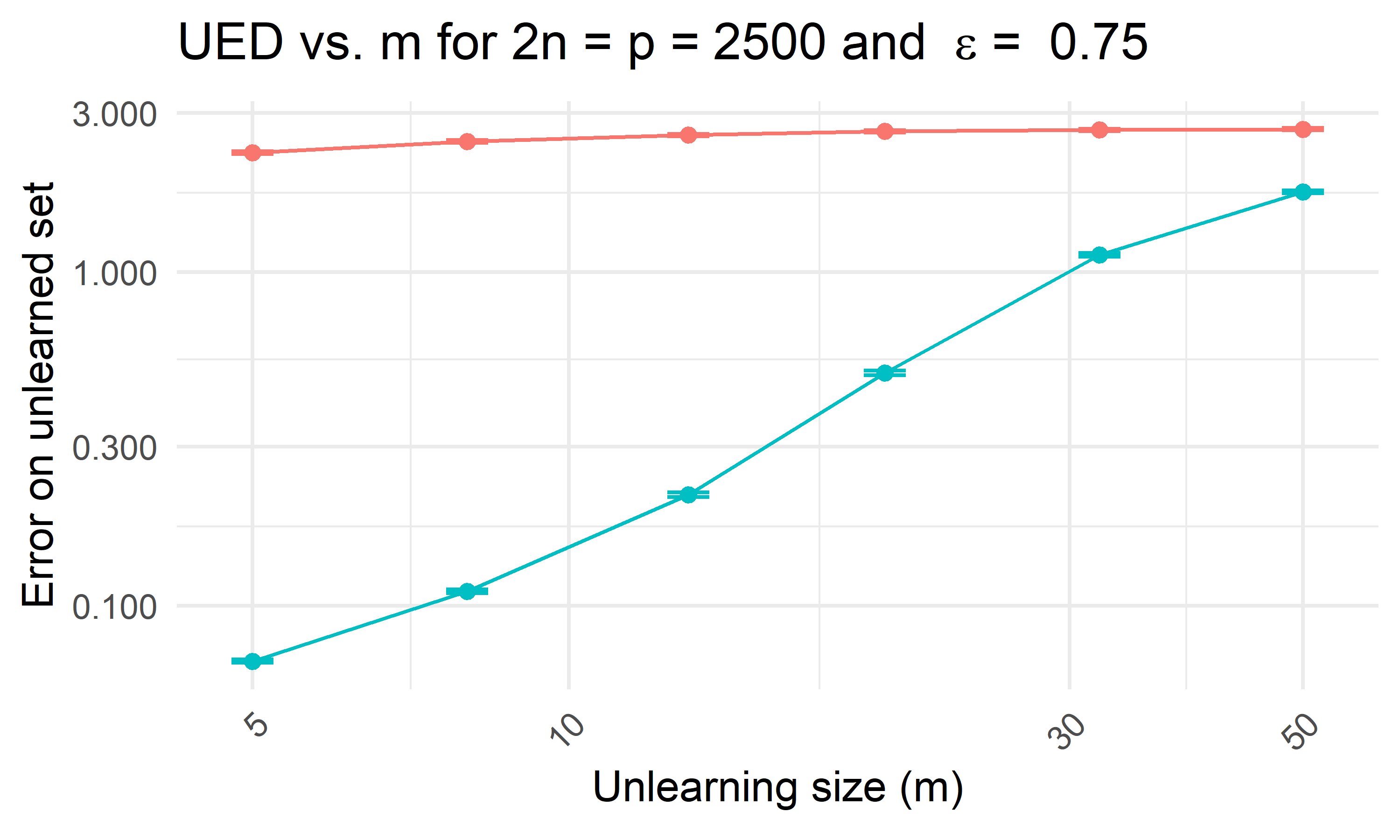}
    \caption{Comparison of UED (plotted in $\log$ scale) across the unlearning size $m$ (plotted in $\log$ scale) for Laplace noise (in red) vs. Gaussian noise (in cyan).}
    \label{fig:forget-across-m-del0_5}
\end{figure}

\textbf{Conclusions from figures 11 and 12:} For all three plots, as $m$ increases, unlearning accuracy as reflected in GED, worsens for both the estimators. Results are in Figure~\ref{fig:ged-across-m-del0_5} and Figure~\ref{fig:forget-across-m-del0_5}.

\end{document}